\pdfoutput=1
\documentclass[10pt,twocolumn,letterpaper]{article}
\usepackage[pagenumbers]{cvpr}

\usepackage{graphicx}	
\usepackage{amsmath}	
\usepackage{amssymb}	
\usepackage{booktabs}
\usepackage{times}
\usepackage{microtype}
\usepackage{epsfig}
\usepackage{caption}
\usepackage{float}
\usepackage{placeins}
\usepackage{color, colortbl}
\usepackage{stfloats}
\usepackage{enumitem}
\usepackage{tabularx}
\usepackage{xstring}
\usepackage{multirow}
\usepackage{xspace}
\usepackage{url}
\usepackage{subcaption}
\usepackage{listings}
\usepackage[hang,flushmargin]{footmisc}
\usepackage{algorithmicx,algorithm,algpseudocode}
\usepackage{longtable}

\definecolor{light-gray}{RGB}{245, 245, 245}   
\definecolor{purple-keyword}{RGB}{136, 18, 128} 
\definecolor{green-string}{RGB}{0, 128, 0}     
\definecolor{black-identifier}{RGB}{36, 41, 46} 
\definecolor{gray-comment}{RGB}{106, 153, 85}  
\definecolor{blue-function}{RGB}{0, 92, 197}   
\definecolor{brilliantlavender}{rgb}{0.96, 0.73, 1.0}

\definecolor{coverage_by_input}{rgb}{229,239,225}
\definecolor{coverage_by_answer}{rgb}{230,221,231}
\definecolor{coverage_by_answer_then_input}{rgb}{182,164,195}
\definecolor{sdv14}{rgb}{246,233,241}
\definecolor{sdxl3}{rgb}{222,183,208}
\definecolor{lmguided}{rgb}{176,89,141}

\lstdefinestyle{onelight}{
    language=Python,
    backgroundcolor=\color{light-gray},           
    basicstyle=\ttfamily\fontsize{8}{8}\selectfont\color{black},      
    keywordstyle=\color{purple-keyword}\bfseries, 
    stringstyle=\color{green-string},             
    commentstyle=\color{gray-comment}\itshape,    
    emph={ImagePatch,__init__,print},          
    emphstyle=\color{blue-function},
    identifierstyle=\color{gray}, 
    numberstyle=\tiny\color{gray},                
    showstringspaces=false,                       
    tabsize=4,                                    
    frame=none,                                   
    breaklines=true,                              
    breakatwhitespace=true,
    morekeywords={self},                          
    alsoletter={.} ,
    escapeinside={||},
}
\usepackage{xr-hyper}

\makeatletter
\newcommand*{\addFileDependency}[1]{
  \typeout{(#1)}
  \@addtofilelist{#1}
  \IfFileExists{#1}{}{\typeout{No file #1.}}
}

\makeatother

\definecolor{cvprblue}{rgb}{0.21,0.49,0.74}
\usepackage[pagebackref,breaklinks,colorlinks,citecolor=cvprblue]{hyperref}
\usepackage[capitalize]{cleveref}
\crefname{section}{Sec.}{Secs.}
\crefname{table}{Table}{Tables}
\crefname{figure}{Fig.}{Figs.}

\def\viUnitTitle{\includegraphics[height=1.5\baselineskip]{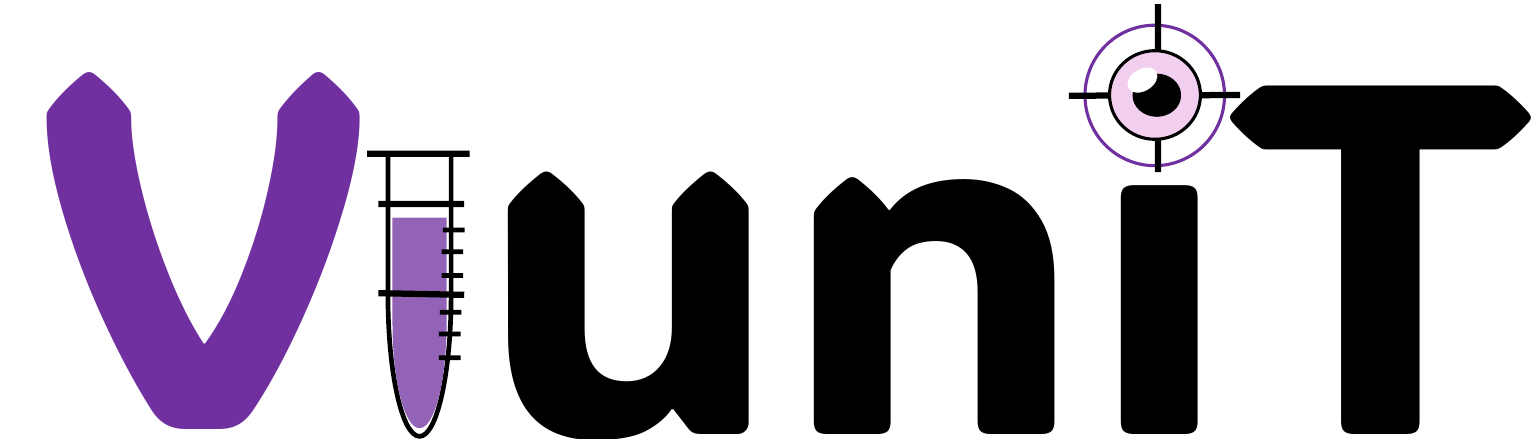}}

 \def\viUnit{\includegraphics[height=0.9\baselineskip]{figs/viunit_logov5.pdf}}

\def\paperTitle{\viUnitTitle: Visual Unit Tests for More Robust Visual Programming}

\def\authorBlock{
Artemis Panagopoulou$^{\dagger,}$\thanks{Work done during internship at Salesforce.} \qquad 
Honglu Zhou$^{\ddagger}$ \qquad 
Silvio Savarese$^{\ddagger}$ \qquad 
Caiming Xiong$^{\ddagger}$\\ 
Chris Callison-Burch$^{\dagger}$ \qquad 
Mark Yatskar$^{\dagger}$ \qquad 
Juan Carlos Niebles$^{\ddagger}$ \\ 
$^\ddagger$Salesforce AI Research \qquad 
$^\dagger$University of Pennsylvania \\
\href{https://artemisp.github.io/viunit/}{\texttt{https://artemisp.github.io/viunit/}}
}

\begin{document}
\title{\paperTitle}
\author{\authorBlock}
\maketitle
\begin{abstract}
Programming based approaches to reasoning tasks have substantially expanded the types of questions models can answer about visual scenes.
Yet on benchmark visual reasoning data, when models answer correctly, they produce incorrect programs 33\% of the time.
These models are often right for the wrong reasons and risk unexpected failures on new data.
Unit tests play a foundational role in ensuring code correctness and could be used to repair such failures.
We propose Visual Unit Testing (ViUniT), a framework to improve the reliability of visual programs by automatically generating unit tests.
In our framework, a unit test is represented as a novel image and answer pair meant to verify the logical correctness of a program produced for a given query.
Our method leverages a language model to create unit tests in the form of image descriptions and expected answers and image synthesis to produce corresponding images.
We conduct a comprehensive analysis of what constitutes an effective visual unit test suite, exploring unit test generation, sampling strategies, image generation methods, and varying the number of programs and unit tests. 
Additionally, we introduce four applications of visual unit tests: best program selection, answer refusal, re-prompting, and unsupervised reward formulations for reinforcement learning. 
Experiments with two models across three datasets in visual question answering and image-text matching demonstrate that ViUniT improves model performance by 11.4\%.
Notably, it enables 7B open-source models to outperform gpt-4o-mini by an average of 7.7\% and reduces the occurrence of programs that are correct for the wrong reasons by 40\%.
\end{abstract}

\section{Introduction}
\label{sec:intro}
\begin{figure*}[t]
    \centering
    \includegraphics[width=\linewidth]{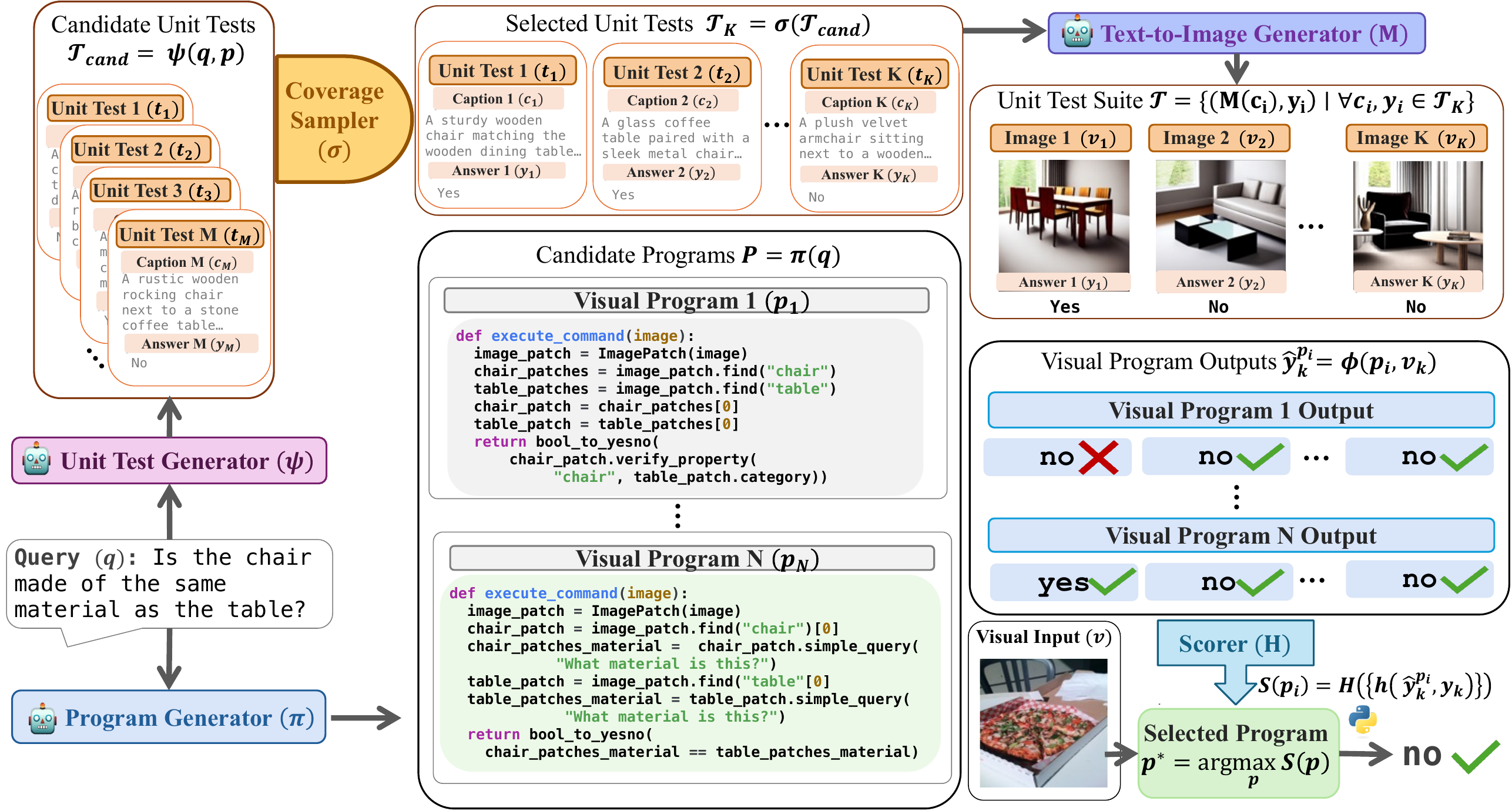}
    \vspace{-0.7cm}
    \caption{\viUnit~Framework Overview. Given a query \( q \) about an image, the unit test generator \( \psi \) generates a set \( \mathcal{T}_{\text{cand}} = \psi(q, p) \) of \( M \) candidate pairs \( t_i = (c_i, y_i) \), each consisting of an image caption \( c_i \) and an expected answer \( y_i \)  (Section~\ref{sec:candidates}). The coverage sampler \( \sigma \) then subsamples \( K \) pairs from \( \mathcal{T}_{\text{cand}} \), forming the subset \( \mathcal{T}_K \) (Section~\ref{sec:sampling}). These captions are passed to an image generator \(M\) to create the corresponding images \(v_i= M(c_i)\) for each unit test (Section~\ref{sec:image}). Each candidate program is subsequently executed, and gets assigned a score  \( S(p) \) by the scorer \( H \) based on its performance on the unit tests (Section~\ref{sec:unit_test_scoring}). Finally, the highest scoring program is selected.}
    \label{fig:framework_fig}
    \vspace{-3mm}
\end{figure*}

Visual Programming~\citep{gupta2023visual,suris2023vipergpt}, which involves generating executable programs that leverage state-of-the-art specialist systems (e.g. object detection, captioning, etc.), has emerged as an effective method for tackling compositional reasoning tasks~\citep{stanictowards,hanimage}.
Often correct visual programs must be inferred without training programs because they are expensive to annotate.
Recently, some methods improve the performance of visual program synthesis by leveraging programs that yield correct results on training data~\citep{khan2024self,li2024synthesize}. 
While these approaches have shown improvements, a critical limitation persists: \textbf{visual programs can be right for the wrong reasons.} 
For example, human evaluation of 100 visual programs resulting in correct responses generated by CodeLlama-7B\footnote{CodeLlama-7B~\citep{roziere2023code} is a leading open source large language model (LLM)} for questions in GQA~\cite{hudson2019gqa}, showed that only 33\% of them were actually correct and 70\% of the incorrect programs (23\% of the total) would require significant rewriting to be correct.  

To mitigate this prevailing issue, we propose \textbf{Visual Unit Testing}~(\viUnit), a framework for automatically generating unit tests for visual programs. 
While automatic unit test generation has gained momentum in text-based tasks~\citep{chen2022codet,siddiq2023exploring,alagarsamy2023a3test,guilherme2023initial,takerngsaksiri2024tdd}, its application to visual program synthesis has been limited. 
Recent efforts toward visual units tests focused primarily on checking program return value types~(e.g.~the output falling outside a range of options, like yes or no)~\citep{koo-etal-2024-proptest}. 
However, this approach does not assess the program's execution or logical correctness, limiting the types of errors it can address.
In this work, we bridge this gap by addressing challenges that have hindered unit test use in visual question answering~(VQA) and image-text-matching~(ITM).

As seen in Figure~\ref{fig:framework_fig}, visual programming converts  queries  to code that executes on test images to provide a response. 
For such programs, unit tests take the form of images and expected answers.
Units test are difficult to construct because they need to have sufficient coverage to diagnose errors.
To solve this problem, we leverage language models to generate candidate sets of descriptions of images that could test the code (Section \ref{sec:candidates}).
We formulate an optimization criterion to select ones that maximize coverage of possible program inputs and outputs (Section \ref{sec:sampling}), and convert selected descriptions to images (Section \ref{sec:image}).
Our approach is entirely unsupervised with no accompanying annotations.

\begin{figure*}[tb]
    \centering
    \includegraphics[width=\linewidth]{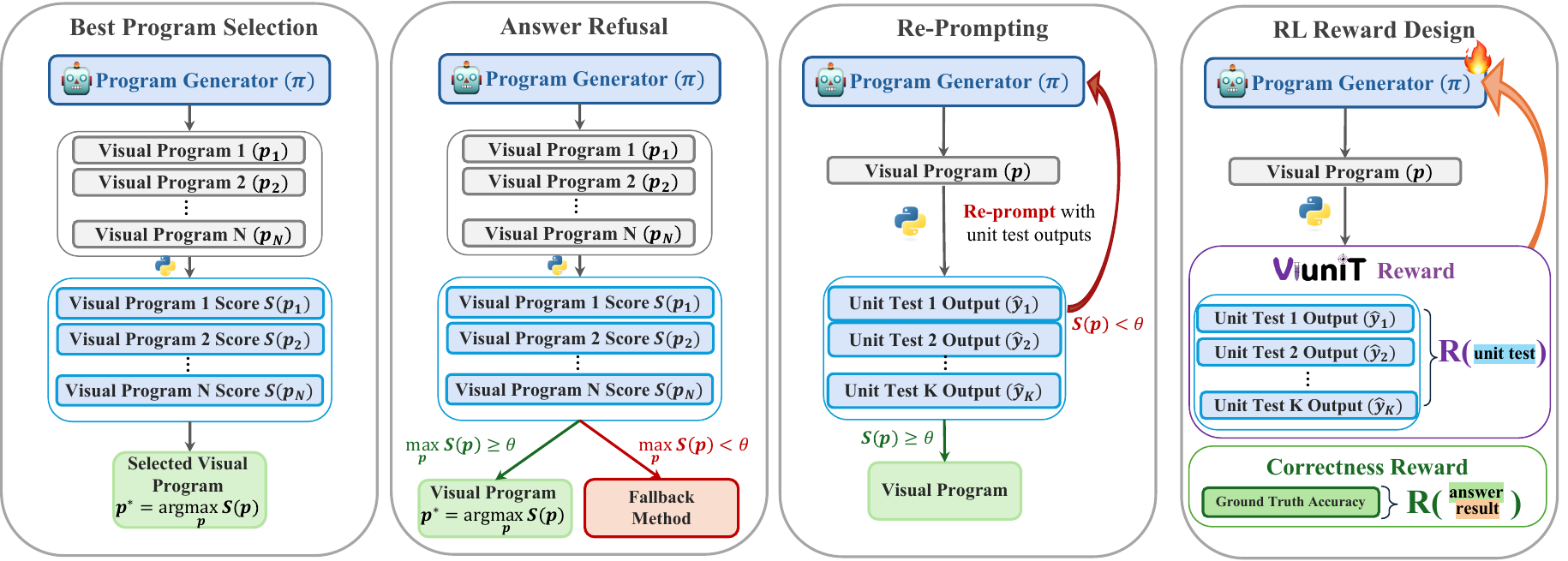}
    \vspace{-.6cm}\caption{Visual Unit Testing Utilization Strategies (Section \ref{sec:method_applicatons}). }
    \label{fig:applications}
    \vspace{-5mm}
\end{figure*}

Unit tests can be used to identify incorrect programs but integrating this signal to improve model behavior is challenging. 
In Section \ref{sec:method_applicatons} we explore several mechanisms, summarized in Figure~\ref{fig:applications}, including:

\begin{enumerate}[wide, labelindent=0pt]
\item \textbf{Best program selection}: Given a set of program candidates we select the one that passes the most test cases. This approach achieves a 7.7-point improvement over \texttt{gpt-4o-mini} (Table \ref{tab:app_best_program}) and reduces  right-for-wrong-reason programs by 40\% (Section \ref{sec:human_evaluation}).
\item \textbf{Re-prompting}: We use unit test outputs to guide the generation of improved programs when initial programs perform poorly on the unit test suite. Relative to regeneration without unit tests, programs are over 3\% more accurate (Table \ref{tab:performance_reprompting}).
\item \textbf{Unsupervised Reinforcement Learning (RL) Reward Design}: We use unit test scores as feedback to fine-tune an LLM on programs more likely correct for the right reasons, surpassing supervised correctness-based rewards by an average of 1.3 points across tasks (Table \ref{tab:performance_rl}).
\item \textbf{Answer refusal}: Unit test scores are used to assess program confidence, reverting to an end-to-end model if the program is not robust, achieving up to 0.8 F1 score in correctly refusing programs that would fail (Figure \ref{fig:refusal_plot}).
\end{enumerate}

To summarize our contributions, we present \viUnit, the first framework to introduce unit tests that verify the logical correctness of visual programs.
We conduct a broad exploration of unit test generation configurations (Section \ref{sec:ablations}), showing that maximizing coverage is an important criterion.
We introduce four ways to leverage unit-tests to improve models (Section \ref{sec:method_applicatons}): best program selection, answer refusal, re-prompting, and unsupervised reward design for reinforcement learning.
Overall, integrating unit-tests improves frozen-LLM accuracy by 11.4\% and enables 7B open-source LLMs to outperform proprietary models like \texttt{gpt-4o-mini} by an average of 7.7 points, while improving 
underlying code correctness. 
Broader adoption of unit-test suits will significantly enchase robustness and trust of visual programming approaches.

\section{Related Work}
\label{sec:related}\vspace{-.2cm}
\noindent\textbf{Visual Program Synthesis:} The recent advancements in LLMs~\citep{achiam2023gpt,longpre2023flan,touvron2023llama,touvron2023llama2,brown2020language,anil2023palm,jiang2023mistral,nijkamp2023xgen} have led to their use as a planning interface for the modularization of tools to execute complex reasoning tasks involving multiple modalities~\cite{lu2024chameleon,schick2024toolformer,singh2019towards,gao2024clova,li2024synthesize} and as a reasoning module for visual agents~\citep{yang2022empirical,yang2023mm,wei2024editable,hong2024cogagent}. Specialized coding LLMs~\citep{roziere2023code,le2022coderl,team2024codegemma,guo2024deepseek} have demonstrated significant potential in addressing visual challenges by generating executable code based on contextual demonstrations~\citep{suris2023vipergpt,gupta2023visual,hanimage,ge2025recursive,ukai2024adacoder} with comparable or better performance to vision language models~\citep{li2023blip,panagopoulou2023x,liu2024visual,dai2023instructblip}. Attempts to improve the initial paradigm involve automatically generating a pool of effective programs to retrieve as in-context examples~\citep{stanictowards} and tuning a model through reinforcement learning by sampling programs that succeed on the training set~\citep{khan2024self}. More relevant to this work, \citet{hu2024visual} distill program reasoning into a VLM as chain-of-thought reasoning by generating multiple programs per query and selecting the best one, either by using the ground truth answer as a proxy for correctness or by having it evaluated by an LLM.
However, a critical issue remains: some generated programs achieve correct outcomes without sound reasoning, which we address in this paper.

\noindent\textbf{LLM Unit Test Generation:} Unit tests have been used as reinforcement learning signal to train code-generating LLMs~\citep{le2022coderl,chen2022codet,shen2023pangu,guilherme2023initial,shojaeeexecution,dou2024stepcoder}. Existing methods for automatic unit test generation with LLMs \citep{chen2022codet,alagarsamy2023a3test,guilherme2023initial,takerngsaksiri2024tdd} focus primarily on text-based tasks, generating entire unit test scripts. However, these approaches often result in issues like compilation errors, low coverage, redundant assertions, and empty tests \citep{siddiq2023exploring}. Recent work \citep{koo-etal-2024-proptest} proposes property testing on the outputs of visual programs by leveraging LLMs to generate properties that should be satisfied by the output given the query (e.g. the output should be a color if the query asks for one). Yet, this method inherits many limitations of LLM generated script-based unit testing, and crucially, it fails to assess logical correctness—meaning it overlooks cases where program outputs may be right for the wrong reasons. Instead, we propose a method of generating unit tests to verify the \textit{execution} of visual programs, without requiring an LLM to directly generate unit-test scripts, avoiding such issues that tend to accompany the automatic generation of unit tests using LLMs. In particular, we use LLMs to generate image descriptions and expected answers without requiring any direct code generation. Image descriptions and expected answers are then transformed to a unit test using a text-to-image diffusion model~\citep{Rombach_2022_CVPR}.

\label{sec:applications}

\section{Method}\label{sec:method}
\vspace{-.2cm}
In this section, we formalize the tasks of visual program synthesis and unit test generation (Section~\ref{sec:task_definition}) and introduce our \viUnit~ framework (Section~\ref{sec:unit_test_generation}). 
Our method comprises two main components: unsupervised generation of visual unit tests (Section~\ref{sec:unit_test_generation}) and unit test scoring (Section~\ref{sec:unit_test_scoring}). We propose four ways to leverage unit tests in Section \ref{sec:method_applicatons}: Best Program Selection, Answer Refusal, Re-Prompting, and Unsupervised RL Reward Design. 

\vspace{-.1cm}
\subsection{Task Definition}\label{sec:task_definition}
\vspace{-.1cm}
\noindent\textbf{Visual Program Synthesis:} Given a visual input \( v \) and a textual query \( q \) about \( v \), our goal is to synthesize a program \( p \) that correctly answers \(q\) about \(v\). Each program \( p \in \mathcal{P} \) is executed on the visual input \( v \) using an execution engine \( \phi \), yielding a predicted answer \(\hat{y} = \phi(p, v)\). Our objective is to select the program \( p^\ast \) that is most likely to produce the correct answer \( y^\ast \) to the query \( q \) about \( v \), formalized as:
{\small \begin{align}\vspace{-.4cm}
p^\ast &= \arg\max_{p \in \mathcal{P}} \Pr\left( \phi(p, v) \equiv y^\ast \right).\label{eq:best_program}
\end{align}}

\noindent\textbf{Visual Unit Testing:} To assess the candidate programs, we employ a unit test generator \( \psi \), which generates a set of unit tests \(\mathcal{T} = \psi(q)\).
Each unit test \( t_i \in \mathcal{T} \) consists of a test visual input \( v_i \) and the corresponding correct answer \( y_i \) to the query \( q \) on that input \(t_i = (v_i, y_i)\).
For each candidate program \( p \in \mathcal{P} \), we execute it on all test inputs \( v_i \) to obtain outputs \(\hat{y}_i = \phi(p, v_i),~\text{for } t_i \in \mathcal{T}\).

\vspace{-.1cm}\subsection{Unsupervised Visual Unit Test Generation}\label{sec:unit_test_generation}\vspace{-.2cm}
\begin{figure*}[tb]
    \centering
    \includegraphics[width=\linewidth]{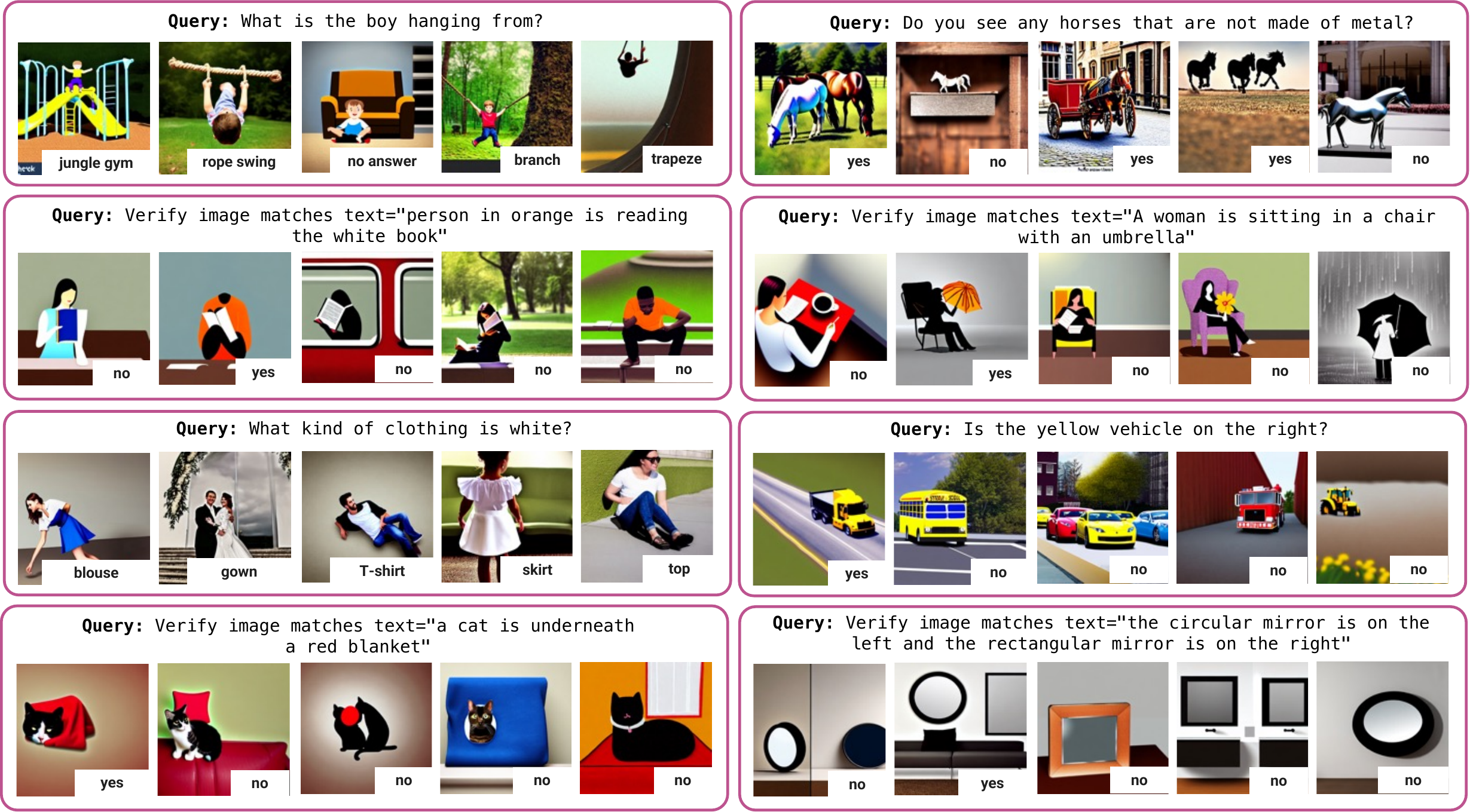}\vspace{-.3cm}
    \caption{Unit Test Examples generated by \viUnit}
    \label{fig:qualitative_examples}
    \vspace{-3mm}
\end{figure*}
Given a program \( p \) to solve a query \( q \), our goal is to generate a set of unit tests \( \mathcal{T} \) comprising input images and expected answers, as shown in Figure \ref{fig:qualitative_examples}. This process involves three main steps: \textbf{Candidate Unit Test Generation} (Section~\ref{sec:candidates}), \textbf{Unit Test Sampling} (Section~\ref{sec:sampling}), and 
 \textbf{Image Generation} (Section~\ref{sec:image}).
\vspace{-.2cm}\subsubsection{Candidate Unit Test Generation $\psi$}\label{sec:candidates}\vspace{-.2cm}
As illustrated in Figure \ref{fig:framework_fig}, rather than generating images directly for unit tests, we first create image descriptions with expected answers. This approach reduces computational overhead during the preliminary stage of unit test coverage sampling, after which we generate images only for those tests that are included in the final unit test suite \(\mathcal{T}\).
In particular, we first generate a superset of \( M \) candidate unit tests using the unit test generator \( \psi \), which is implemented as an auto-regressive large language model. The unit test generator \( \psi \) can take both the query \( q \) and the program implementation \( p \) as inputs \(\mathcal{T}_{\text{cand}} = \psi(q, p) = \{ t_1, t_2, \ldots, t_M \}\).
Each candidate unit test \( t_i \) consists of an image caption \( c_i \) and an expected answer \( y_i\). We explore whether including the program implementation \( p \) provides useful signals for unit test generation (Section~\ref{sec:ablations}), despite conventional engineering practices that advocate for implementation-independent unit tests. This allows us to investigate whether this principle extends to visual unit testing.

\vspace{-.4cm}\subsubsection{Unit Test Coverage Sampling $\sigma$}\label{sec:sampling}\vspace{-.2cm}
Unit tests 
verify the behavior of code and should exhibit high \emph{isolation} and \emph{coverage}~\citep{khorikov2020unit}. In the context of visual programs, isolation is trivial since each program is a self-contained function. However, achieving high coverage---ensuring that the tests collectively exercise as much of the codebase as possible---is non-trivial due to the computational overhead of executing all candidate tests. To address this, we define coverage metrics tailored for visual programming unit tests, focusing on maximizing the diversity of both expected answers and visual inputs. The coverage sampler \( \sigma \) subsamples \( K \) pairs from \( \mathcal{T}_{\text{cand}} \), forming the subset \( \mathcal{T}_K \).\\
\noindent\textbf{Coverage by Answer:} We aim to include tests that cover all possible expected answers present in the candidate set. Let \(Y = \{ y_i \mid t_i \in \mathcal{T}_{\text{cand}} \}\) be the set of all expected answers in \( \mathcal{T}_{\text{cand}} \).
We define the \emph{answer diversity} criterion as ensuring that for every possible answer \( y \in Y \), there is at least one test \( t_i \in \mathcal{T}_K \) such that \( y_i = y \):
{\small \begin{align}\vspace{-.6cm}
\forall y \in Y, \quad \exists t_i \in \mathcal{T}_K \text{ such that } y_i \equiv y.\label{eq:answer_diversity}
\end{align}}

\noindent\textbf{Coverage by Input:} To maximize the diversity of visual inputs without generating all possible images, we operate on the image captions. We define an encoding function \( E \) that maps a caption \( c \) to a feature vector. We aim to maximize the \emph{input diversity} score \( \sigma_V(\mathcal{T}_K) \), defined as the maximum pairwise distance between the encoded captions:
{\small \begin{align}\vspace{-.4cm}
\sigma_V(\mathcal{T}_K) &= \max_{t_i, t_j \in \mathcal{T}_K,\, i \neq j} \left\| E(c_i) - E(c_j) \right\|\vspace{-.2cm}\label{eq:input_diversity}
\end{align}}This encourages the selection of tests with diverse descriptions, which in turn is likely to yield diverse images.\\
\noindent\textbf{Coverage by Answer then Input:} We begin by selecting one test for each possible answer to satisfy the answer diversity criterion (Equation~\ref{eq:answer_diversity}). Then, we iteratively select additional tests to maximize \( \sigma_V(\mathcal{T}_K) \) using the following criterion until \( K \) tests are selected, forming the subset \( \mathcal{T}_K \).
{\small \begin{align}\vspace{-.4cm}
t_{\text{new}} &= \arg\max_{t \in \mathcal{T}_{\text{cand}} \setminus \mathcal{T}_K} \max_{t' \in \mathcal{T}_K} \left\| E(c_t) - E(c_{t'}) \right\|.\vspace{-.2cm}\label{eq:sampling_criterion}
\end{align}}
\vspace{-.4cm}\subsubsection{Image Generation $M$}\label{sec:image}\vspace{-.2cm} For each selected unit test \( t_i = (c_i, y_i) \in \mathcal{T}_K \), we generate the corresponding image \( v_i \) using a text-to-image model \( M \) to yield the final unit-test suite \(\mathcal{T}=\{(M(c_i), y_i)\mid \forall t_i\in \mathcal{T}_K\}\). We employ three state-of-the-art diffusion models: SDv1.4~\citep{Rombach_2022_CVPR}, SDXL3~\citep{podell2024sdxl}, and LM Guided Diffusion~\citep{lian2024llmgrounded} which utilizes automatically generated templates with phrases and bounding boxes for spatial conditioning~\citep{lian2024llmgrounded}. To provide these additional signals, we prompt an LLM with in-context examples and the caption \( c_i \) to generate pairs of phrases and bounding boxes \( (ph_i, bb_i) \) to feed into the text-to-image model: \(v_i = M(c_i, (ph_i, bb_i))\).

\vspace{-.1cm}\subsection{Program Selection Based on Unit Test Scores}
\label{sec:unit_test_scoring}\vspace{-.2cm}
We select the program \( p^\ast \) that succeeds on most unit tests by Equation \ref{eq:optimal_program}, where the overall score \( S(p) \) is computed by an aggregator \(H\) over individual scores \(s_{t_i}=h(\hat{y_i}, y_i)\).\\

\noindent\textbf{Individual Unit Test Scorer $h$:}
For each program \( p \) and test \( t_i = (v_i, y_i) \in \mathcal{T}_K \), we execute \(p\) on 
\( v_i \) to obtain the predicted answer \( \hat{y}_i = \phi(p, v_i)\).
We define a scoring function \( h \) that assigns a score \( s_{t_i} \) based on the program's output:
{\small \begin{align}\vspace{-.4cm}
s_{t_i} &= h(\hat{y}_i, y_i) = \begin{cases}
    -\epsilon_r,& \text{if runtime error}, \\
    -\epsilon_c,& \text{if compilation error}, \\
    \mathbb{I}\{\hat{y}_i \equiv y_i\},& \text{otherwise}
\end{cases}\label{eq:scoring_function}
\end{align}
}where \( \epsilon_r \) and \( \epsilon_c \) are runtime and compilation error penalties and $\mathbb{I}$ is the indicator function.\\

\noindent\textbf{Score Aggregator $H$:}  The individual scores $s_{t_i}$ are aggregated to compute an overall score \(\small S(p)=H(\{ s_{t_i} \mid t_i \in \mathcal{T} \})\). Here, $H$ represents the averaging function. The program \( p^\ast \) with the highest score is selected as the best candidate approximating Equation \ref{eq:best_program} by:
{\small \begin{align}\vspace{-.6cm}
p^\ast &= \arg\max_{p \in \mathcal{P}} S(p).\vspace{-.2cm}\label{eq:optimal_program}
\end{align}}

\vspace{-.2cm}\subsection{Visual Unit Test Utilization Methods}\label{sec:method_applicatons}\vspace{-.2cm}

Figure~\ref{fig:applications} illustrates how to leverage visual unit tests in four ways, further elaborated below:

\noindent\textbf{Best Program Selection:}
\label{sec:applications_best_program_selection}
Given a set of candidate programs \( \mathcal{P} = \{ p_1, p_2, \ldots, p_N \} \) for a query \( q \), our goal is to select the program \( p^\ast \) that is most likely to produce the correct answer when executed on the visual input \( v \). We utilize the unit test scores \( S(p) \) computed for each program \( p \in \mathcal{P} \) as described in Section~\ref{sec:unit_test_scoring}. The best program--the program succeeds on most unit tests-- is selected by solving the optimization problem in Equation \ref{eq:optimal_program}.\\
\noindent\textbf{Answer Refusal:} If the maximum unit test score \( S(p^\ast) \) falls below a threshold \( \theta \), indicating low confidence in all candidate programs, we refuse to provide a programmatic answer. Instead, we retreat to an end-to-end fallback method (refer to supplement for details). Formally, the decision rule is: \(\text{If } S(p^\ast)< \theta,\text{ refuse to answer and redirect}\).
Otherwise, we proceed to execute the selected program \( p^\ast \) on the original visual input \( v \) to obtain the final answer \(\hat{y} = \phi(p^\ast, v)\).
The hyperparameter \( \theta \) balances a trade-off between attempting to answer with potentially incorrect programs and deferring to a more reliable but less interpretable method.\\
\noindent\textbf{Re-Prompting:} If all generated programs \( \mathcal{P} \) fail to meet the threshold \( \theta \) (i.e., \( \max_{p \in \mathcal{P}} S(p) < \theta \)), we employ a re-prompting strategy to generate better candidate programs using feedback from unit tests:
{\small \begin{align}\vspace{-.6cm}
    \mathcal{P}' &= \pi\left( x'(q) + \mathcal{F} \right)
    \vspace{-.4cm}\label{eq:reprompting_function}
\end{align}
}where: \( x'(q) \) is an adaptation of the original input containing the API, the query \( q \), and in-context examples of unit-test-feedback corrections, and \( \mathcal{F} \) is the feedback derived from unit test results~\footnote{~$\mathcal{F}$ comprises unit test image descriptions, expected answers, and the predicted answers generated by the program in the current iteration.}, summarizing the discrepancies between expected and actual outputs, and \( \pi \) is the program generator.

We select the best program \( p^{\ast\ast} \) from the new set \( \mathcal{P}' \) based on their unit test scores \(p^{\ast\ast} = \arg\max_{p' \in \mathcal{P}'} S(p')\).
If \( S(p^{\ast\ast}) \geq \theta \), we execute \( p^{\ast\ast} \) on the original visual input \( v \). Otherwise, we may repeat the re-prompting process until a predefined number of iterations is reached.\\
\noindent\textbf{Unsupervised Reinforcement Learning Reward Design}\label{sec:applications_rl} We propose to design RL rewards based on visual unit tests, aiming not only to provide extra supervision but also curtail policy deterioration due to logically incorrect programs~\citep{khan2024self}. The goal is to optimize a policy implemented as an autoregressive language model for program generation \( \pi_w \), parameterized by \( w \), by minimizing the reward-weighted loss over the dataset \( D \), where each example consists of an image \( v \), user query \( q \), generated program \( p \) by the previous iteration's policy \( \pi_{w^{\text{itr}-1}} \), and ground truth answer \( y \):
{\small \begin{align}\vspace{-.4cm}
    J(w) &= \mathbb{E}_{(v, q, p, y) \sim D} \left[ R(v, p, y) \, L_{\text{NLL}}(p, q; w) \right],
    \vspace{-.2cm}\label{eq:objective}
\end{align}
}where \(\small L_{\text{NLL}}(p, q; w)= - \sum_{l=1}^L \log \pi_w(p_l|p_{1:l-1}, x(q))\) is the negative log-likelihood loss on next token prediction and \(L\) is the sequence length .

\citet{khan2024self} introduce a correctness reward based on performance on the training set:
{\small \begin{align}\vspace{-.4cm}
   R_{\text{Correct}}(v, p, y) &= \begin{cases}
      1, & \text{if } \phi(p, v) \equiv y, \\
      0, & \text{otherwise}.
   \end{cases}
   \vspace{-.4cm}\label{eq:vi_rep_reward}
\end{align}}

However, this approach can lead to sparse rewards and may falsely reward programs that are right for incorrect reasons. \citet{khan2024self} address this issue through human corrections to stabilize training. Instead we reformulate the reward using feedback from the visual unit tests:
{\small \begin{align}\vspace{-.4cm}
   R_{\text{ViUnit}}(v, p) &= \begin{cases}
      1, & \text{if } S(p) \geq \theta, \\
      S(p), & \text{otherwise},
   \end{cases}
   \vspace{-.4cm}\label{eq:vi_unit_reward}
\end{align}
}where \( \theta \) is a passing threshold.
We terminate policy iteration on declining reward. Following earlier work~\cite{karwowskigoodhart}, we assume that an optimal policy will keep increasing an optimal reward function \( R^\ast \). Thus, when our proxy reward \( R \) declines (i.e., regret increases), there are theoretical guarantees that we are not far from the optimal policy that can be learned under \( R \).

\section{Experimental Setup}\label{sec:experiments}
\vspace{-.2cm}Below is the experimental setup: datasets (Section \ref{sec:data}), baselines (Section \ref{sec:baselines}), and implementation details (Section \ref{sec:implementation_details}).

\vspace{-.1cm}\subsection{Data}\label{sec:data}\vspace{-.2cm}
We utilize three compositional reasoning datasets: GQA~\citep{hudson2019gqa} for Visual Question Answering (VQA), SugarCREPE~\citep{hsieh2024sugarcrepe}, and  Winoground~\citep{thrush2022winoground} for Image-Text Matching (ITM), assessing model performance via accuracy metrics. For GQA, we calculate accuracy using an implementation by \citet{suris2023vipergpt}, which standardizes and compares generated answers for exact matches.\footnote{\href{https://github.com/cvlab-columbia/viper/blob/main/datasets/gqa.py}{https://github.com/cvlab-columbia/viper/blob/main/datasets/gqa.py}} Our experimental setup incorporates training and testing splits sampled similar to \citet{khan2024self}, specifically testing on 502 examples from the GQA \texttt{balanced-val} split and training on 1022 examples from the \texttt{balanced-train} split, with 10 samples per question \texttt{group}. In SugarCREPE, we utilize 788 examples for training by subsampling approximately 10\% of the dataset balanced across question types, excluding our validation split. The validation subset consists of 560 examples and includes both positive and negative image-text pairings from 40 samples from each of the 7 question types. The full Winoground dataset is used, encompassing all possible positive and negative pairings for a total of 1600 test examples, with the SugarCREPE dataset employed for training purposes. Refer to the supplement for further dataset details.

\begin{figure*}[ht]
    \centering
    \begin{subfigure}[b]{0.475\linewidth}
        \centering
        \includegraphics[width=\linewidth]{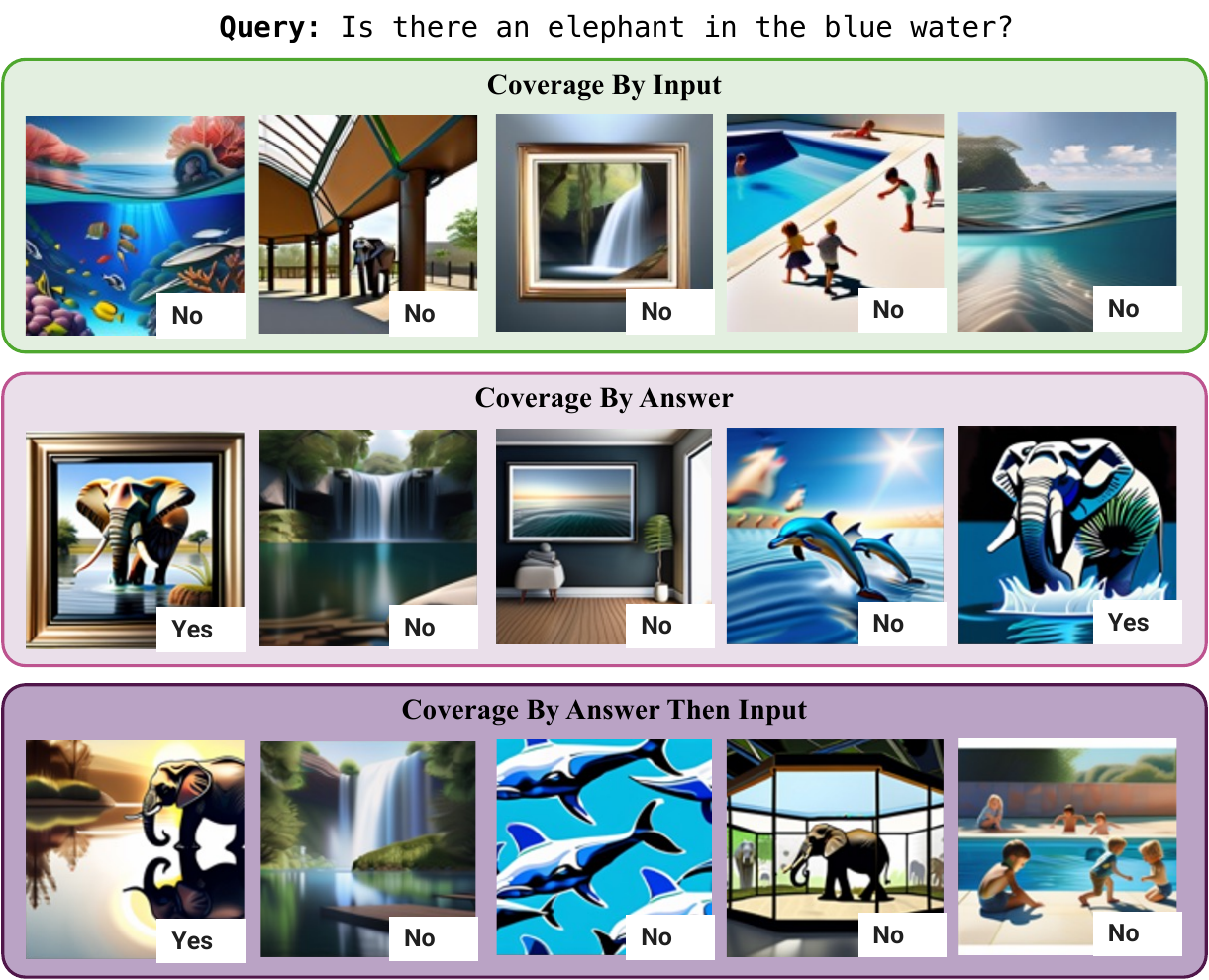}
        \caption{Unit Tests Generated by Different Sampling Methods}
        \label{fig:sampling_examples}
    \end{subfigure}
    \hfill
    \begin{subfigure}[b]{0.49\linewidth}
        \centering
        \includegraphics[width=\linewidth]{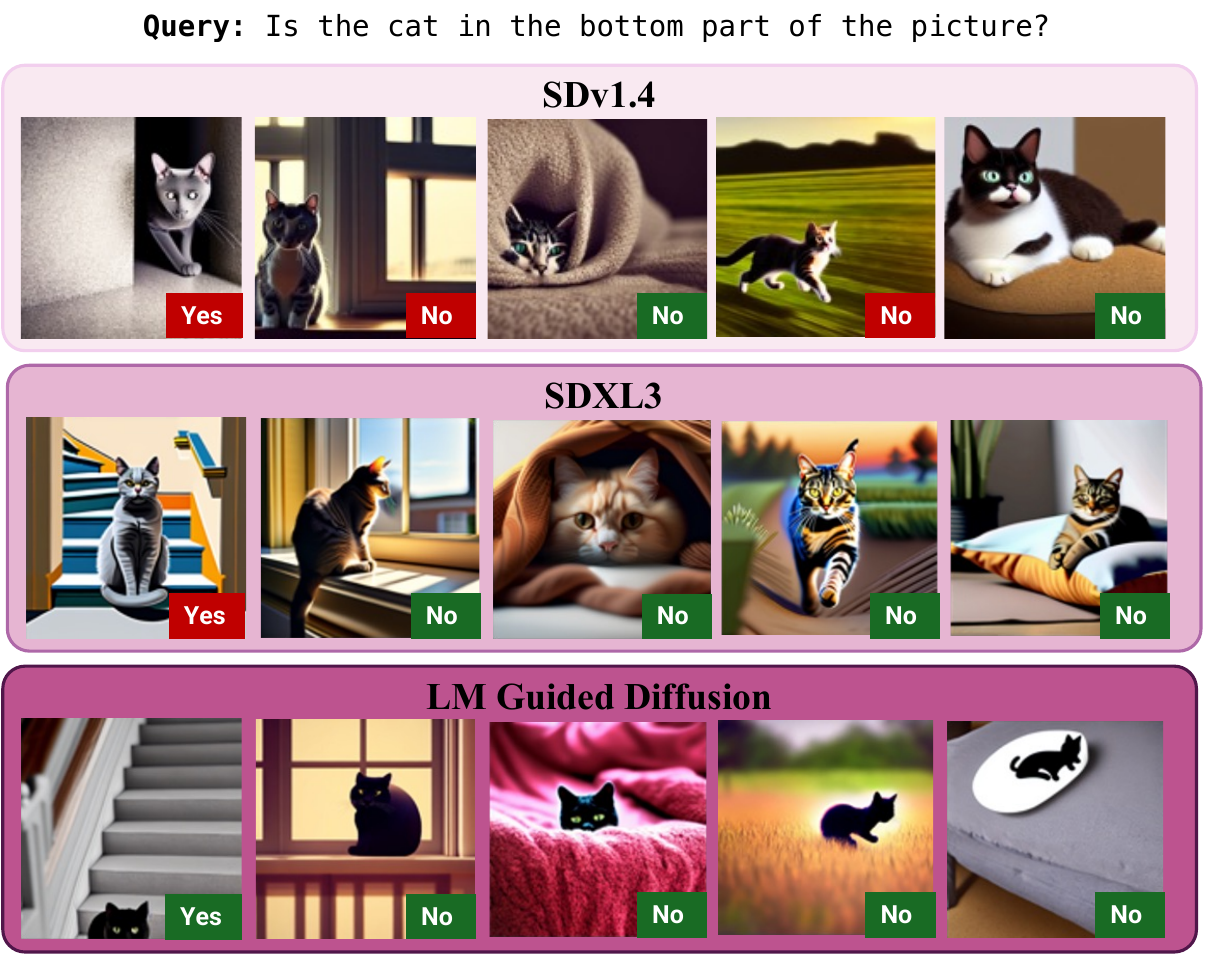}
        \caption{Unit Tests Generated by Different Diffusion Methods}
        \label{fig:diffusion_examples}
    \end{subfigure}
    \vspace{-.2cm}\caption{Comparison of Unit Tests Generated by Different Methods}
    \vspace{-5mm}
    \label{fig:combined_examples}
\end{figure*}
\vspace{-.1cm}\subsection{Baselines}\label{sec:baselines}\vspace{-.2cm}
We evaluate against the following baselines:~\\
\noindent\textbf{Base Setup:} Following the prototypical use of visual programs~\cite{suris2023vipergpt,gupta2023visual}, we prompt the LLM to generate a single program per query, which is executed to retrieve a response.\\
\noindent\textbf{Most Common Answer:} To leverage multiple programs, we compare performance with selecting the most common answer across executed programs if one exists.\\
\noindent\textbf{Error Re-prompting:} To evaluate the effectiveness of unit-test incorporation in program correction via unit-test re-prompting, we benchmark performance against a method that leverages error-traces as feedback \(\mathcal{F}\) in Equation \ref{eq:reprompting_function}. Further details are provided in the supplement.~\\
\noindent\textbf{Correctness Reward:} We baseline unsupervised unit-test RL reward fomulation against the supervised correctness reward described by Equation \ref{eq:vi_rep_reward}.

\vspace{-.1cm}\subsection{Implementation Details}\label{sec:implementation_details}\vspace{-.2cm}
We provide a summary of key implementation details, with additional information in the supplement. Experiments were conducted on two A100 40GB GPUs, though a single GPU suffices for smaller API models. Results report the mean and standard deviation across 3 runs.\\
\noindent\textbf{Program Generation Models:} Three program generator models are employed, \href{https://huggingface.co/codellama/CodeLlama-7b-hf}{codellama/CodeLlama-7b-Python-hf}~\cite{roziere2023code} and \href{https://huggingface.co/google/codegemma-7b-it}{google/codegemma-7b-it}~\cite{team2024codegemma} hosted on \href{https://huggingface.co}{Hugginface} and served by \href{https://github.com/vllm-project/vllm}{VLLM}~\citep{kwon2023efficient}, as well as \href{https://openai.com/index/gpt-4o-mini-advancing-cost-efficient-intelligence/}{gpt-4o-mini}~\cite{achiam2023gpt} served by \href{https://platform.openai.com/docs/overview}{OpenAI}.  We use HuggingFace's \href{https://huggingface.co/docs/trl/en/sft_trainer}{SFT-Trainer} to train the RL policy using LoRA~\citep{hu2022lora} with $\theta=0.8$ in Equation \ref{eq:vi_unit_reward}. Models are prompted with an API adapted from ViperGPT~\citep{suris2023vipergpt} and 4 in-context examples.\\
\noindent\textbf{API Models:} Object detection is performed using \href{https://huggingface.co/IDEA-Research/grounding-dino-base}{IDEA-Research/grounding-dino-base}~\citep{liu2023grounding}. For image-text matching, we use \href{https://huggingface.co/openai/clip-vit-large-patch14-336}{openai/clip-vit-large-patch14-336}~\citep{radford2021learning}, and for VQA answering, we employ \href{https://huggingface.co/Salesforce/blip2-flan-t5-xxl}{Salesforce/blip2-flan-t5-xxl}~\citep{li2023blip}. All models are accessed through \href{https://huggingface.co/}{HuggingFace}.\\
\noindent\textbf{Unit Test Generation Models:} We use \href{https://huggingface.co/meta-llama/Meta-Llama-3-8B-Instruct}{meta-llama/Meta-Llama-3-8B-Instruct}~\citep{dubey2024llama} to generate image descriptions and expected answers for unit test candidates. The unit test sampler is implemented with \href{https://sbert.net/}{sentence-transformers}, using the \href{https://huggingface.co/sentence-transformers/all-MiniLM-L6-v2}{all-MiniLM-L6-v2}~\citep{wang2020minilm} model to embed image descriptions. For image generation, we use the \href{https://huggingface.co/docs/diffusers/en/index}{diffusers} library, specifically \href{https://huggingface.co/CompVis/stable-diffusion-v1-4}{CompVis/stable-diffusion-v1-4} for SDv1.4, \href{https://huggingface.co/longlian/lmd_plus}{longlian/lmd\_plus} for LM Guided Diffusion, and \href{https://huggingface.co/stabilityai/stable-diffusion-xl-base-1.0}{stabilityai/stable-diffusion-xl-base-1.0} for SDXL3.~\\ 
\textbf{Program Scoring and Execution: } Program executions are capped at 120 seconds. Unit test scoring error penalties are set to $\epsilon_r =\epsilon_c=0.1$ (Equation \ref{eq:scoring_function}). Unless specified, no end-to-end model retreat was employed on exception.

\section{Strategies for Visual Unit Test Generation}\label{sec:ablations}\vspace{-.2cm}
We explore different unit test generation configurations applied on \textit{best program selection} using a smaller dataset of three questions from each group in GQA, and each tag in WinoGround, yielding 303 and 504 samples, respectively.
\begin{figure}[t]
    \centering
\includegraphics[width=\linewidth]{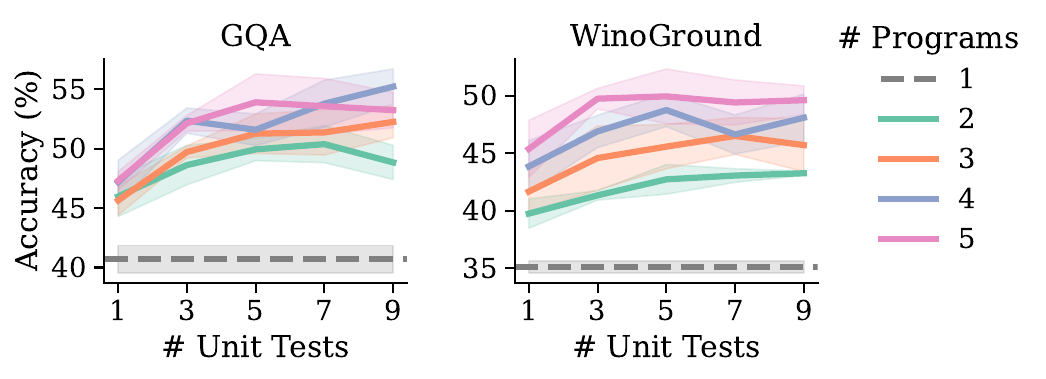}
\vspace{-.8cm}\caption{Accuracy across varying unit test and program counts.}
\vspace{-3mm}
    \label{fig:num_ut}
\end{figure}

\noindent\textbf{Number of unit tests $K$.}  Figure~\ref{fig:num_ut} illustrates that increasing both the number of unit tests and the number of candidate programs improves accuracy on both datasets. Accuracy rises substantially with the addition of unit tests, particularly from 1 to 5 tests, after which gains diminish. Higher numbers of programs (e.g., 4 or 5) consistently yield better accuracy compared to fewer programs, underscoring the benefit of exploring multiple candidate solutions. 
\begin{figure}[tb]
    \centering
        \includegraphics[width=\linewidth]{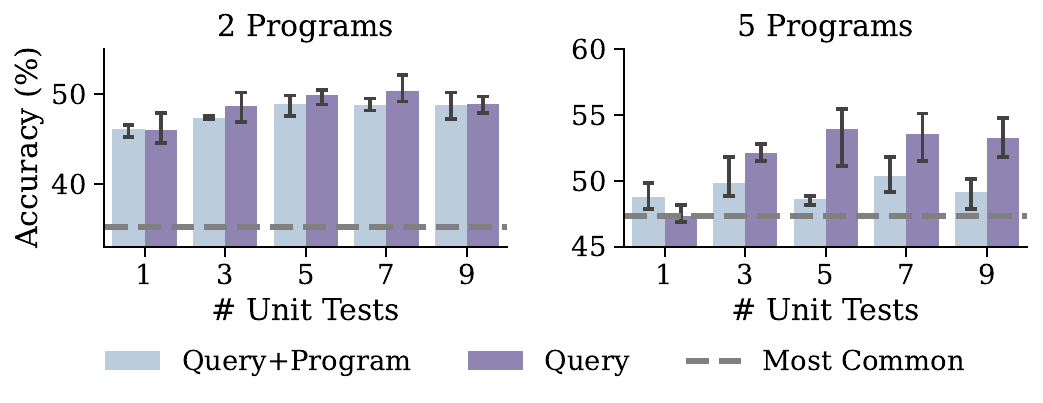}
    \vspace{-.8cm}\caption{Program in context in unit test generation for GQA.}
    \vspace{-3mm}
    \label{fig:prompting_ablation}
\end{figure}

\noindent\textbf{Unit Test Generator $\psi$}. Figure~\ref{fig:prompting_ablation} demonstrates that in low unit test settings, incorporating program information into unit test generation yields comparable results to query-only approaches. However, as the number of unit tests and programs increases, disregarding implementation details proves significantly more effective. This aligns with software engineering best practices, where unit tests are designed to remain independent of specific implementations.

\begin{figure}[tb]
    \centering
    \includegraphics[width=\linewidth]{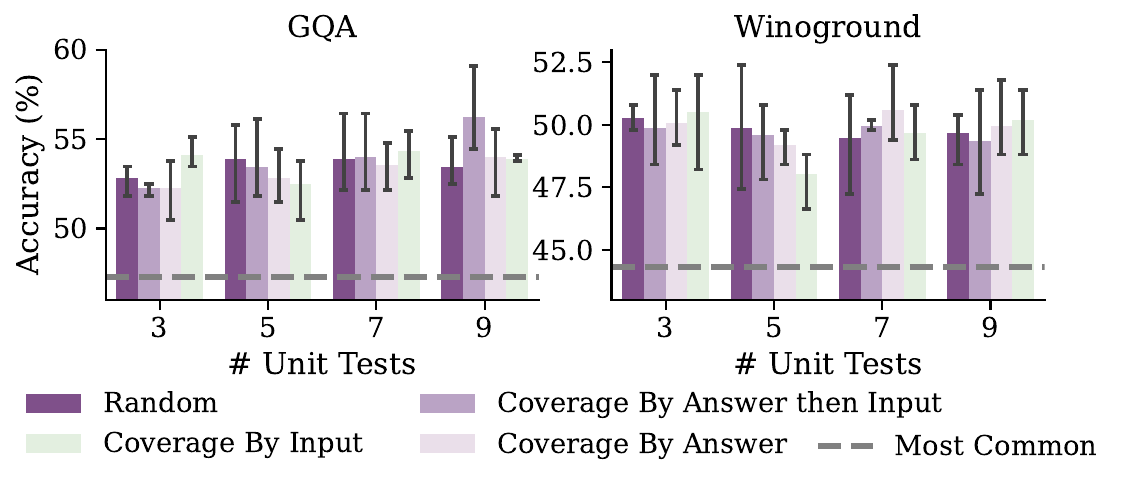}
    \vspace{-.8cm}\caption{Sampling method comparison at 5 programs.} 
    \vspace{-3mm}
    \label{fig:sampling_methods}
    \end{figure}
    
\noindent\textbf{Unit Test Sampler $\sigma$}.
Figure~\ref{fig:sampling_methods} demonstrates the impact of different unit test sampling methods on model accuracy. In GQA, ``Coverage By Answer then Input'' shows increasing performance as the number of unit tests grows, thus allowing the saturation of possible answers. Figure~\ref{fig:sampling_examples} highlights limitations of the other methods: ``Coverage by Input'' may suffer from reduced answer diversity, and ``Coverage by Answer'' could involve repetitive inputs. In WinoGround there is negligible difference across methods, due to its restriction to two answers, preventing significant sampling diversity. Nevertheless, an analysis of performance by question-type in the supplement shows that this sampling method yields higher results for attribute-related queries in both datasets. 

\begin{figure}[tb]
   \centering
       \includegraphics[width=\linewidth]{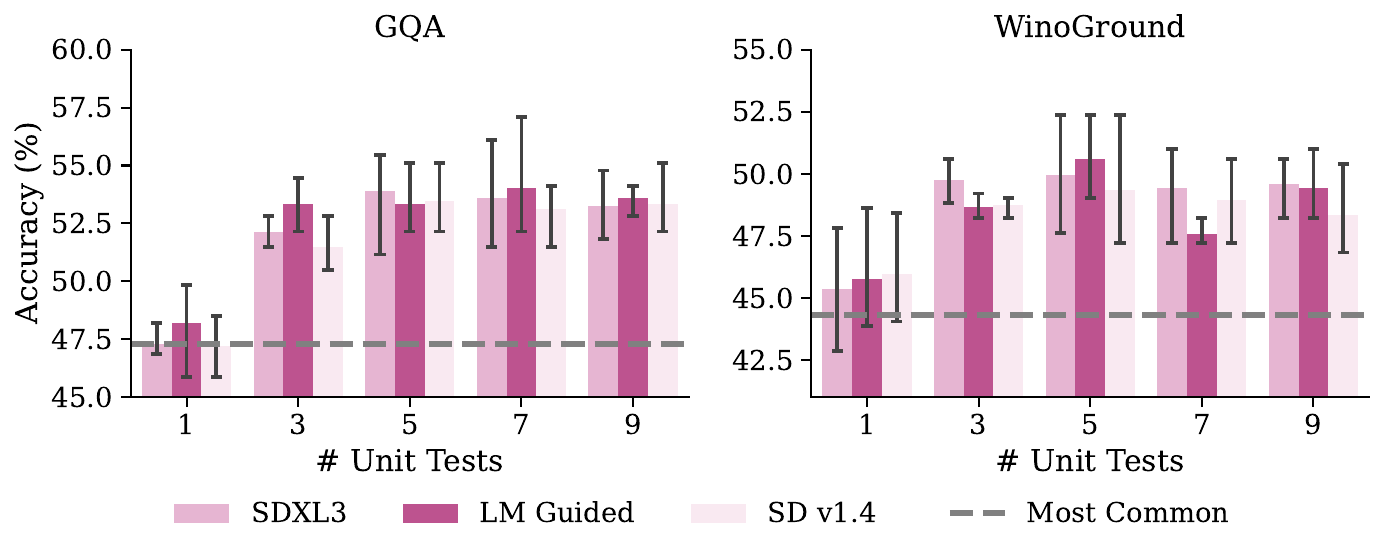}
   \vspace{-.8cm}\caption{Image generator comparison at 5 programs.}
   \vspace{-3mm}
    \label{fig:image_generator_ablation}
\end{figure}
\noindent\textbf{Image Generator $M$}. Figure~\ref{fig:image_generator_ablation} illustrates the impact of different diffusion models. In GQA at lower unit test settings LM Guided diffusion yields some accuracy improvements, while for WinoGround, LM Guided diffusion only helps in lower program settings, with quick convergence as the number of program increases. The benefit of LM Guided diffusion is primarily driven by improved tests when spatial positioning is critical as shown with the result breakdowns in the supplement and illustrated in Figure~\ref{fig:diffusion_examples}.
\begin{figure}
    \centering
\includegraphics[width=\linewidth]{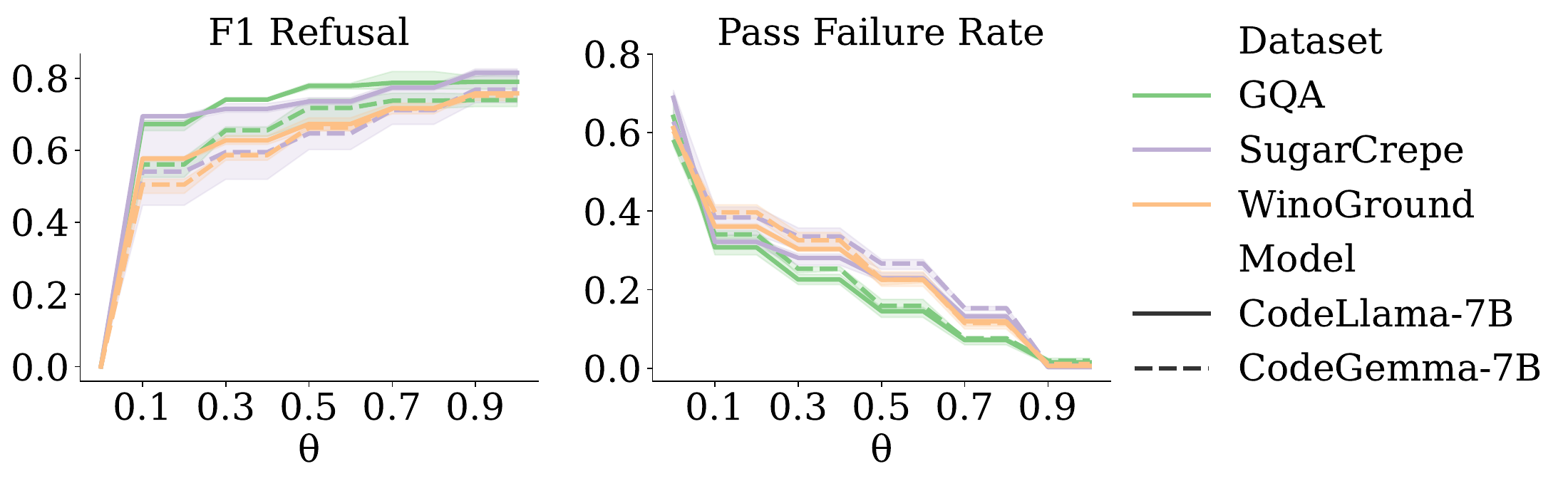}
  \vspace{-.8cm}\caption{Refusal evaluation at different passing thresholds.}
  \vspace{-3mm}
    \label{fig:refusal_plot}
\end{figure}

\noindent\textbf{Scoring function $h$}. The supplement presents results with varying error penalties, illustrating that in few unit test settings imposing error penalties enhances the likelihood of selecting a successful program.

\begin{table}[tb]
\resizebox{\columnwidth}{!}{
\begin{tabular}{@{}lcccllc@{}}
\toprule
     & & &\multicolumn{1}{c}{VQA}  & \multicolumn{2}{c}{Image-Text Matching} & \\ \midrule
LLM & \#~Prog & \#~UT &GQA           & Winoground         & SugarCREPE     & Avg.    \\ \midrule
\rowcolor{lightgray!30}\multicolumn{7}{c}{Base Setup}\\
gpt-4o-mini & 1 & 0 & 42.03$_{\pm1.21}$ &  44.98$_{\pm0.75}$ & 38.75$_{\pm0.47}$ & 41.92$_{\pm0.81}$\\

 CodeLlama-7B  & 1 & 0 & 35.99$_{\pm2.94}$ &  38.83$_{\pm0.45}$ & 30.54$_{\pm0.99}$ & 35.12$_{\pm1.46}$\\

 CodeGemma-7B  & 1 & 0 &   41.83$_{\pm2.26}$ &  39.60$_{\pm1.38}$ & 42.56$_{\pm1.52}$ & 41.33$_{\pm1.72}$\\

\rowcolor{lightgray!30}\multicolumn{7}{c}{Most Common Answer Setup}\\
CodeLlama-7B  &5 & 0 &42.50$_{\pm1.50}$& 45.85$_{\pm0.77}$  &41.67$_{\pm1.79}$ & 43.34$_{\pm1.35}$\\

CodeGemma-7B  &5 & 0 &43.89$_{\pm0.98}$& 46.04$_{\pm1.48}$  &46.67$_{\pm1.69}$ & 45.53$_{\pm1.38}$\\
\rowcolor{brilliantlavender!30}\multicolumn{7}{c}{ViUniT Setup (Ours)}\\
CodeLlama-7B & 5 & 5 &\textbf{49.27$_{\pm1.33}$}&\textbf{ 49.73$_{\pm0.73}$}  &\textbf{47.02$_{\pm1.19}$} & \textbf{48.67$_{\pm1.08}$} \\
CodeGemma-7B & 5 & 5 &\textbf{48.01$_{\pm1.05}$}& \textbf{51.92$_{\pm0.90}$}  &\textbf{51.85$_{\pm2.16}$} & \textbf{50.59$_{\pm1.37}$} \\

\bottomrule
\end{tabular}}
\vspace{-.3cm}\caption{Accuracy on Best Program Selection. \textbf{Bold} is best.}
\vspace{-3mm}
\label{tab:app_best_program}
\end{table}
\begin{table}[tb]
\resizebox{\columnwidth}{!}{
\begin{tabular}{@{}llccclc@{}}
\toprule
      & & &\multicolumn{1}{c}{VQA}  & \multicolumn{2}{c}{Image-Text Matching} \\ \midrule
 LLM & \#~Prog & \#~UT &GQA           & Winoground         & SugarCREPE  & Avg.       \\ \midrule
\rowcolor{lightgray!30}\multicolumn{7}{c}{Reverting on Error}\\
CodeLlama-7B  & 1 & 0 & 44.89$_{\pm2.04}$& \textbf{51.67$_{\pm1.16}$}  &\textbf{49.29$_{\pm0.99}$} & 48.61$_{\pm1.40}$\\

CodeGemma-7B  & 1 & 0 & 44.89$_{\pm2.19}$& 47.25$_{\pm2.17}$  &49.58$_{\pm0.88}$ &47.24$_{\pm1.74}$\\
\rowcolor{brilliantlavender!30}\multicolumn{7}{c}{Reverting on ViUniT Threshold $\theta=0.7$ (Ours)}\\
CodeLlama-7B & 1 & 5 &\textbf{54.18$_{\pm0.40}$}& 50.67$_{\pm1.28}$  &49.05$_{\pm0.82}$ & \textbf{51.30$_{\pm0.84}$} \\
CodeGemma-7B & 1 & 5 &\textbf{54.58$_{\pm1.24}$}& \textbf{50.73$_{\pm0.94}$}  &\textbf{50.12$_{\pm1.62}$}&\textbf{51.81$_{\pm1.27}$} \\
\bottomrule
\end{tabular}}
\vspace{-.3cm}\caption{Answer Refusal: Reverting to end-to-end model on error or unit test passing failure ($\theta=0.7$). \textbf{Bold} is best.}
\label{tab:app_refusal}
\end{table}
\begin{table}[]
\resizebox{\columnwidth}{!}{
\begin{tabular}{@{}llcccclc@{}}
\toprule
      & &&&\multicolumn{1}{c}{VQA}  & \multicolumn{2}{c}{Image-Text Matching} &\\ \midrule
 LLM & Iter. & \#~Prog & \#~UT &GQA           & Winoground         & SugarCREPE & Avg.        \\ \midrule
\rowcolor{lightgray!30}\multicolumn{8}{c}{Error Reprompting}\\
CodeLlama-7B & 1 & 1& 0 &37.92$_{\pm2.68}$& 42.46$_{\pm0.57}$  &33.21$_{\pm0.64}$&37.86$_{\pm1.30}$\\
CodeGemma-7B & 1 & 1& 0 &42.63$_{\pm2.42}$& 42.42$_{\pm1.91}$  &44.52$_{\pm1.05}$ &42.63$_{\pm2.42}$\\

\rowcolor{brilliantlavender!30}\multicolumn{8}{c}{ViUniT Reprompting $\theta = 0.7$ (Ours)}\\
CodeLlama-7B & 1 & 1& 5 &\textbf{46.68$_{\pm2.52}$}& \textbf{51.85$_{\pm0.40}$}  &\textbf{47.68$_{\pm2.17}$} &\textbf{48.74$_{\pm1.69}$}\\
CodeGemma-7B & 1 & 1& 5 &\textbf{45.75$_{\pm0.30}$}& \textbf{48.19$_{\pm2.28}$}  &\textbf{48.21$_{\pm1.12}$} & \textbf{47.38$_{\pm1.23}$} \\
\bottomrule
\end{tabular}}
\vspace{-.3cm}\caption{Accuracy of different re-prompting methods. \textbf{Bold} is best.}
\label{tab:performance_reprompting}
\end{table}
\begin{table}[tb]
\resizebox{\columnwidth}{!}{
\begin{tabular}{@{}lcccccc@{}}
\toprule
      & & &\multicolumn{1}{c}{VQA}  & \multicolumn{2}{c}{Image-Text Matching}& \\\midrule
LLM  &\#~Prog & \#~UT &GQA           & Winoground         & SugarCREPE   & Avg.     \\ \midrule
\rowcolor{lightgray!30}\multicolumn{7}{c}{\textbf{Supervised} Correctness Reward}\\
CodeLlama-7B & 1 & 0 & 39.18$_{\pm4.88}$&\textbf{48.65$_{\pm0.87}$}&39.58$_{\pm2.75}$ & 42.47$_{\pm2.83}$\\

CodeGemma-7B & 1 & 0  &43.03$_{\pm5.08}$&45.98$_{\pm2.64}$&46.31$_{\pm2.26}$ &45.11$_{\pm3.33}$  \\

\rowcolor{brilliantlavender!30}\multicolumn{7}{c}{\textbf{Unsupervised} ViUniT Reward (Ours)}\\

CodeLlama-7B & 1 & 0 &\textbf{40.57$_{\pm2.10}$}&46.52$_{\pm0.81}$&\textbf{41.85$_{\pm1.44}$} & \textbf{42.98$_{\pm1.45}$} \\
CodeGemma-7B &1 & 0 &  \textbf{45.68$_{\pm2.45}$}&\textbf{49.29$_{\pm0.43}$}&\textbf{46.55$_{\pm0.69}$}&\textbf{47.17$_{\pm1.19}$} \\

\bottomrule
\end{tabular}}
\vspace{-.2cm}
\caption{Comparison of RL with supervised correctness rewards versus unsupervised unit-test-based rewards. \textbf{Bold} is best.\vspace{-.4cm}}
\label{tab:performance_rl}
\end{table}
\vspace{-.1cm}\section{Strategies of Visual Unit Test Utilization}\label{sec:applications_results}\vspace{-.1cm}
\noindent\textbf{Best Program Selection:}
Table \ref{tab:app_best_program} underscores the efficacy of \viUnit~ selection in identifying the most optimal program. Our approach demonstrates a notable average improvement of 11.4 accuracy points over the base setup and a substantial 7.7-point average gain over the \texttt{gpt-4o-mini} configuration. Furthermore, it surpasses most common answer selection by an average margin of 5.2 points.

\noindent\textbf{Answer Refusal:}
Figure \ref{fig:refusal_plot} illustrates the impact of varying the threshold $\theta$ on the F1 score of refusing programs with incorrect answers (left), and the false pass failure rate (right), measured relative to the total number of programs. The minimal false pass failure rate at higher thresholds supports the use of unit test scores as a proxy for correctness during unsupervised model fine-tuning. Table \ref{tab:app_refusal} showcases an improvement of 3.6 points of reverting to a fixed model when \(S(p)<\theta=0.7\) compared to reverting only on error.  
For CodeLlama-7B, 
performance on image-text matching
is similar between the two methods, as some programs yield correct answers despite failing unit tests. Although such programs impact final performance, a human inspection of 40 samples revealed that 65\% were unreliable from the start. 

\noindent\textbf{Re-prompting:} Table \ref{tab:performance_reprompting} demonstrates that re-prompting with \viUnit~ achieves an average improvement of 7.5 points over error-based re-prompting, with a notable 10.9-point increase for CodeLlama-7B, which performs lower in the base setting.  The unit tests offer additional opportunities for refining the method’s initial response, as they go beyond error detection to assess program confidence, while also providing a measure of comparison between the programs.

\noindent\textbf{RL Reward Design:} The pattern of improvements is particularly interesting in the RL setting, where we find that \viUnit~ rewards outperform correctness rewards by an average of 1.3 points in accuracy despite not relying on the training labels. Additionally, we observe a notable reduction in the percentage of code leading to exceptions; errors decrease from 14.47\% to 11.76\% for CodeLlama and even more sharply from 11.73\% to 4.68\% for CodeGemma. These results indicate that heavily rewarding higher-quality code, as filtered through unit tests, encourages the development of a more robust and error-resistant policy.

\section{Human Evaluation}\label{sec:human_evaluation}\vspace{-.2cm}
We summarize key findings from two human evaluations that assess unit test quality and improvements in program reliability. Full details are available in the supplement.\\
\noindent\textbf{Unit Test Evaluation:} We randomly sampled 20 examples from each of three datasets, each corresponding to 5 unit tests, resulting in a total of 300 unit tests, each of which was judged by three annotators. Based on the majority annotator response, 75\% of unit tests per sample were correct. 
Annotators could optionally comment on errors, with ``Missing Object'' noted as the most frequent issue.\\
\noindent\textbf{Program Evaluation:} To measure the effectiveness of unit tests in enhancing program reliability, we evaluated 100 VQA  
 programs that correctly answered the queries both from the base and the unit-test best program selection setups. Two annotators with 3+ years of Python experience graded programs from 0 (Fully Correct) to 3 (Irrelevant). 
 Under the unit test setup, 86\% of programs were fully correct, compared to 77\% in the base setup. Additionally, only 5\% of programs were marked completely incorrect—with none deemed irrelevant—compared to 14\% and 4\%, respectively, in the base setup. Notably, the most common error type shifted from ``Incorrect Logic'' in the base setup to ``Missing Checks (e.g., list index out of range)'' in the unit-test setup. 

\section{Conclusion and Future Work}
\label{sec:conclusion}\vspace{-.2cm}
We introduce \viUnit, the first framework to automatically generate unit tests for verifying visual program correctness, addressing cases where programs may appear correct for the wrong reasons. Unit tests are leveraged in four ways: best program selection (+11.4 points over the base setup and +7.7 points over \texttt{gpt4o-mini}), answer refusal, re-prompting, and unsupervised RL reward design (+1.3 points over supervised rewards). Future directions include fine-grained test generation and broader task applications. By reinforcing logical correctness, \viUnit~ advances robustness and interpretability in visual programs.

{\small
\bibliographystyle{ieeenat_fullname}
\bibliography{_main}
}

\clearpage \appendix 
\section{Data}
The three compositional reasoning datasets used in this work are GQA~\citep{hudson2019gqa}, SugarCREPE~\citep{hsieh2024sugarcrepe}, and WinoGround~\citep{thrush2022winoground}. Table \ref{tab:data_samples} shows examples from each dataset, and table \ref{tab:data_statistics} summarizes the dataset statistics. For GQA validation we sample 5 questions from each of the 102 question groups from the \texttt{balanced-val} split with a total of 502 examples. For testing, we sample 10 questions per group from the \texttt{balanced-train} split yielding 1022 examples. Note that some groups such as \texttt{typeVerifyC}, \texttt{stateChoose}, and \texttt{companyVerify} do not have a sufficient amount of questions, so we sample the whole group. For SugarCREPE, we utilize 788 examples for training by subsampling 10\% of the dataset balanced across the 7 question types, excluding our validation split. This validation subset consists of 560 examples and includes both positive and negative image-text pairings from 40 samples from each of the 7 question types. The full Winoground dataset is used, encompassing all possible positive and negative pairings for a total of 1600 examples, with the SugarCREPE dataset employed for training.
\begin{table}[h]
\centering
\small
\resizebox{\linewidth}{!}{%
\begin{tabular}{>{\centering}m{2cm}>{\centering\arraybackslash}m{3.7cm}>{\centering\arraybackslash}m{1cm}}

\toprule
 \textbf{Image} & \centering\textbf{Question} & \textbf{Answer} \\
\midrule
\rowcolor{gray!10}\multicolumn{3}{c}{GQA}\\
\includegraphics[width=1.4cm]{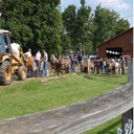}
&Are there any guys to the right of the brown horse?
&no \\
 \includegraphics[width=1.4cm]{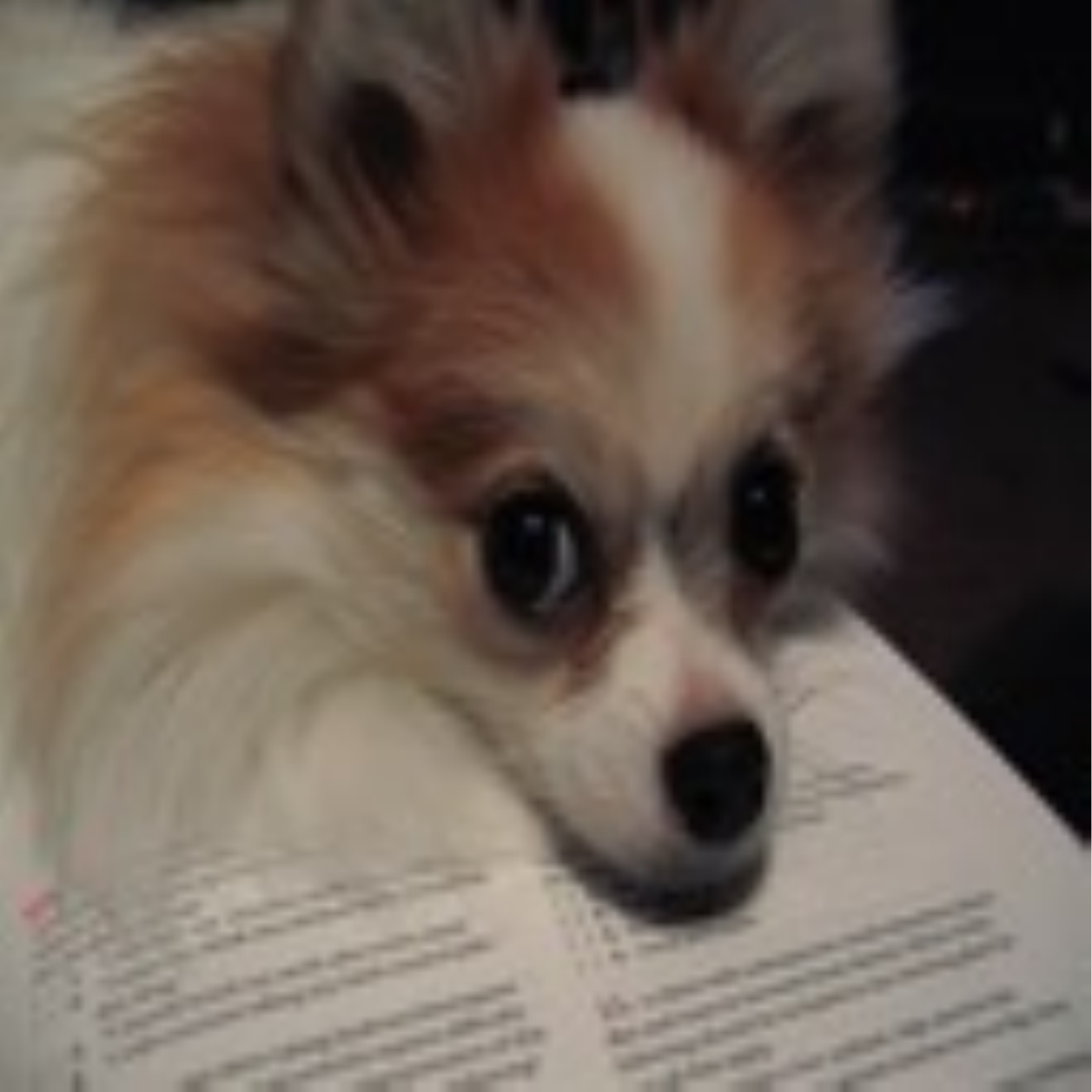}
& Which direction is the animal that looks white and brown looking at?
& forward \\
\includegraphics[width=1.4cm]{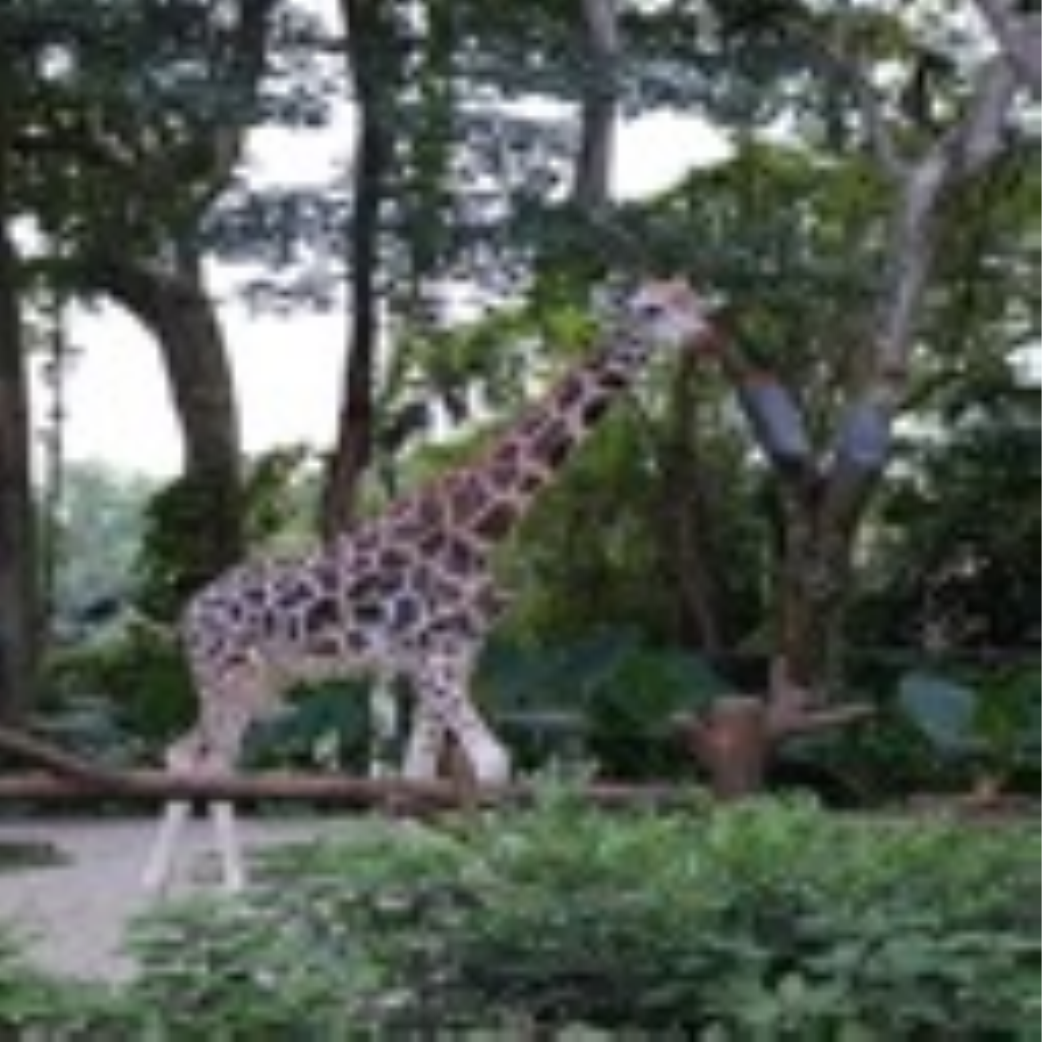}
& What type of animal is that fence behind of, an elephant or a giraffe?
& giraffe \\\midrule

\rowcolor{gray!10}\multicolumn{3}{c}{SugarCREPE}\\
    \includegraphics[width=1.4cm]{figs/sc_val_10.pdf}
&Is there a white pitcher holding flowers in a window sill?
&yes \\
\includegraphics[width=1.4cm]{figs/sc_val_81.pdf}
& Are a cat and a dog napping together under a blanket on the couch?
& no \\

    \includegraphics[width=1.4cm]{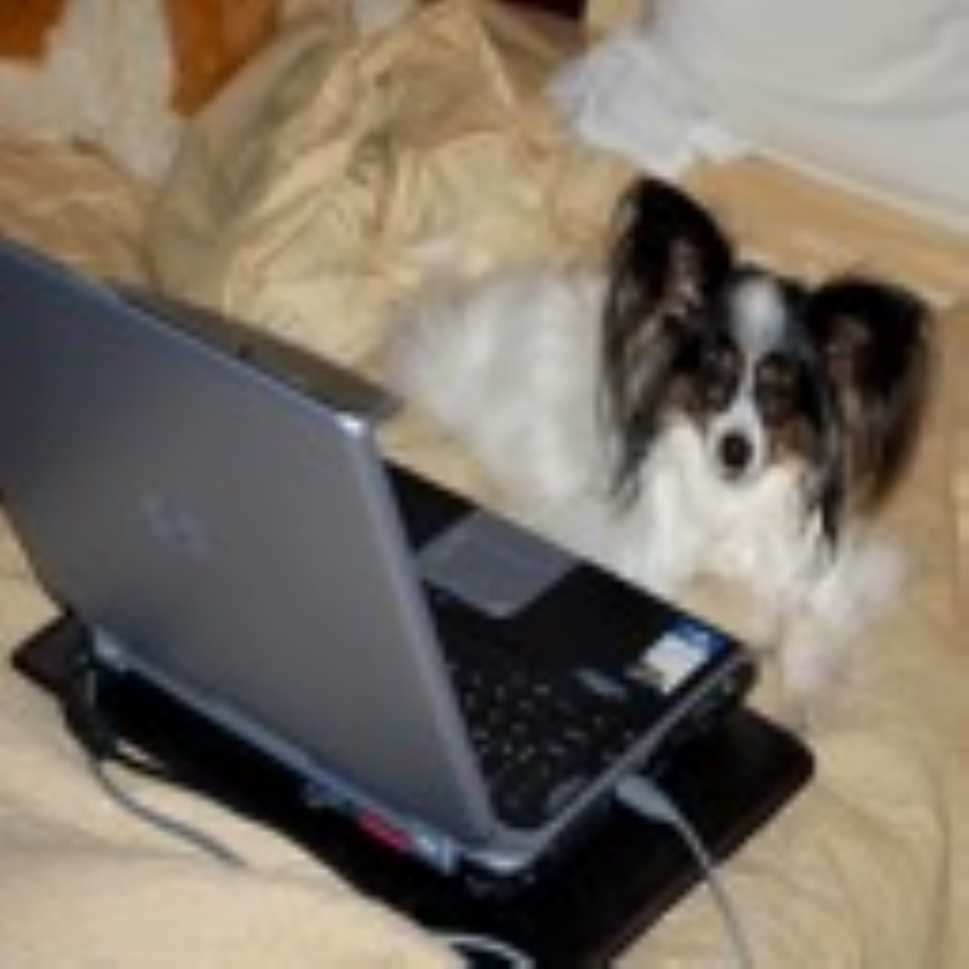}
& Is a dog sitting in front of a laptop on top of a bed?
& yes \\\midrule

\rowcolor{gray!10}\multicolumn{3}{c}{WinoGround}\\
    \includegraphics[width=1.4cm]{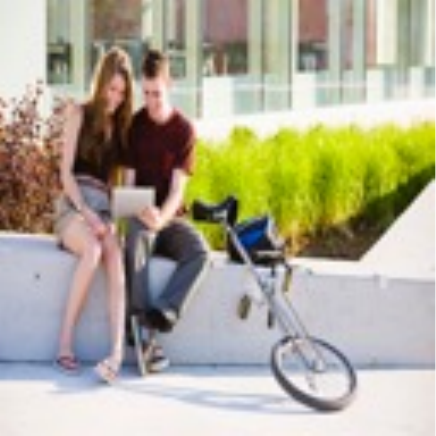}
& Verify image matches text=``two humans and one wheel''
&yes \\
  
    \includegraphics[width=1.4cm]{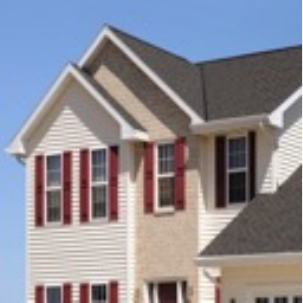}
& Verify image matches text=``red building with white shutters''
&no \\

    \includegraphics[width=1.4cm]{figs/gqa_val_469.pdf}
& Verify image matches text=``the person with the white collared shirt waters the plant while the other holds it''
&yes \\
\bottomrule
\end{tabular}}
\vspace{-.3cm}\caption{Dataset Samples}
\label{tab:data_samples}
\end{table}

\begin{table}[h]
\centering
\small

\resizebox{\linewidth}{!}{%
\begin{tabular}{ lccccc}
\toprule
 \textbf{\#~Samples} & \textbf{\#~Images}&\textbf{\#~Questions}&\textbf{\#~Answers} &\textbf{\#~Question Types} &\textbf{\#~Questions/Type} \\
\midrule
\rowcolor{gray!10}\multicolumn{6}{c}{GQA}\\
1022/502 & 1014/487 & 937/474 & 176/122 & 105/102 & 10/5\\
\rowcolor{gray!10}\multicolumn{6}{c}{WinoGround}\\
-/1600 & -/800 & -/800 & -/2 & -/70 & -/8\\
\rowcolor{gray!10}\multicolumn{6}{c}{SugarCREPE}\\
788/560 & 335/260 & 765/557 & 2/2 & 7/7 & 52/80\\

\bottomrule
\end{tabular}}
\vspace{-.3cm}\caption{Dataset Statistics: Values are shown in \{train/test\} format. For SugarCREPE and WinoGround, both positive and negative image-text pairings are included. In GQA, question types are divided by the data field \texttt{group}, and in WinoGround by the data field \texttt{tag}. The training data for WinoGround consists of SugarCREPE.}
\label{tab:data_statistics}
\end{table}

\section{Unit Test Sampling Pseudocode}
For clarity, Algorithm \ref{alg:sampling} presents the pseudocode for the unit test coverage sampling method described in Section \ref{sec:method}.
\begin{algorithm}
\caption{Unit Test Sampling Algorithm}
\label{alg:sampling}
\begin{algorithmic}[1] 
\Require \(T = \{t_1, t_2, \dots, t_n\}\), the set of texts
\Require \(A = \{a_1, a_2, \dots, a_m\}\), the set of answers
\Require \(f: T \rightarrow A\), a function mapping each text to an answer
\Require \(E(t)\), embedding function for text \(t\)
\Require \(k\), number of samples
\Require \(\texttt{\small use\_answers}\), a boolean flag
\Ensure \(S\), a subset of \(T\) of size \(k\)

\Function{SampleTexts}{T, A, f, E, k, \texttt{\small use\_answers}}
    \State Initialize \(S \gets \emptyset\)
    \If{\(\texttt{\small use\_answers} = \text{True}\)}
        \For{each \(a_i \in A\)}
            \State Select \(t\) from \(T\) such that \(f(t) = a_i\)
            \State \(S \gets S \cup \{t\}\)
            \State \(T \gets T \setminus \{t\}\)
        \EndFor
    \Else
        \State Select a random \(t\) from \(T\)
        \State \(S \gets \{t\}\)
        \State \(T \gets T \setminus \{t\}\)
    \EndIf
    \While{\(|S| < k\)}
        \State \(s_{\text{new}} \gets \arg\max_{t \in T} \max_{s \in S} \|E(t) - E(s)\|\)
        \State \(S \gets S \cup \{s_{\text{new}}\}\)
        \State \(T \gets T \setminus \{s_{\text{new}}\}\)
    \EndWhile
    \State \Return \(S\)
\EndFunction
\end{algorithmic}
\end{algorithm}


\section{Program Generation and Execution}
\label{app:program_generation}
In this section, we outline the implementation details for program generation and execution. 
\subsection{Generation Details}
\label{app:prog_gen_implementation}
For program generation we use in context examples both in of-the-shelf inference, and finetuned model inference. Generation is conducted using \href{https://github.com/vllm-project/vllm}{VLLM} with the following generation parameters: \texttt{temperature=1.0}, \texttt{top\_p=0.9}, \texttt{top\_k=0.0}, \texttt{max\_new\_tokens=320}, and \texttt{num\_beams=1}. We set the temperature at a high value to ensure diversity in generated programs. For CodeLLaMA we prefix the prompt with \texttt{<s>}, and for CodeGemma we enclose it in \texttt{<bos><start\_of\_turn>[..]<end\_of\_turn>}
\subsection{Image Patch API}
\label{app:imagepatch_api}
We present the \texttt{ImagePatch} API in Code \ref{code:api_prompt} which we adapt the  from \citet{khan2024self} which is in turn adapted from ViperGPT~\citet{suris2023vipergpt}. 
We implement object detection using \href{https://huggingface.co/IDEA-Research/grounding-dino-base}{IDEA-Research/grounding-dino-base}~\citep{liu2023grounding} with  \texttt{text\_threshold=box\_threshold=0.2}, image-text-matching using \href{https://huggingface.co/openai/clip-vit-large-patch14-336}{openai/clip-vit-large-patch14-336}~\citep{radford2021learning} using 0.8 similarity threshold for detection, and the underlying visual question answering module is \href{https://huggingface.co/Salesforce/blip2-flan-t5-xxl}{Salesforce/blip2-flan-t5-xxl}~\citep{li2023blip} loaded in 8-bits using \href{https://huggingface.co/docs/bitsandbytes/main/en/index}{BitsAndBytes} with a maximum batch size of 4 and generation hyperparameters 
\texttt{length\_penalty=-1}, \texttt{num\_beams=5}, \texttt{max\_length=10},\texttt{min\_length=1},\texttt{do\_sample=False}, \texttt{top\_p=0.9}, \texttt{repetition\_penalty=1.0}, and \texttt{temperature=1} for QA and set \texttt{length\_penalty=1} and \texttt{max\_length=30} for captioning. All models are served by \href{https://huggingface.co/}{HuggingFace}. 
\subsection{In-Context Examples}
\label{app:incontext_examples}
We present the in-context examples used for visual question answering and image-text matching in Codes \ref{code:vqa_ice} and \ref{code:itm_ice} respectively. Code execution is handled using multiprocessing with a batch size of 30, and a timeout of 120 seconds, after which a \texttt{TimeOutException} is raised if execution exceeds the limit.

\section{Unit Test Generation}
\label{app:ut_generation}
\subsection{Implementation Details}
\label{app:ut_gen_implementation}
To generate the unit test imaage descriptions and expected answers we prompt  \href{https://huggingface.co/meta-llama/Meta-Llama-3-8B-Instruct}{meta-llama/Meta-Llama-3-8B-Instruct}, executed via \href{https://github.com/vllm-project/vllm}{VLLM} with the following generation parameters: \texttt{temperature=0.7}, \texttt{top\_p=0.9}, \texttt{top\_k=0.0}, \texttt{max\_new\_tokens=512}, and \texttt{num\_beams=1}. We return 3 output sequences, from which we extract the unit tests, deduplicate them, and filter answers longer than five words since they are out of distribution to the task before feeding them to the sampling module. 

\subsection{In-Context Examples}
\label{app:in_context_examples}

We prompt the LLM with the system prompt presented below, as well as in-context examples presented in Codes \ref{code:ut_gen_ice_vqa} and \ref{code:ut_gen_ice_itm} for VQA and ITM respectively.
\begin{quote} 
\label{quote:ut_gen_system}
    \texttt{You are a skilled AI assistant specialized in generating test cases for programs that respond to queries about images.}
\end{quote}

\subsection{Unit Test Candidate Generation}
\label{app:ut_gen_candidates}
We experiment with two prompting methodologies for the unit test generation: \texttt{Query-Only} and \texttt{Query+Implementation}. The former only takes into account the user query to generate the unit-tests, while the latter takes into account also each generated program. We prompt the Visual Program Generator in the same way, but instead also include implementation examples, and the current implementation as shown in Code \ref{code:vqa_ice_implementation}.

\subsection{Image Generation}
\label{app:image_generation}
To generate the images we use the \href{https://huggingface.co/docs/diffusers/en/index}{diffusers} library, and prompt each of the models with generation hyperaparameters \texttt{guidance\_scale=16.0} and \texttt{num\_inference\_steps=50}. In the case of NSFW image generation, we update the seed by 1 and regenerate an image up to 10 times. Effectively, all unit tests have a corresponding image. We use the following implementations:  \href{https://huggingface.co/CompVis/stable-diffusion-v1-4}{CompVis/stable-diffusion-v1-4} for SDv1.4, \href{https://huggingface.co/longlian/lmd_plus}{longlian/lmd\_plus} for LM Guided Diffusion, and \href{https://huggingface.co/stabilityai/stable-diffusion-xl-base-1.0}{stabilityai/stable-diffusion-xl-base-1.0} for SDXL3. 

\subsubsection{LM Grounded Diffusion}
\label{app:lm_grounded_diffusion}
To generate the bounding boxes and phrases for LM Grounded Diffusion we prompt  \href{https://huggingface.co/meta-llama/Meta-Llama-3-8B-Instruct}{meta-llama/Meta-Llama-3-8B-Instruct}, executed via \href{https://github.com/vllm-project/vllm}{VLLM} with the following generation parameters: \texttt{temperature=1.0}, \texttt{top\_p=0.9}, \texttt{top\_k=0.0}, \texttt{max\_new\_tokens=320}, and \texttt{num\_beams=1}. We return 5 candidate sequences to collect multiple candidates since we notice that often the extracted phrases can be empty, leading to failure in image generation. We present the prompt and in-context examples used for this part in Code \ref{code:lm_grounded}. 

\section{Strategies for Visual Unit Test Generation}
\subsection{Unit Test Sampler $\sigma$}
\begin{figure}[h]
    \centering
    \includegraphics[width=\linewidth]{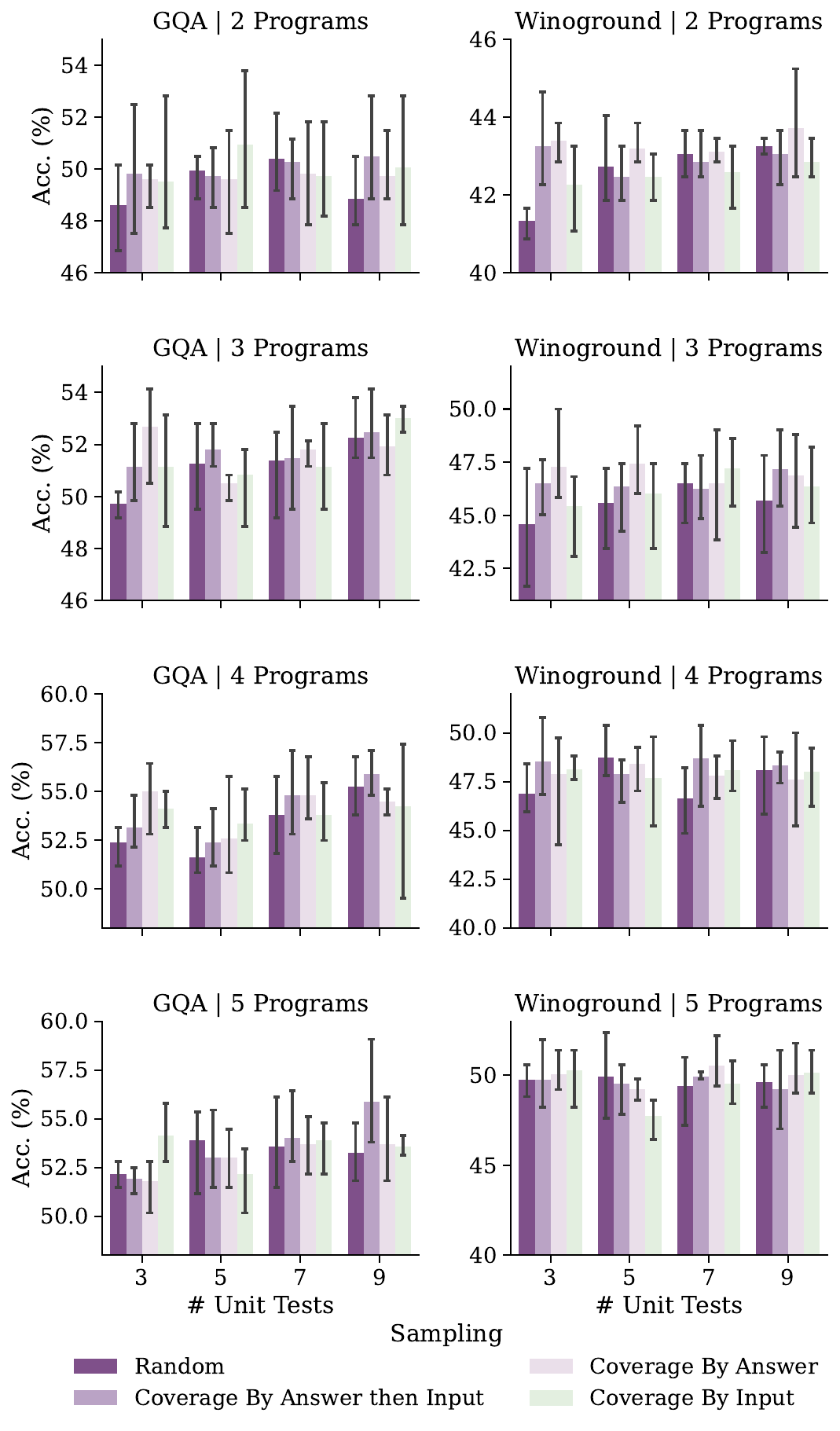}
    \vspace{-.3cm}\caption{Effect of sampling methods on performance across varying numbers of unit tests and program configurations.}
    \label{fig:gqa_winoground_sampling_all}
\end{figure}

\begin{figure}[h]
    \centering
    \includegraphics[width=\linewidth]{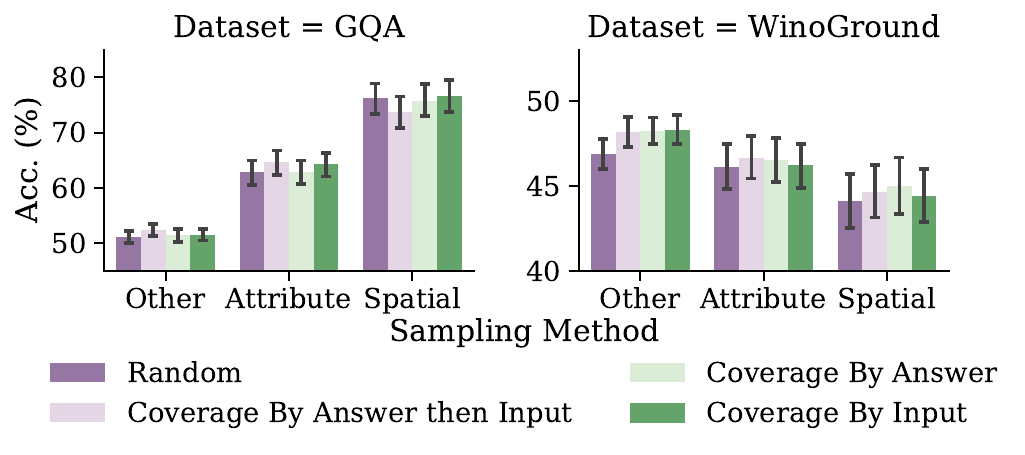}
    \vspace{-.3cm}\caption{Performance of sampling methods across question categories. Results are averaged over scenarios with at least five unit tests and three program configurations.}
    \label{fig:gqa_winoground_sampling_categories}
\end{figure}

Figure \ref{fig:gqa_winoground_sampling_all} illustrates the impact of different sampling strategies with varying the number of unit tests and program configurations. Our results indicate that `Coverage by Answer then Input', consistently outperforms other methods. To gain deeper insights, we categorize the questions into three groups: \texttt{Spatial}, \texttt{Attribute}, and \texttt{Other}. For GQA, we classify any question groups containing \texttt{Attr} as \texttt{Attribute} and those mentioning \texttt{location} or \texttt{position} as \texttt{Spatial}.
Figure \ref{fig:gqa_winoground_sampling_categories} presents the average performance across scenarios with at least five unit tests and three program configurations. Notably, the Coverage by Answer Then Input strategy emerges as the most effective for questions in the \texttt{Attribute} category.

\subsection{Image Generator $M$}
\begin{figure}[h]
    \centering
    \includegraphics[width=\linewidth]{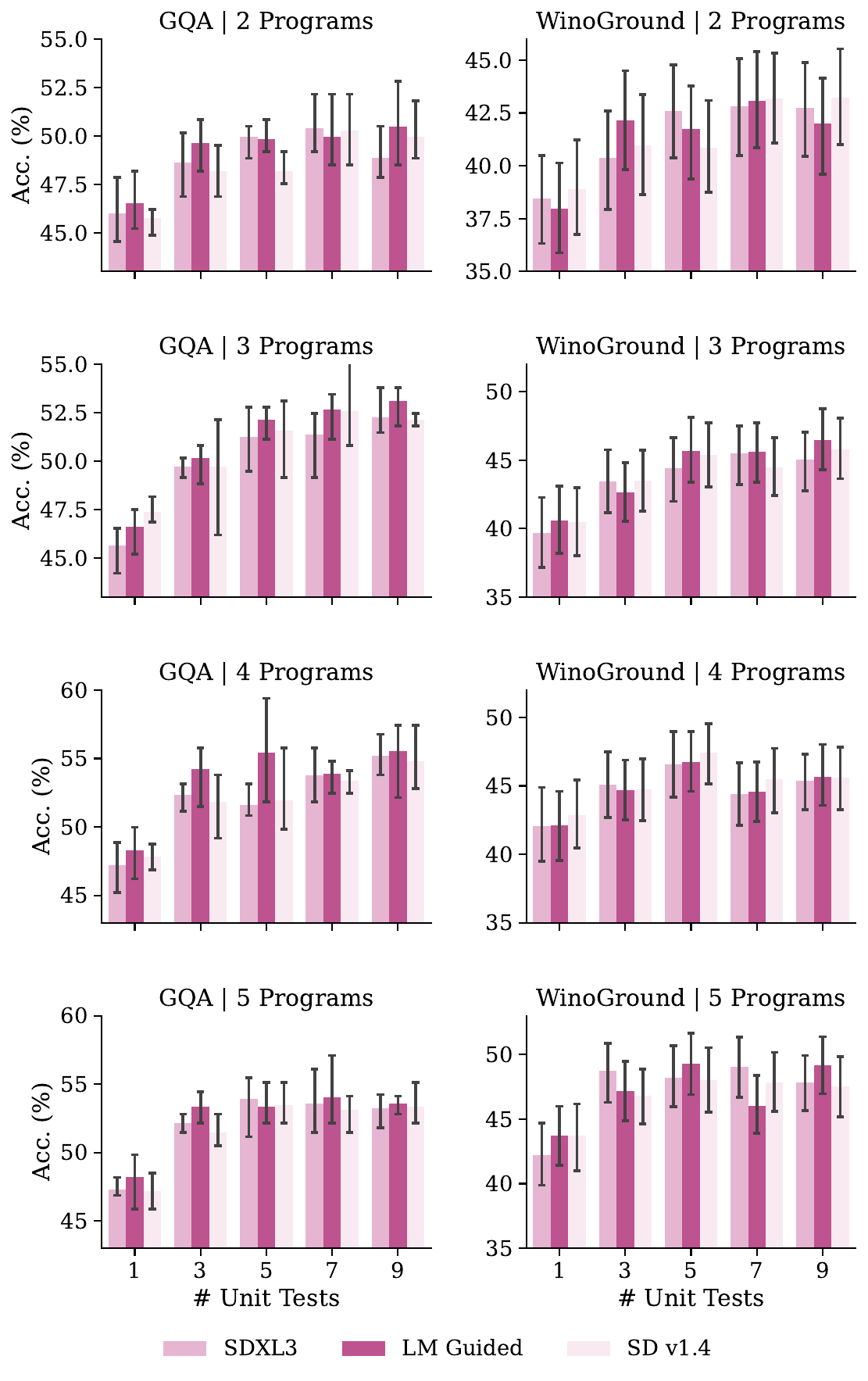}
    \vspace{-.3cm}\caption{Effect of diffusion model on performance across varying numbers of unit tests and program configurations.}
    \label{fig:diffusion_all}
\end{figure}

\begin{figure}[h]
    \centering
    \includegraphics[width=\linewidth]{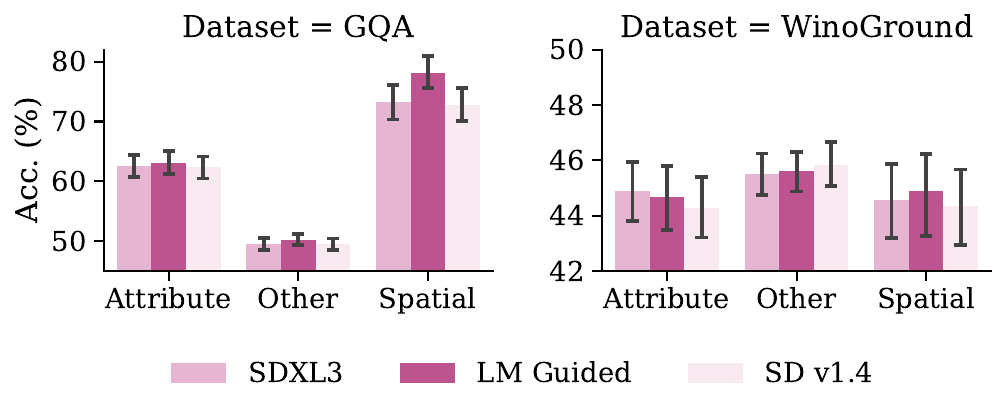}
    \vspace{-.3cm}\caption{Performance of different diffusion models across question categories. Results are averaged over scenarios with at least three unit tests and two program configurations.}
    \label{fig:gqa_winoground_diffusion_categories}
\end{figure}

Figure \ref{fig:diffusion_all} shows the impact of various diffusion models across different numbers of unit tests and program configurations. Our analysis reveals that LM-Guided diffusion consistently outperforms other methods, particularly in scenarios with more programs, where the likelihood of finding a suitable program for execution is higher. To gain deeper insights, figure \ref{fig:gqa_winoground_sampling_categories} presents the average performance across scenarios with at least three unit tests and two program configurations on the categories introduced in the previous subsection. To provide a deeper understanding, Figure \ref{fig:gqa_winoground_diffusion_categories} illustrates the average performance across scenarios involving at least three unit tests and two program configurations, focusing on the categories defined earlier. Notably, LM-Guided diffusion proves most effective for questions in the \texttt{Spatial} category, highlighting the advantages of more controllable generation in achieving higher spatial fidelity. 

\subsection{Scoring function $h$}
\begin{figure}
    \centering
    \includegraphics[width=\linewidth]{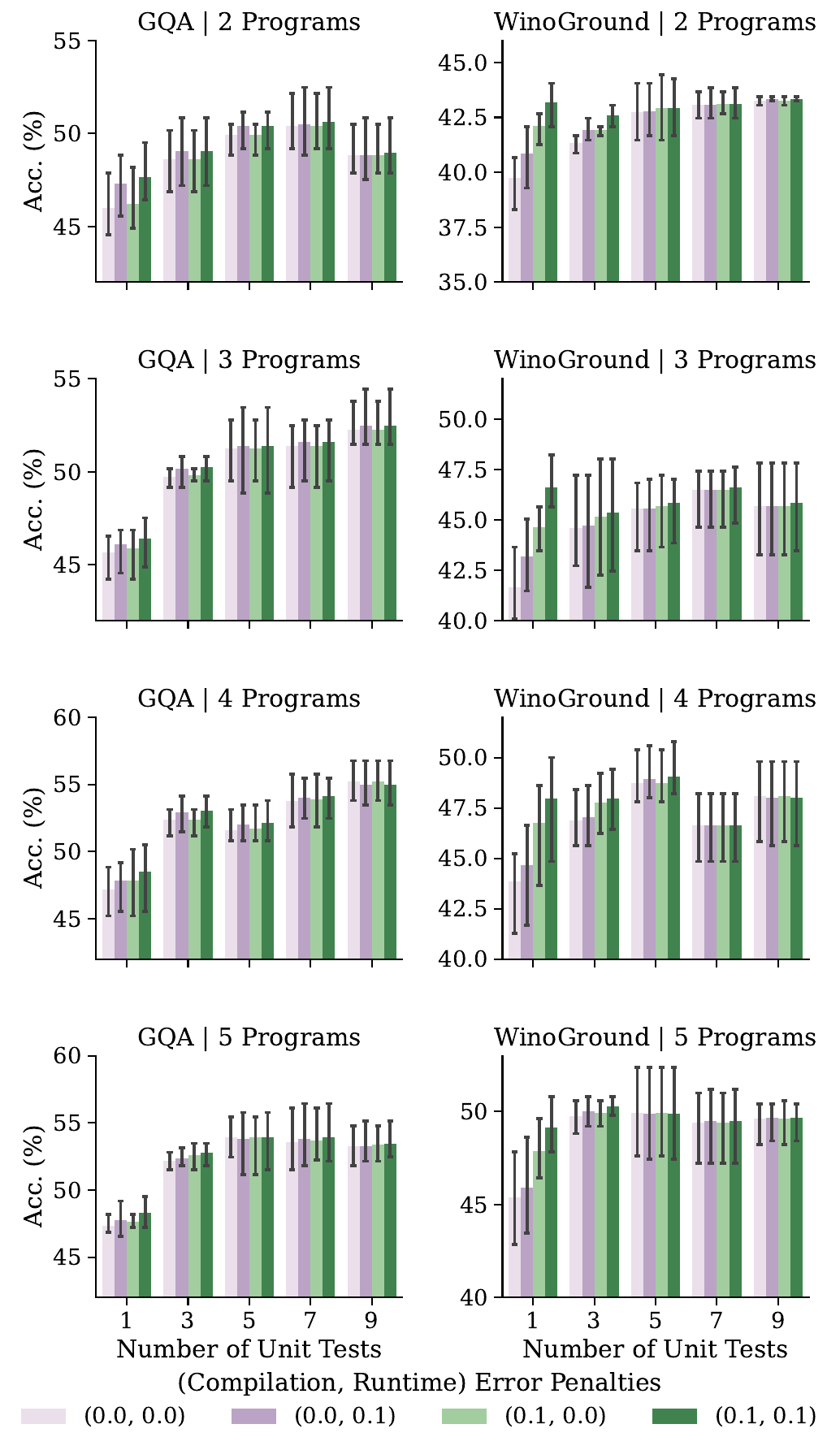}
    \vspace{-.3cm}\caption{Effect of error penalties on accuracy.}
    \label{fig:penalties_plot}
\end{figure}

Figure \ref{fig:penalties_plot} highlights the impact of error penalties across varying configurations of unit tests and programs. While their effect becomes negligible in higher-resource configurations with more programs and unit tests, error penalties prove beneficial in lower-resource settings. In these scenarios, they help prioritize the selection of executable programs, thereby improving performance. Notably, runtime error penalties are more impactful for GQA, whereas compilation error penalties play a larger role in WinoGround. This difference likely stems from the higher complexity of WinoGround programs, which are more prone to compilation errors.

\subsection{Aggregate Scorer $H$}
\begin{figure}
    \centering
    \includegraphics[width=\linewidth]{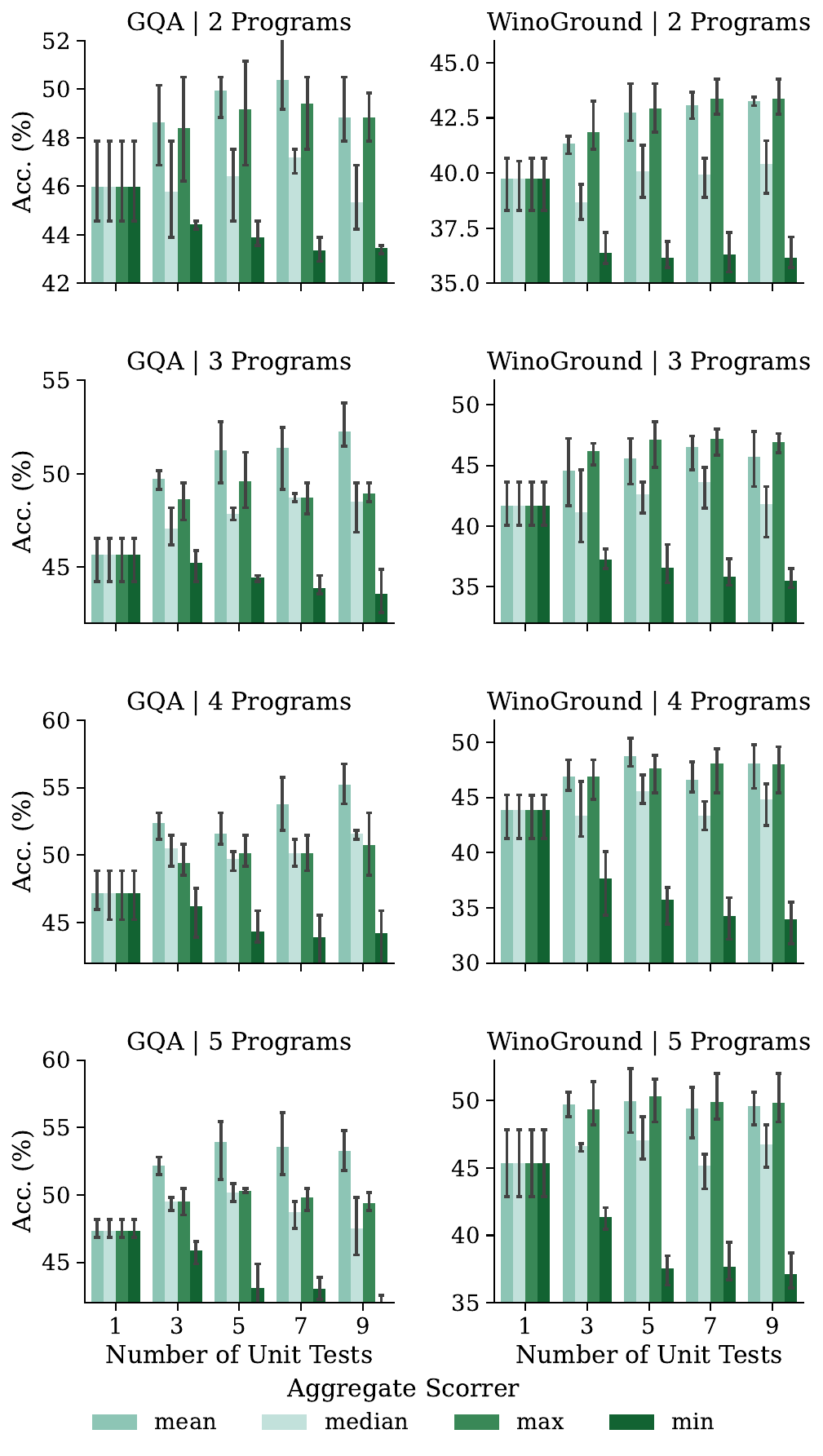}
     \vspace{-.3cm}\caption{Effect of aggregator function on accuracy.}
    \label{fig:aggregators_plot}
\end{figure}

Figure \ref{fig:aggregators_plot} illustrates the impact of various aggregator functions on accuracy. Among these, mean score aggregation consistently outperforms other methods, particularly in configurations with a higher number of programs. In the case of WinoGround, however, max aggregation also performs competitively, occasionally surpassing mean aggregation. This is likely due to the binary nature of the answers in WinoGround and the increased likelihood of selecting correct for incorrect reasons programs. 

\section{Visual Unit Test Utilization Methods}
\subsection{Best Program Selection}
Tab \ref{tab:app_best_program_supplement} shows additional results on best program selection with varrying number of programs. 

\begin{table}[h]
\resizebox{\columnwidth}{!}{
\begin{tabular}{@{}lcccllc@{}}
\toprule
     & & &\multicolumn{1}{c}{VQA}  & \multicolumn{2}{c}{Image-Text Matching} & \\ \midrule
LLM & \#~Prog & \#~UT &GQA           & Winoground         & SugarCREPE     & Avg.    \\ \midrule
\rowcolor{lightgray!30}\multicolumn{7}{c}{Base Setup}\\
gpt-4o-mini & 1 & 0 & 42.03$_{\pm1.21}$ &  44.98$_{\pm0.75}$ & 38.75$_{\pm0.47}$ & 41.92$_{\pm0.81}$\\

 CodeLlama-7B  & 1 & 0 & 35.99$_{\pm2.94}$ &  38.83$_{\pm0.45}$ & 30.54$_{\pm0.99}$ & 35.12$_{\pm1.46}$\\
 CodeGemma-7B  & 1 & 0 &   41.83$_{\pm2.26}$ &  39.60$_{\pm1.38}$ & 42.56$_{\pm1.52}$ & 41.33$_{\pm1.72}$\\

\rowcolor{lightgray!30}\multicolumn{7}{c}{Most Common Answer Setup}\\
CodeLlama-7B  &2 & 0 &27.76$_{\pm0.41}$& 36.19$_{\pm0.66}$  &32.02$_{\pm2.25}$ & 31.99$_{\pm1.11}$\\
CodeLlama-7B  &3 & 0 &35.99$_{\pm0.70}$& 42.40$_{\pm0.85}$  &37.26$_{\pm2.70}$ & 38.55$_{\pm1.42}$\\
CodeLlama-7B  &4 & 0 &38.71$_{\pm1.61}$& 42.12$_{\pm0.60}$  &39.17$_{\pm2.01}$ & 40.00$_{\pm1.41}$\\
CodeLlama-7B  &5 & 0 &42.50$_{\pm1.50}$& 45.85$_{\pm0.77}$  &41.67$_{\pm1.79}$ & 43.34$_{\pm1.35}$\\
CodeGemma-7B  &2 & 0 &31.87$_{\pm0.80}$& 33.04$_{\pm0.67}$  &36.37$_{\pm1.62}$ & 33.76$_{\pm1.03}$\\
CodeGemma-7B  &3 & 0 &40.31$_{\pm1.00}$& 40.50$_{\pm1.33}$  &44.58$_{\pm0.55}$ & 41.80$_{\pm0.96}$\\
CodeGemma-7B  &4 & 0 &40.44$_{\pm0.53}$& 43.06$_{\pm1.89}$  &44.46$_{\pm1.17}$ & 42.66$_{\pm1.20}$\\
CodeGemma-7B  &5 & 0 &43.89$_{\pm0.98}$& 46.04$_{\pm1.48}$  &46.67$_{\pm1.69}$ & 45.53$_{\pm1.38}$\\

\rowcolor{brilliantlavender!30}\multicolumn{7}{c}{ViUniT Setup (Ours)}\\
CodeLlama-7B & 2 & 5 &41.90$_{\pm1.74}$& 46.65$_{\pm1.63}$  &40.24$_{\pm0.82}$ & 42.93$_{\pm1.40}$\\
CodeLlama-7B & 3 & 5 &45.68$_{\pm0.94}$& 48.54$_{\pm0.37}$  &43.93$_{\pm1.09}$ & 46.05$_{\pm0.80}$\\
CodeLlama-7B & 4 & 5 &49.07$_{\pm2.39}$& 50.17$_{\pm0.54}$  &45.65$_{\pm1.22}$ & 48.30$_{\pm1.38}$\\
CodeLlama-7B & 5 & 5 &\textbf{49.27$_{\pm1.13}$} & 49.73$_{\pm0.73}$  &47.02$_{\pm1.19}$ & 48.67$_{\pm1.02}$\\

CodeGemma-7B & 2 & 5 &44.02$_{\pm0.72}$& 49.27$_{\pm0.57}$  &46.73$_{\pm2.30}$ & 46.67$_{\pm1.20}$\\
CodeGemma-7B & 3 & 5 &46.08$_{\pm0.41}$& 51.17$_{\pm1.98}$  &48.93$_{\pm1.86}$ & 48.73$_{\pm1.42}$\\
CodeGemma-7B & 4 & 5 &47.88$_{\pm1.36}$& \textbf{52.25$_{\pm1.35}$}  &50.83$_{\pm1.32}$ & 50.32$_{\pm1.34}$\\
CodeGemma-7B & 5 & 5 &48.01$_{\pm1.05}$& 51.92$_{\pm0.90}$  &\textbf{51.85$_{\pm2.16}$} & \textbf{50.59$_{\pm1.37}$} \\
\bottomrule
\end{tabular}}
\vspace{-.3cm}\caption{Accuracy on Best Program Selection with varying number of programs. \textbf{Bold} is best.}
\label{tab:app_best_program_supplement}
\end{table}

\subsection{Answer Refusal}
\begin{figure}
    \centering
    \includegraphics[width=\linewidth]{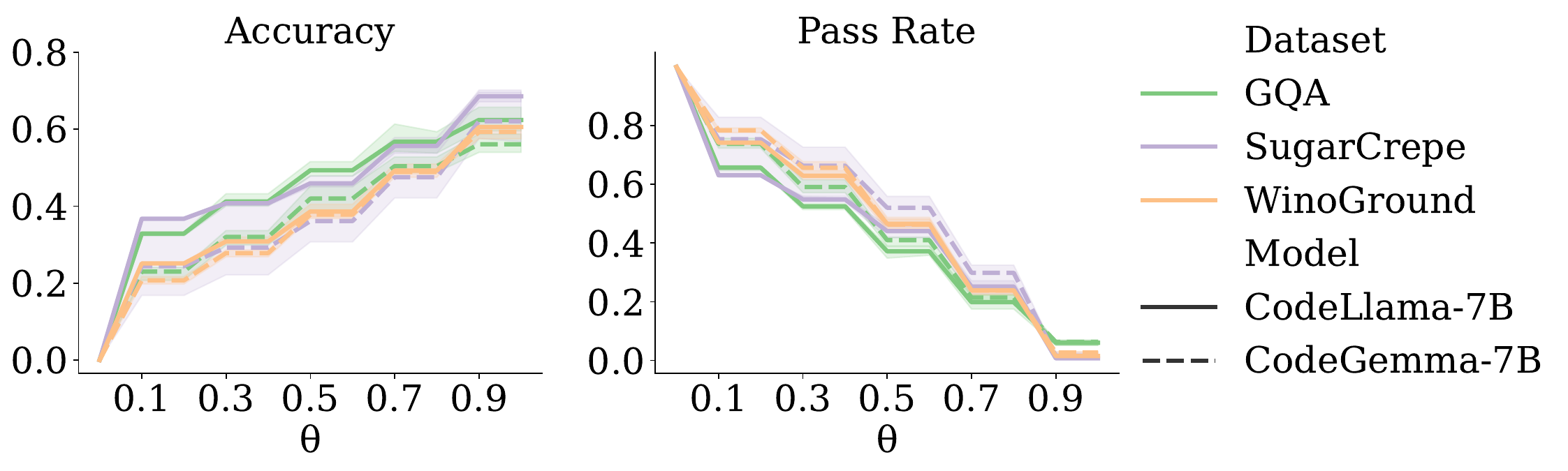}
    \caption{Accuracy and Program Pass Rate for different thereshold values for answer refusal.}
    \label{fig:refusal_plot_appendix}
\end{figure}

Figure \ref{fig:refusal_plot_appendix} shows additional statistics on answer refusal, in particular the accuracy of selecting programs that will provide the final answer and the programs that succeed on the unit tests at different thresholds. 

\subsection{Re-prompting}
\subsubsection{Implementation Details}
We consider an application of the unit tests to generate different candidate programs if the generated program falls below a threshold. To do so, we maintain the same hyperparameters in the program generator, but adapt the prompt to include the outputs of the unit tests as well as use suitable in context examples as shown in Codes \ref{code:viunit_reprompting_vqa}  and \ref{code:viunit_reprompting_itm} for VQA and ITM respectively.

\noindent\textbf{Error Reprompting Baseline}
We employ the same model and hyperparamters as the \viUnit reprompting, but instead adapt the prompt to take into account the error messages instead of the unit tests as shown in Codes \ref{code:error_reprompting_vqa} and \ref{code:error_reprompting_itm} for VQA and ITM respectively.

\subsubsection{Additional Results}
Table \ref{tab:performance_reprompting_appendix} presents the results of an additional reprompting iteration, highlighting that while \viUnit~ continues to achieve higher performance overall, there is a slight drop in accuracy compared to the previous iteration. This decline can be attributed to its attempts to refine programs that may already produce correct answers for the wrong reasons. Such corrections can inadvertently cause shifts in the generated answers, leading to decreased accuracy despite the method's focus on improving program fidelity.

\begin{table}[h]
\resizebox{\columnwidth}{!}{
\begin{tabular}{@{}llcccclc@{}}
\toprule
      & &&&\multicolumn{1}{c}{VQA}  & \multicolumn{2}{c}{Image-Text Matching} &\\ \midrule
 LLM & Iter. & \#~Prog & \#~UT &GQA           & Winoground         & SugarCREPE & Avg.        \\ \midrule
 
\rowcolor{lightgray!30}\multicolumn{8}{c}{Base Setup (Iteration = 0)}\\
 CodeLlama-7B  &0& 1 & 0 & 35.99$_{\pm2.94}$ &  38.83$_{\pm0.45}$ & 30.54$_{\pm0.99}$ & 35.12$_{\pm1.46}$\\

 CodeGemma-7B  &0& 1 & 0 &   41.83$_{\pm2.26}$ &  39.60$_{\pm1.38}$ & 42.56$_{\pm1.52}$ & 41.33$_{\pm1.72}$\\
\rowcolor{lightgray!30}\multicolumn{8}{c}{Error Reprompting}\\
CodeLlama-7B & 1 & 1& 0 &37.92$_{\pm2.68}$& 42.46$_{\pm0.57}$  &33.21$_{\pm0.64}$&37.86$_{\pm1.30}$\\
CodeLlama-7B & 2 & 1& 0 &38.78$_{\pm2.22}$& 44.58$_{\pm0.44}$  &37.08$_{\pm1.08}$&40.15$_{\pm1.25}$\\
CodeGemma-7B & 1 & 1& 0 &42.63$_{\pm2.42}$& 42.42$_{\pm1.91}$  &44.52$_{\pm1.05}$ &42.63$_{\pm2.42}$\\
CodeGemma-7B & 2 & 1& 0 &42.90$_{\pm2.65}$& 43.08$_{\pm1.73}$  &45.30$_{\pm0.92}$ &42.90$_{\pm2.65}$\\

\rowcolor{brilliantlavender!30}\multicolumn{8}{c}{ViUniT Reprompting $\theta = 0.7$ (Ours)}\\
CodeLlama-7B & 1 & 1& 5 &\textbf{46.68$_{\pm2.52}$}& \textbf{51.85$_{\pm0.40}$}  &\textbf{47.68$_{\pm2.17}$} &\textbf{48.74$_{\pm1.69}$} \\
CodeLlama-7B & 2 & 1& 5 &\textbf{46.95$_{\pm1.33}$}& \textbf{52.04$_{\pm0.83}$}  &\textbf{48.04$_{\pm1.64}$} &\textbf{49.01$_{\pm1.26}$} \\
CodeGemma-7B & 1 & 1& 5 &\textbf{45.75$_{\pm0.30}$}& \textbf{48.19$_{\pm2.28}$}  &\textbf{48.21$_{\pm1.12}$} & \textbf{47.38$_{\pm1.23}$} \\
CodeGemma-7B & 2 & 1& 5 &\textbf{44.42$_{\pm1.00}$}& \textbf{49.25$_{\pm2.66}$} &\textbf{48.81$_{\pm1.19}$ }& \textbf{47.49$_{\pm1.62}$} \\
\bottomrule
\end{tabular}}
\vspace{-.3cm}\caption{Accuracy of different re-prompting methods with an additional iteration. \textbf{Bold} is best.}
\label{tab:performance_reprompting_appendix}
\end{table}

\subsection{Reward Design for Reinforcement Learning}
\subsubsection{Implementation Details}
Table \ref{tab:rl_hyperparameters} contains additional hyperparameters used for training. Each RL epoch requires about 30 minutes with correctness reward, and 90 minutes with \viUnit~ reward since it requires execution of unit tests.   
\begin{table}[h]
\fontsize{8}{8}\selectfont
\centering
\begin{tabular}{@{}ll@{}}
\toprule
\textbf{Parameter}           & \textbf{Value}               \\ \midrule

\texttt{warmup\_ratio}        & 0.1                         \\
\texttt{max\_grad\_norm}      & 0.3                         \\
\texttt{lr\_scheduler\_type}  & \texttt{linear}             \\
\texttt{learning\_rate}       & 2e-4                        \\
\texttt{lora\_config.r}       & 16                          \\
\texttt{lora\_config.lora\_alpha} & 32                     \\
\texttt{lora\_config.lora\_dropout} & 0.05                 \\
\texttt{lora\_config.bias}    & \texttt{none}               \\
\texttt{lora\_config.target\_modules} & 
\begin{tabular}[c]{@{}ll@{}}
\texttt{k\_proj} &
\texttt{v\_proj} \\
\texttt{q\_proj} &
\texttt{o\_proj}
\end{tabular}\\\bottomrule
\end{tabular}
\caption{RL training hyperparameters.}
\label{tab:rl_hyperparameters}
\end{table}
\subsubsection{Additional Analysis}
Table \ref{tab:error_rate_rl} highlights the reduced error rates—measured as the number of programs leading to exceptions—achieved using the \viUnit~ reward. Additionally, Table \ref{tab:generalization_rl} presents the results of cross-task and cross-dataset generalization on policies trained with GQA, following the approach of \citep{khan2024self}. For VQAv2~\cite{goyal2017making}, we sample 10 questions for each of the 50 most common answers from the validation split of the compositional subset curated by \citep{selvaraju2020squinting}, similar to \citep{khan2024self}. For OKVQA~\cite{marino2019ok}, we sample 10 questions per question type, resulting in a total of 110 questions. The results indicate that while both reward types demonstrate strong generalization across tasks and datasets, the \viUnit~ reward consistently delivers superior performance.

\begin{table}[h]
\resizebox{\columnwidth}{!}{
\begin{tabular}{@{}lcccccc@{}}
\toprule
      & & &\multicolumn{1}{c}{VQA}  & \multicolumn{2}{c}{Image-Text Matching}& \\\midrule
LLM  &\#~Prog & \#~UT &GQA           & Winoground         & SugarCREPE   & Avg.     \\ \midrule
\rowcolor{lightgray!30}\multicolumn{7}{c}{\textbf{Supervised} Correctness Reward}\\
CodeLlama-7B & 1 & 0 &15.14$_{\pm7.74}$&\textbf{8.21$_{\pm1.72}$}&20.06$_{\pm3.62}$& 14.47$_{\pm4.36}$\\
CodeGemma-7B & 1 & 0 &9.10$_{\pm9.35}$&13.25$_{\pm6.30}$ &12.86$_{\pm4.41}$&11.73$_{\pm6.69}$\\

\rowcolor{brilliantlavender!30}\multicolumn{7}{c}{\textbf{Unsupervised} ViUniT Reward (Ours)}\\

CodeLlama-7B & 1 & 0 & \textbf{9.56$_{\pm2.13}$}&10.31$_{\pm1.55}$&\textbf{15.42$_{\pm3.03}$} & \textbf{11.76$_{\pm2.24}$} \\
CodeGemma-7B &1 & 0 &  \textbf{1.99$_{\pm0.91}$}&\textbf{5.81$_{\pm0.49}$} &\textbf{6.25$_{\pm1.02}$} &\textbf{4.68$_{\pm0.80}$} \\

\bottomrule
\end{tabular}}
\vspace{-.2cm}
\caption{Comparison of \textit{Error Rates} in models trained with supervised correctness rewards versus unsupervised unit-test-based rewards. Lower is better. \textbf{Bold} is best.\vspace{-.4cm}}
\label{tab:error_rate_rl}
\end{table}

\begin{table}[h]
\resizebox{\columnwidth}{!}{
\begin{tabular}{@{}lcccccc@{}}
\toprule
      & & &\multicolumn{2}{c}{X-Dataset Generalization}  & \multicolumn{2}{c}{X-Task Generalization}\\\midrule
LLM  &\#~Prog & \#~UT &VQAv2 & OK-VQA  & Winoground         & SugarCREPE       \\ \midrule
\rowcolor{lightgray!30}\multicolumn{7}{c}{Base Setup}\\
 CodeLlama-7B  &1 & 0 &25.67$_{\pm2.20}$ &16.09$_{\pm2.02}$   & 30.54$_{\pm0.99}$ & 35.12$_{\pm1.46}$\\

 CodeGemma-7B  &1 & 0 & 36.40$_{\pm1.44}$  &  27.58$_{\pm 2.48}$ & 42.56$_{\pm1.52}$ & 41.33$_{\pm1.72}$\\
\rowcolor{lightgray!30}\multicolumn{7}{c}{\textbf{Supervised} Correctness Reward}\\
CodeLlama-7B & 1 & 0 & 34.33$_{\pm7.82}$&24.12$_{\pm5.98}$ &41.02$_{\pm3.05}$& 37.14$_{\pm6.48}$ \\
CodeGemma-7B & 1 & 0 &42.47$_{\pm6.03}$ &28.12$_{\pm6.20}$&47.98$_{\pm4.98}$  &39.94$_{\pm11.58}$\\

\rowcolor{brilliantlavender!30}\multicolumn{7}{c}{\textbf{Unsupervised} ViUniT Reward (Ours)}\\

CodeLlama-7B & 1 & 0 &\textbf{35.87$_{\pm2.31}$}  & \textbf{25.64$_{\pm 0.91}$}&\textbf{43.63$_{\pm2.89}$}& \textbf{44.35$_{\pm3.18}$} \\
CodeGemma-7B &1 & 0 & \textbf{44.00$_{\pm4.20}$}  & \textbf{36.85$_{\pm 3.48}$}&\textbf{51.78$_{\pm0.41}$ } &\textbf{49.23$_{\pm2.54}$} \\

\bottomrule
\end{tabular}}
\vspace{-.2cm}
\caption{GQA policy generalization across tasks and datasets\vspace{-.4cm}}
\label{tab:generalization_rl}
\end{table}

\section{End-to-End Fallback Methods}
\label{app:end2end}
\subsection{Implementation Details}
\subsubsection{VQA}
\label{app:end2end_vqa}
For VQA we revert to ask the query directly to \href{https://huggingface.co/Salesforce/blip2-flan-t5-xxl}{Salesforce/blip2-flan-t5-xxl}~\citep{li2023blip} loaded in 8-bits using \href{https://huggingface.co/docs/bitsandbytes/main/en/index}{BitsAndBytes} with a maximum batch size of 4 and generation hyperparameters 
\texttt{length\_penalty=-1}, \texttt{num\_beams=5}, \texttt{max\_length=10},\texttt{min\_length=1},\texttt{do\_sample=False}, \texttt{top\_p=0.9}, \texttt{repetition\_penalty=1.0}, and \texttt{temperature=1}.
\subsubsection{Image-Text-Matching}
\label{app:end2end_itm}
 For image-text-matching we revert to  \href{https://huggingface.co/openai/clip-vit-large-patch14-336}{openai/clip-vit-large-patch14-336}~\citep{radford2021learning} using 0.8 similarity threshold for positive match, and negative otherwise.

\subsection{Results with Fallback Method on Exception}
In this work, we report results without employing a fallback method on exceptions, treating such cases as failures to better assess the quality of programs generated by different methods. However, it is common in the literature to report accuracy with a fallback method applied on exceptions. In Table \ref{tab:app_best_program_with_fallback} we present the best program selection results using this fallback approach on error.
\begin{table}[h]
\resizebox{\columnwidth}{!}{
\begin{tabular}{@{}lcccllc@{}}
\toprule
     & & &\multicolumn{1}{c}{VQA}  & \multicolumn{2}{c}{Image-Text Matching} & \\ \midrule
LLM & \#~Prog & \#~UT &GQA           & Winoground         & SugarCREPE     & Avg.    \\ \midrule
\rowcolor{lightgray!30}\multicolumn{7}{c}{Base Setup}\\
gpt-4o-mini\textdagger & 1 & 0 & 43.76$_{\pm1.72}$ &  \textbf{51.94$_{\pm0.56}$} & 49.46$_{\pm1.25}$ & 48.39$_{\pm1.17}$\\
CodeLlama-7B\textdagger  & 1 & 0 & 44.75$_{\pm2.01}$ &  51.65$_{\pm1.09}$ & 48.57$_{\pm0.82}$ &48.32$_{\pm1.31}$\\
CodeGemma-7B\textdagger  & 1 & 0 & 44.82$_{\pm2.30}$ &  47.23$_{\pm2.26}$ & 50.18$_{\pm0.71}$ & 47.41$_{\pm1.76}$\\
\rowcolor{lightgray!30}\multicolumn{7}{c}{Most Common Answer Setup}\\
CodeLlama-7B\textdagger  &5 & 0 &49.07$_{\pm2.79}$& 51.29$_{\pm0.87}$ &46.79$_{\pm1.29}$ &49.05$_{\pm1.65}$\\
CodeGemma-7B\textdagger  &5 & 0 &46.61$_{\pm1.24}$& 49.10$_{\pm1.32}$  &49.17$_{\pm1.52}$ & 48.29$_{\pm1.36}$\\
\rowcolor{brilliantlavender!30}\multicolumn{7}{c}{ViUniT Setup (Ours)}\\
CodeLlama-7B\textdagger  &5 & 5 &\textbf{49.27$_{\pm1.33}$}& 49.73$_{\pm0.73}$  &47.02$_{\pm1.19}$ & \textbf{48.67$_{\pm1.08}$} \\
CodeGemma-7B\textdagger  &5 & 5 &\textbf{48.14$_{\pm1.02}$}& 51.92$_{\pm0.90}$ &\textbf{51.85$_{\pm2.16}$} & \textbf{50.63$_{\pm1.36}$} \\
\bottomrule
\end{tabular}}
\vspace{-.3cm}\caption{Accuracy on Best Program Selection using fallback method on exception (indicated by \textdagger). \textbf{Bold} is best.}
\label{tab:app_best_program_with_fallback}
\end{table}

\section{Human Evaluation}\label{app:human_eval}
This section presents details on the human evaluations on the quality of unit tests, and program correctness. We used \href{https://www.google.com/forms/about/}{Google-Forms} to conduct the evaluations. 
\subsection{Unit Test Evaluation} 
\begin{figure}[h]
    \centering
    \includegraphics[width=\linewidth]{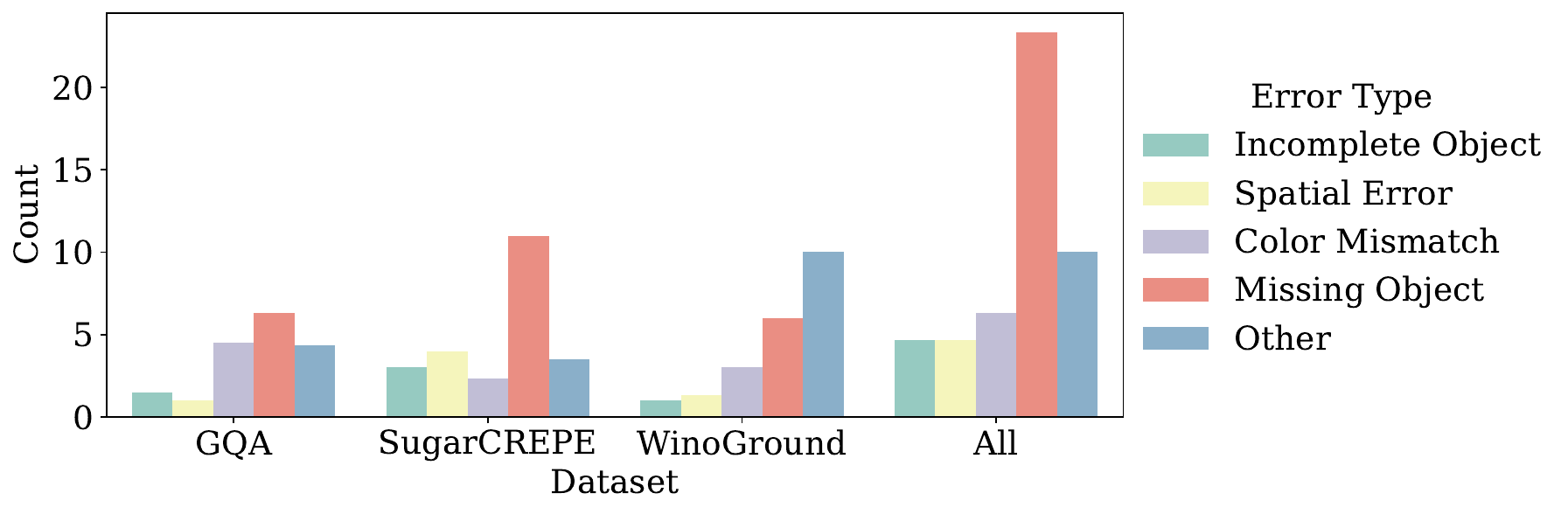}
    \vspace{-.3cm}\caption{Human Evaluation of Unit Test Quality. Bars show the average number of times annotators selected a source of error.}
    \label{fig:human_eval_unit_test_error_types}
\end{figure}

To assess the quality of unit tests we randomly sample 20 exampels from each of the three datasets, each corresponding to 5 unit tests, resulting in a total of 300 unit tests for evaluation. The unit tests were judged by three independent annotators, instructed with \texttt{\small Is the answer {answer} correct given the image?}, were \texttt{answer} was populated with the unit test expected answer, with binary yes/no answers. Table \ref{tab:human_eval_unit_test} breaks down the results showing that on average 75\% of unit tests are correct. Then the annotators optionally annotated the reason of failure by selecting from ``Missing Object'', ``Spatial Error'', ``Incomplete object'', ``Color Mismatch'', or ``Other''.  Figure \ref{fig:human_eval_unit_test_error_types} shows the break down by error type, highlighting ``Missing Object'' as the most common source of error.
\begin{table}[h]
\resizebox{\columnwidth}{!}{
\begin{tabular}{@{}cc|cc|cc|cc@{}}
\toprule
 \multicolumn{2}{c|}{GQA}          & \multicolumn{2}{c|}{WinoGround}         & \multicolumn{2}{c|}{SugarCREPE}     & \multicolumn{2}{c}{Avg.}   \\  
Acc. & $\kappa$ & Acc. & $\kappa$ & Acc. & $\kappa$ & Acc. & $\kappa$  \\ 

68.00&0.39&75.00&0.70&82.00&0.67&75.00&0.58\\

\bottomrule
\end{tabular}}
\vspace{-.3cm}\caption{Human Evaluation of Unit Test Quality. Accuracy corresponds to how many unit tests from the total were accurate and $\kappa$ is the mean Kohen Kappa across annotators.}
\label{tab:human_eval_unit_test}
\end{table}

\subsection{Program Correctness Evaluation}
\begin{figure}[h]
    \centering
    \includegraphics[width=\linewidth]{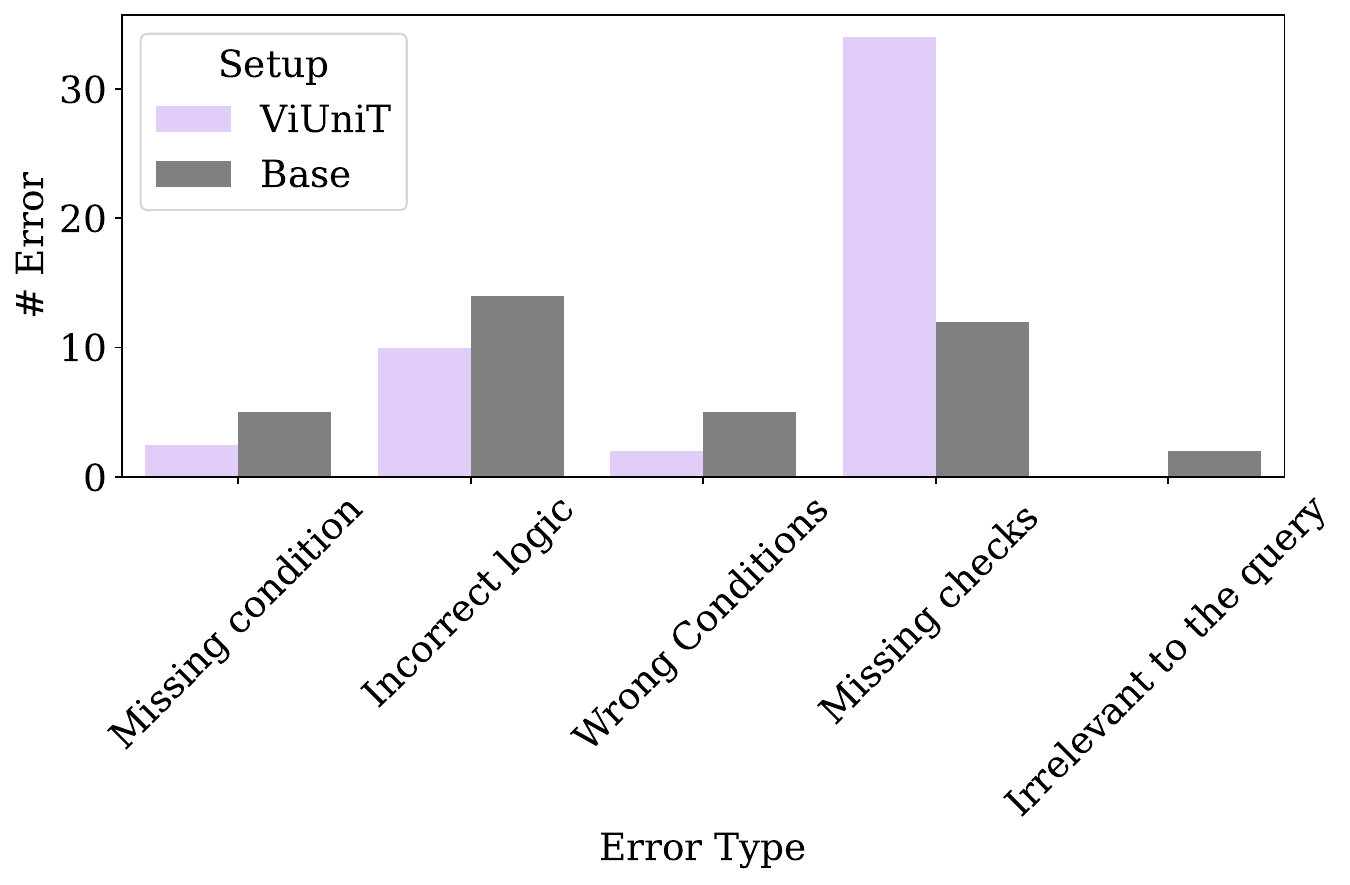}
    \vspace{-.3cm}\caption{Human Evaluation of Program Quality.}
    \label{fig:human_eval_program_error_types}
\end{figure}

To assess the improvements on program quality by applying \viUnit~ we conduct a human evaluation to rate GQA programs generated by the Base Setup and the programs selected from 5 candidate programs and 5 unit tests. Two annotators with 3+ years of Python experience graded programs using the following grading scheme: ``Correct: The code accurately and fully answers the query.'' (0), ``Partially Correct: The code answers the query but has some issues.'' (1), ``Incorrect: The code does not answer the query correctly.'' (2), and ``Irrelevant: The code is unrelated to the query.'' (3). In addition, they were optionally asked to select the source of error from ``Missing Condition'', ``Incorrect Logic'', ``Irrelevant to the query'', ``Wrong Conditions'', ``Missing Checks (e.g. could get list index out of range)'', ``Performance Issues'', ``Other''. Table \ref{tab:human_eval_programs_colored} shows the break down of program correctness improvements using \viUnit~ and Figure \ref{fig:human_eval_program_error_types} shows the error types identified in each method. \viUnit~ has ``Missing Checks'' as the most common error type, which mostly involves cases of not checking array length before accessing indices, typically still leading to correct solutions with reasonable programs, whereas the main culprit for program incorrectness in the base setup is ``Incorrect Logic''. 

\begin{table}[h]
\resizebox{\columnwidth}{!}{
\begin{tabular}{@{}cccc@{}}
\toprule
 & Base Setup\cellcolor{lightgray!10} & \cellcolor{brilliantlavender!30}ViUniT Setup (Ours) \\ 
Fully Correct ($\leq1$) & \cellcolor{lightgray!10}77\% & \cellcolor{brilliantlavender!30}\textbf{86\%} \\ 
Partially Correct ($<2$) & \cellcolor{lightgray!10}86\% & \cellcolor{brilliantlavender!30}\textbf{95\%} \\ 
Incorrect ($\geq 2$) & \cellcolor{lightgray!10}14\% & \cellcolor{brilliantlavender!30}\textbf{5\%} \\ 
Irrelevant ($>2$) & \cellcolor{lightgray!10}4\% & \cellcolor{brilliantlavender!30}\textbf{0\%} \\ 
$\kappa$ & \cellcolor{lightgray!10}0.24 & \cellcolor{brilliantlavender!30}0.30 \\ 
$\kappa_{bin}$ & \cellcolor{lightgray!10}0.59 & \cellcolor{brilliantlavender!30}0.40 \\ 
\bottomrule
\end{tabular}}
\vspace{-.3cm}\caption{Human Evaluation of Program Correctness. \textbf{Bold} is best.}
\label{tab:human_eval_programs_colored}
\end{table}

\section{Limitations and Social Ethics Impact}
\subsection{Limitations}
While \viUnit~provides significant advancements in the logical correctness and robustness of visual programs, our framework has several limitations that present opportunities for future enhancement. First, although \viUnit~improves program selection and execution by leveraging unit tests, it does not fully eliminate the issue of programs being correct for the wrong reasons, as shown by the human evaluation in Table \ref{tab:human_eval_programs_colored}. Our approach does not provide a formal guarantee of logical correctness, as it relies on automatically generated tests to evaluate candidate programs. Addressing this challenge opens avenues for integrating formal verification methods and more sophisticated testing strategies to further enhance program correctness. Second, while we optimize for maximizing input and output coverage during unit test generation, it is possible that the generated tests do not fully capture the space of edge cases or subtle logical errors in complex programs. This limitation highlights the potential for future work to develop more comprehensive coverage metrics and testing methodologies, possibly incorporating code-line execution coverage or other verifiable metrics. Third, the improved accuracy and robustness achieved by \viUnit~, as seen in Table \ref{tab:app_best_program}, come with an increase in computational effort. Generating candidate programs, sampling unit tests, and executing them on generated images introduce additional overhead. This trade-off between accuracy and efficiency presents an exciting challenge for future research to optimize the framework for real-time or resource-constrained applications, possibly through algorithmic improvements or efficient execution strategies. Additionally, enhancing the explainability of program failures remains an area for further development. Providing clear and interpretable feedback when a program is rejected or not selected due to poor performance on unit tests can improve user trust and facilitate debugging. Future work could focus on combining unit test outputs to offer detailed explanations of program failures. Finally, while \viUnit~has demonstrated effectiveness on VQA and ITM tasks, exploring its applicability to other domains or tasks involving different modalities or reasoning paradigms presents an opportunity to extend its impact. Adapting the framework to diverse domains can unlock new possibilities and broaden its utility. Despite these limitations, the advancements introduced by \viUnit~lay a strong foundation for future innovations in visual programming. By addressing these challenges, we can further enhance the robustness, efficiency, and applicability of the framework.

\subsection{Social Ethics Impact}
\viUnit~enhances the robustness and correctness of visual programming, with applications in critical domains like autonomous driving, healthcare, and education. By reducing instances where programs are correct for the wrong reasons, it helps build more trustworthy AI systems. However, ethical considerations are crucial for its responsible deployment: First, \viUnit~relies on pre-trained models, which may propagate biases (e.g., gender, racial, or cultural). Future work should focus on integrating bias detection and correction into unit test generation to promote fairness. Second, computational demands may limit access for resource-constrained organizations. Advancing efficiency and optimization can broaden accessibility and foster inclusivity. Third, increased computational needs may raise energy consumption. Optimizing for energy efficiency and using renewable energy can reduce the environmental impact, while improved AI reliability could deliver long-term sustainability benefits. Finally, in sensitive domains like healthcare or law, rigorous validation and transparency are essential. Finally, in sensitive domains such as healthcare or legal decision-making, while \viUnit~has the potential to enhance the correctness of visual programs, it is crucial to carefully communicate the framework's limitations and ensure rigorous validation. By proactively addressing ethical challenges and focusing on responsible development, we can maximize the positive societal impact of \viUnit, paving the way for more reliable, fair, and trustworthy AI systems.

\section{Qualitative Examples}
\begin{figure*}
    \centering
    \includegraphics[width=\linewidth]{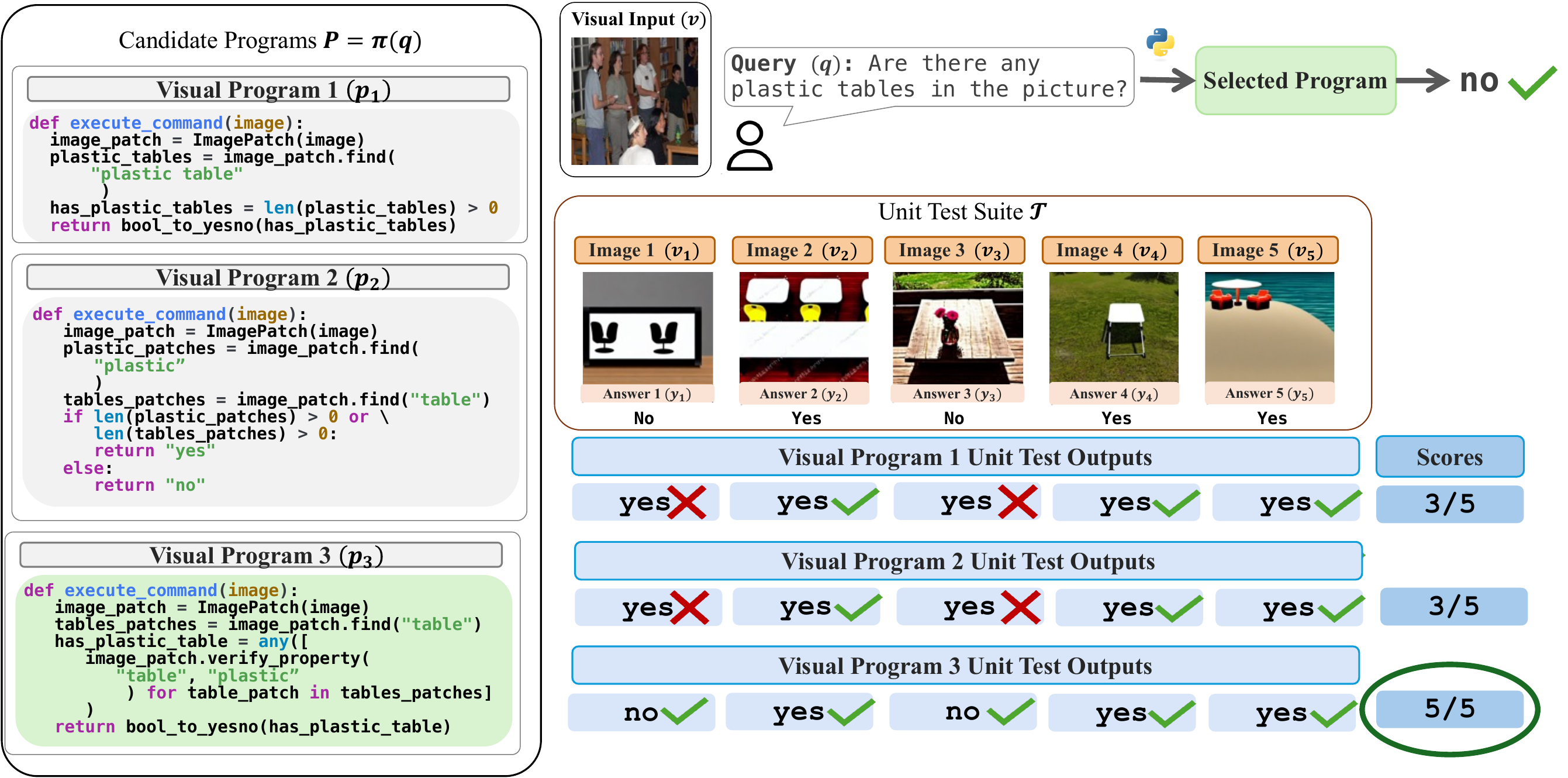}
    \caption{Program Selection Example}
    \label{fig:program_selection_A}
\end{figure*}

\begin{figure*}
    \centering
    \includegraphics[width=\linewidth]{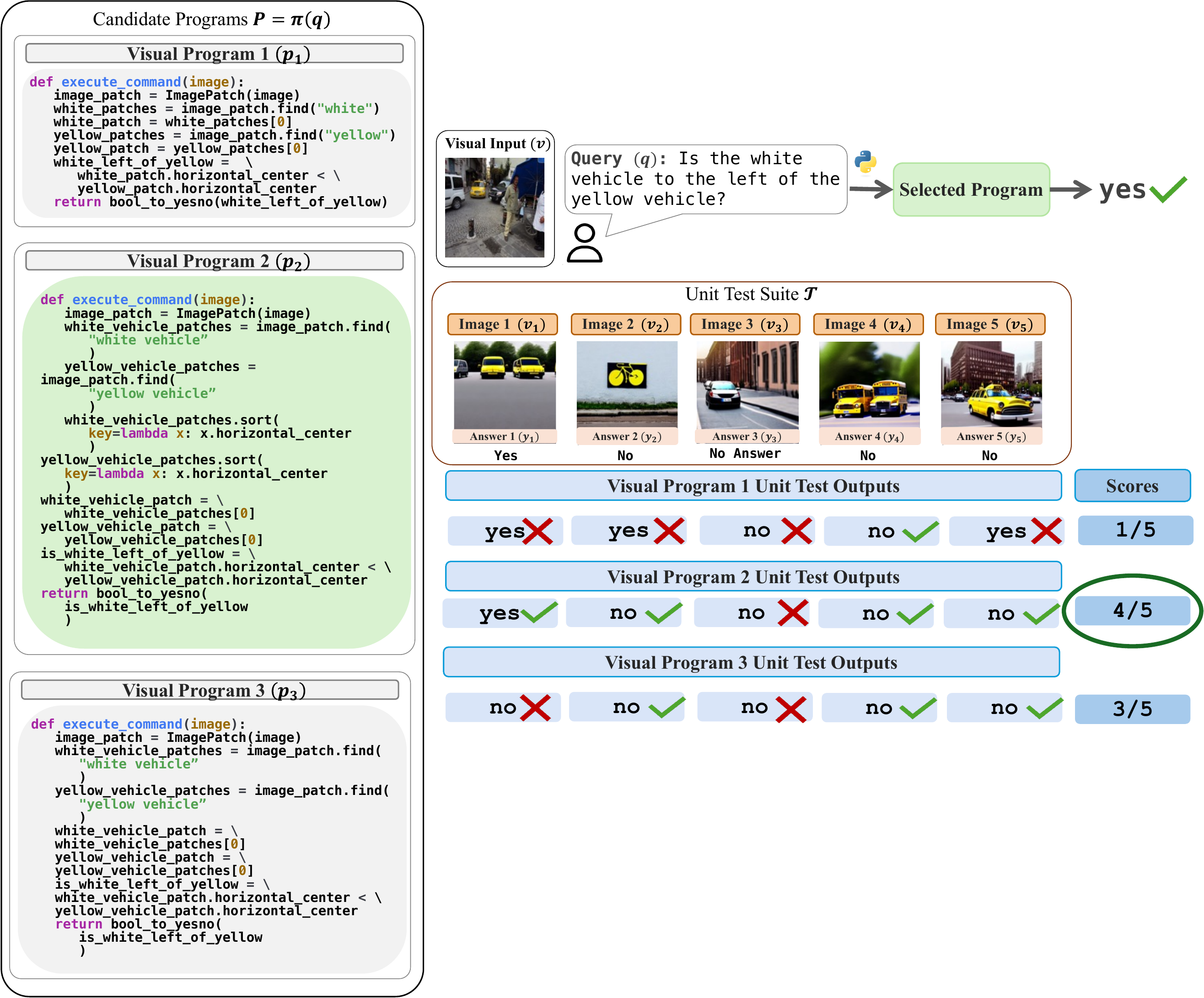}
    \caption{Program Selection Example}
    \label{fig:program_selection_B}
\end{figure*}

We present two program selection examples in Figures \ref{fig:program_selection_A} and \ref{fig:program_selection_B}.

\onecolumn

    \begin{lstlisting}[caption=API Prompt, label=code:api_prompt]
import math 

class ImagePatch:
    pass

    def __init__(
        self, image, left=None, lower=None, right=None, upper=None, category=None
    ):
        """Initializes an ImagePatch object by cropping the image at the given
        coordinates and stores the coordinates as attributes. If no coordinates are
        provided, the image is left unmodified, and the coordinates are set to the
        dimensions of the image.
        Parameters
        -------
        image : array_like
            An array-like of the original image.
        left, lower, right, upper : int
            An int describing the position of the (left/lower/right/upper) border of the
             crop's bounding box in the original image.
        category : str
            A string describing the name of the object in the image."""

        # Rectangles are represented as 4-tuples, (x1, y1, x2, y2),
        # with the upper left corner given first. The coordinate
        # system is assumed to have its origin in the upper left corner, so
        # upper must be less than lower and left must be less than right.

        self.left = left if left is not None else 0
        self.lower = lower if lower is not None else image.height
        self.right = right if right is not None else image.width
        self.upper = upper if upper is not None else 0
        self.cropped_image = image[:, image.shape[1]-upper:image.shape[1]-lower, left:right]
        self.horizontal_center = (self.left + self.right) / 2
        self.vertical_center = (self.upper + self.lower) / 2
        self.category = category

    def from_bounding_box(cls, image, bounding_box):
        """Initializes an ImagePatch object by cropping the image at the given
        coordinates and stores the coordinates as attributes.
        Parameters
        -------
        image : array_like
            An array-like of the original image.
        bounding_box : dict
            A dictionary like {"box": [left, lower, right, upper], "category": str}."""
        pass

    @property
    def area(self):
        """
        Returns the area of the bounding box.

        Examples
        --------
        >>> # What color is the largest foo?
        >>> def execute_command(image) -> str:
        >>>     image_patch = ImagePatch(image)
        >>>     foo_patches = image_patch.find("foo")
        >>>     foo_patches.sort(key=lambda x: x.area)
        >>>     largest_foo_patch = foo_patches[-1]
        >>>     return largest_foo_patch.simple_query("What is the color?")
        """
        pass

    def find(self, object_name):
        """Returns a list of ImagePatch objects matching object_name contained in the
        crop if any are found.
        Otherwise, returns an empty list.
        Parameters
        ----------
        object_name : str
            the name of the object to be found

        Returns
        -------
        List[ImagePatch]
            a list of ImagePatch objects matching object_name contained in the crop

        Examples
        --------
        >>> # return the foo
        >>> def execute_command(image) -> List[ImagePatch]:
        >>>     image_patch = ImagePatch(image)
        >>>     foo_patches = image_patch.find("foo")
        >>>     return foo_patches
        """
        pass

    def exists(self, object_name):
        """Returns True if the object specified by object_name is found in the image,
        and False otherwise.
        Parameters
        -------
        object_name : str
            A string describing the name of the object to be found in the image.

        Examples
        -------
        >>> # Are there both foos and garply bars in the photo?
        >>> def execute_command(image)->str:
        >>>     image_patch = ImagePatch(image)
        >>>     is_foo = image_patch.exists("foo")
        >>>     is_garply_bar = image_patch.exists("garply bar")
        >>>     return bool_to_yesno(is_foo and is_garply_bar)
        """
        pass

    def verify_property(self, object_name, visual_property):
        """Returns True if the object possesses the visual property, and False otherwise.
        Differs from 'exists' in that it presupposes the existence of the object 
        specified by object_name, instead checking whether the object possesses
        the property.
        Parameters
        -------
        object_name : str
            A string describing the name of the object to be found in the image.
        visual_property : str
            String describing the simple visual property (e.g., color, shape, material)
            to be checked.

        Examples
        -------
        >>> # Do the letters have blue color?
        >>> def execute_command(image) -> str:
        >>>     image_patch = ImagePatch(image)
        >>>     letters_patches = image_patch.find("letters")
        >>>     # Question assumes only one letter patch
        >>>     return bool_to_yesno(letters_patches[0].verify_property("letters", "blue"))
        """
        pass

    def simple_query(self, question):
        """Returns the answer to a basic question asked about the image.
        If no question is provided, returns the answer to "What is this?".
        The questions are about basic perception, and are not meant to be used for
        complex reasoning or external knowledge.
        Parameters
        -------
        question : str
            A string describing the question to be asked.

        Examples
        -------

        >>> # Which kind of baz is not fredding?
        >>> def execute_command(image) -> str:
        >>>     image_patch = ImagePatch(image)
        >>>     baz_patches = image_patch.find("baz")
        >>>     for baz_patch in baz_patches:
        >>>         if not baz_patch.verify_property("baz", "fredding"):
        >>>             return baz_patch.simple_query("What is this baz?")

        >>> # What color is the foo?
        >>> def execute_command(image) -> str:
        >>>     image_patch = ImagePatch(image)
        >>>     foo_patches = image_patch.find("foo")
        >>>     foo_patch = foo_patches[0]
        >>>     return foo_patch.simple_query("What is the color?")

        >>> # Is the second bar from the left quuxy?
        >>> def execute_command(image) -> str:
        >>>     image_patch = ImagePatch(image)
        >>>     bar_patches = image_patch.find("bar")
        >>>     bar_patches.sort(key=lambda x: x.horizontal_center)
        >>>     bar_patch = bar_patches[1]
        >>>     return bar_patch.simple_query("Is the bar quuxy?")
        """
        pass

    def crop_left_of_bbox(self, left, lower, right, upper):
        """Returns an ImagePatch object representing the area to the left of the given
        bounding box coordinates.

        Parameters
        ----------
        left, lower, right, upper : int
            The coordinates of the bounding box.

        Returns
        -------
        ImagePatch
            An ImagePatch object representing the cropped area.

        Examples
        --------
        >>> # Is the bar to the left of the foo quuxy?
        >>> def execute_command(image) -> str:
        >>>     image_patch = ImagePatch(image)
        >>>     foo_patch = image_patch.find("foo")[0]
        >>>     left_of_foo_patch = image_patch.crop_left_of_bbox(
        >>>         foo_patch.left, foo_patch.lower, foo_patch.right, foo_patch.upper
        >>>     )
        >>>     return bool_to_yesno(left_of_foo_patch.verify_property("bar", "quuxy"))
        """
        pass

    def crop_right_of_bbox(self, left, lower, right, upper):
        """Returns an ImagePatch object representing the area to the right of the given
        bounding box coordinates.

        Parameters
        ----------
        left, lower, right, upper : int
            The coordinates of the bounding box.

        Returns
        -------
        ImagePatch
            An ImagePatch object representing the cropped area.

        Examples
        --------
        >>> # Is the bar to the right of the foo quuxy?
        >>> def execute_command(image) -> str:
        >>>     image_patch = ImagePatch(image)
        >>>     foo_patch = image_patch.find("foo")[0]
        >>>     right_of_foo_patch = image_patch.crop_right_of_bbox(
        >>>         foo_patch.left, foo_patch.lower, foo_patch.right, foo_patch.upper
        >>>     )
        >>>     return bool_to_yesno(right_of_foo_patch.verify_property("bar", "quuxy"))
        """
        pass

    def crop_below_bbox(self, left, lower, right, upper):
        """Returns an ImagePatch object representing the area below the given
        bounding box coordinates.

        Parameters
        ----------
        left, lower, right, upper : int
            The coordinates of the bounding box.

        Returns
        -------
        ImagePatch
            An ImagePatch object representing the cropped area.

        Examples
        --------
        >>> # Is the bar below the foo quuxy?
        >>> def execute_command(image) -> str:
        >>>     image_patch = ImagePatch(image)
        >>>     foo_patch = image_patch.find("foo")[0]
        >>>     below_foo_patch = image_patch.crop_below_bbox(
        >>>         foo_patch.left, foo_patch.lower, foo_patch.right, foo_patch.upper
        >>>     )
        >>>     return bool_to_yesno(below_foo_patch.verify_property("bar", "quuxy"))
        """
        pass

    def crop_above_bbox(self, left, lower, right, upper):
        """Returns an ImagePatch object representing the area above the given
        bounding box coordinates.

        Parameters
        ----------
        left, lower, right, upper : int
            The coordinates of the bounding box.

        Returns
        -------
        ImagePatch
            An ImagePatch object representing the cropped area.

        Examples
        --------
        >>> # Is the bar above the foo quuxy?
        >>> def execute_command(image) -> str:
        >>>     image_patch = ImagePatch(image)
        >>>     foo_patch = image_patch.find("foo")[0]
        >>>     above_foo_patch = image_patch.crop_above_bbox(
        >>>         foo_patch.left, foo_patch.lower, foo_patch.right, foo_patch.upper
        >>>     )
        >>>     return bool_to_yesno(above_foo_patch.verify_property("bar", "quuxy"))
        """
        pass


def best_image_match(list_patches: List[ImagePatch], content: List[str], return_index=False) -> Union[ImagePatch, int]:
    """Returns the patch most likely to contain the content.
    Parameters
    ----------
    list_patches : List[ImagePatch]
    content : List[str]
        the object of interest
    return_index : bool
        if True, returns the index of the patch most likely to contain the object

    Returns
    -------
    int
        Patch most likely to contain the object
    """
    return best_image_match(list_patches, content, return_index)

def bool_to_yesno(bool_answer: bool) -> str:
    return "yes" if bool_answer else "no"

Write a function using Python and the ImagePatch class (above) that could be executed to provide an answer to the query.

Consider the following guidelines:
- Use base Python (comparison, sorting) for basic logical operations, left/right/up/down, math, etc.

# Examples of how to use the API
INSERT_CONTEXT_HERE

Query: INSERT_QUERY_HERE
Program:
\end{lstlisting}

    \begin{lstlisting}[caption=ITM In-Context Examples, label=code:itm_ice]
# Query: Verify image matches text="An airplane is flying in the sky, and birds are flying below it."
def execute_command(image) -> str:
    image_patch = ImagePatch(image)
    airplane_patches = image_patch.find("airplane")
    bird_patches = image_patch.find("bird")

    airplane_in_sky = any(
        airplane_patch.vertical_center > image_patch.height * 0.6
        for airplane_patch in airplane_patches
    )

    birds_below_airplane = any(
        bird_patch.upper <= airplane_patch.lower
        for bird_patch in bird_patches for airplane_patch in airplane_patches
    )

    return bool_to_yesno(airplane_in_sky and birds_below_airplane)

# Query: Verify image matches text="The bird is flying above the tree, and a cat is sitting under the tree."
def execute_command(image) -> str:
    image_patch = ImagePatch(image)
    bird_patches = image_patch.find("bird")
    tree_patches = image_patch.find("tree")
    cat_patches = image_patch.find("cat")

    bird_above_tree = any(
        bird_patch.lower >= tree_patch.upper and
        abs(bird_patch.horizontal_center - tree_patch.horizontal_center) < 50
        for bird_patch in bird_patches for tree_patch in tree_patches
    )

    cat_under_tree = any(
        cat_patch.upper <= tree_patch.lower and
        abs(cat_patch.horizontal_center - tree_patch.horizontal_center) < 50
        for cat_patch in cat_patches for tree_patch in tree_patches
    )

    return bool_to_yesno(bird_above_tree and cat_under_tree)

# Query: Verify image matches text="The apple is on top of the book, and the pen is beside the book."
def execute_command(image) -> str:
    image_patch = ImagePatch(image)
    apple_patches = image_patch.find("apple")
    book_patches = image_patch.find("book")
    pen_patches = image_patch.find("pen")

    apple_on_book = any(
        apple_patch.lower >= book_patch.upper and
        book_patch.left <= apple_patch.horizontal_center <= book_patch.right
        for apple_patch in apple_patches for book_patch in book_patches
    )

    pen_beside_book = any(
        abs(pen_patch.horizontal_center - book_patch.horizontal_center) < 50 and
        abs(pen_patch.vertical_center - book_patch.vertical_center) < 100
        for pen_patch in pen_patches for book_patch in book_patches
    )

    return bool_to_yesno(apple_on_book and pen_beside_book)

#Query: Verify image matches text="A man is riding a bicycle, and a dog is running beside him."
def execute_command(image) -> str:
    image_patch = ImagePatch(image)
    man_patches = image_patch.find("man")
    bicycle_patches = image_patch.find("bicycle")
    dog_patches = image_patch.find("dog")

    man_on_bicycle = any(
        man_patch.left <= bicycle_patch.right and man_patch.right >= bicycle_patch.left and
        man_patch.lower <= bicycle_patch.upper and man_patch.upper >= bicycle_patch.lower
        for man_patch in man_patches for bicycle_patch in bicycle_patches
    )

    dog_beside_man = any(
        abs(dog_patch.horizontal_center - man_patch.horizontal_center) < 100 and
        abs(dog_patch.vertical_center - man_patch.vertical_center) < 50
        for dog_patch in dog_patches for man_patch in man_patches
    )

    return bool_to_yesno(man_on_bicycle and dog_beside_man)
\end{lstlisting}

    \begin{lstlisting}[caption=VQA In-Context Examples, label=code:vqa_ice]
# Query: Is the vehicle in the top of the image?
def execute_command(image) -> str:
    image_patch = ImagePatch(image)
    # Assume there's only one vehicle patch.
    vehicle_patch = image_patch.find("vehicle")[0]
    vehicle_in_top_half = vehicle_patch.vertical_center > image_patch.vertical_center
    return bool_to_yesno(vehicle_in_top_half)

# Query: Are there trains or fences in this scene?
def execute_command(image) -> str:
    image_patch = ImagePatch(image)
    trains = image_patch.find("train")
    fences = image_patch.find("fence")
    has_trains_or_fences = len(trains) > 0 or len(fences) > 0
    return bool_to_yesno(has_trains_or_fences)

# Query: Is the pillow in the top part or in the bottom of the picture?
def execute_command(image) -> str:
    image_patch = ImagePatch(image)
    pillow_patches = image_patch.find("pillow")
    pillow_patch = pillow_patches[0]
    pillow_in_top_half = pillow_patch.vertical_center > image_patch.vertical_center
    if pillow_in_top_half:
        return "top"
    else:
        return "bottom"

# Query: What color is the curtain that is to the right of the mirror?
def execute_command(image) -> str:
    image_patch = ImagePatch(image)
    mirror_patches = image_patch.find("mirror")
    mirror_patch = mirror_patches[0]
    right_of_mirror_patch = image_patch.crop_right_of_bbox(
        mirror_patch.left, mirror_patch.lower, mirror_patch.right, mirror_patch.upper
    )
    return right_of_mirror_patch.simple_query("What color is the curtain?")
\end{lstlisting}

    \begin{lstlisting}[caption=Reprompting with Unit Tests VQA, label=code:viunit_reprompting_vqa]
INSERT_IMAGE_PATCH_API

You are provided a Python program that answers a query about an image, with a set of tests with the corresponding outputs and exected responses. 
Correct the Python program such that it passes the tests. 
- Ensure the corrected program is different than the incorrect program provided. 

Query: Is there a blue chair in the image?
Incorrect Program: 
def execute_command(image):
    image_patch = ImagePatch(image)
    blue_chair = image_patch.find("chair")
    if not blue_chair:
        return "No"
    is_blue = any([chair.verify_property("blue") for chair in blue_chair])
    return "Yes" if is_blue else "No"
Test Cases:
Test A
Image Content: "A room with a red chair"
Ground Truth Answer: "No"
Program Output: "Error: verify_property() missing 1 required positional argument: 'visual_property'"
Test B
Image Content: "A room with a blue chair under the window"
Ground Truth Answer: "Yes"
Program Output: "Error: verify_property() missing 1 required positional argument: 'visual_property'"
Test C
Image Content: "An empty room"
Ground Truth Answer: "No"
Program Output: "No"
Test D
Image Content: "A garden with a blue chair"
Ground Truth Answer: "Yes"
Program Output: "Error: verify_property() missing 1 required positional argument: 'visual_property'"
Test E
Image Content: "A room with several chairs, all red"
Ground Truth Answer: "No"
Program Output: "Error: verify_property() missing 1 required positional argument: 'visual_property'"
Corrected Program:
def execute_command(image):
    image_patch = ImagePatch(image)
    chair_patches = image_patch.find("chair")
    if not chair_patches:
        return "No"  # No chairs found
    blue_chair_found = any(chair.verify_property("chair", "blue") for chair in chair_patches)
    return "Yes" if blue_chair_found else "No"

Query: "Are there any flowers to the left of the house?"
Incorrect Program: 
def execute_command(image):
    image_patch = ImagePatch(image)
    house_patches = image_patch.find("house")
    if not house_patches:
        return "No house found"
    left_of_house_patch = image_patch.crop_left_of_bbox(
        house_patches.left, house_patches.lower, house_patches.right, house_patches.upper
    )  # Incorrect attribute access
    return "Yes" if left_of_house_patch.exists("flower") else "No"
Test Cases:
Test A
Image Content: "An image of a garden without any buildings."
Ground Truth Answer: "No house found"
Program Output: "Error: 'list' object has no attribute 'left'"
Test B
Image Content: "A house without a garden"
Ground Truth Answer: "No flowers found"
Program Output: "Error: 'list' object has no attribute 'left'"
Test C
Image Content: "A house with many flowers around"
Ground Truth Answer: "Yes"
Program Output: "Error: 'list' object has no attribute 'left'"
Test D
Image Content: "A house with flowers only on the right side"
Ground Truth Answer: "No"
Program Output: "Error: 'list' object has no attribute 'left'"
Test E
Image Content: "An image with flowers but no house"
Ground Truth Answer: "No house found"
Program Output: "Error: 'list' object has no attribute 'left'"
Corrected Program: 
def execute_command(image):
    image_patch = ImagePatch(image)
    house_patches = image_patch.find("house")
    if not house_patches:
        return "No house found"
    for house_patch in house_patches:
        left_of_house_patch = image_patch.crop_left_of_bbox(
            house_patch.left, house_patch.lower, house_patch.right, house_patch.upper
        )
        flowers_found = left_of_house_patch.find("flower")
        if flowers_found:
            return "Yes"
    return "No"

Query: Who wears a green shirt?
Incorrect Program: 
def execute_command(image):
    image_patch = ImagePatch(image)
    people_patches = image_patch.find("person")
    if not people_patches:
        return "No one"
    person_wearing_green_shirt = None
    for index, person_patch in enumerate(people_patches):
        green_patches = person_patch.find("green")
        if green_patches:
            person_wearing_green_shirt = index
            break 
    if person_wearing_green_shirt == None:
        return "No one"
    else:
        return people_patches[person_wearing_green_shirt].simple_query("Who is this?")
Test Cases:
Test A
Image Content: "An image of an empty room"
Ground Truth Answer: "No one"
Program Output: "No one"
Test B
Image Content: "A young girl wearing a green dress and a boy wearing a blue shirt."
Ground Truth Answer: "No one"
Program Output: "girl"
Test C
Image Content: "A man wearing a red shirt and a woman wearing a green shirt."
Ground Truth Answer: "woman"
Program Output: "woman"
Test D
Image Content: "A boy wearing a green shirt."
Ground Truth Answer: "boy"
Program Output: "boy"
Test E
Image Content: "Two people wearing green shirts: a man and a woman"
Ground Truth Answer: "man and woman"
Program Output: "man"
Corrected Program:
def execute_command(image):
    image_patch = ImagePatch(image)
    people_patches = image_patch.find("person")
    if not people_patches:
        return "No people found"
    people_wearing_green_shirts = []
    for index, person_patch in enumerate(people_patches):
        if person_patch.verify_property("clothing", "shirt") and person_patch.verify_property("color", "green"):
            people_wearing_green_shirts.append(index)
    if not people_wearing_green_shirts:
        return "No one"
    wearing_green_shirts = ' and '.join([people_patches[i].simple_query("Who is this?") for i in people_wearing_green_shirts])
    return wearing_green_shirts

Query: "Is the blue ball inside the box?"
Incorrect Program:
def execute_command(image):
    image_patch = ImagePatch(image)
    ball_patches = image_patch.find("ball")
    if not ball_patches:
        return "No"  
    blue_ball = [ball for ball in ball_patches if ball.verify_property("color", "blue")]
    if not blue_ball:
        return "No"  # No blue ball found
    box_patches = image_patch.find("box")
    if not box_patches:
        return "No"  
    return "Yes" 
Test Cases:
Test A

Image Content: "A blue ball is outside a box"
Ground Truth Answer: "No"
Program Output: "Yes"
Test B
Image Content: "A red ball is inside a box"
Ground Truth Answer: "No"
Program Output: "No"
Test C
Image Content: "A blue ball is inside a box"
Ground Truth Answer: "Yes"
Program Output: "Yes"
Test D
Image Content: "No balls or boxes in the image"
Ground Truth Answer: "No"
Program Output: "No"
Test E
Image Content: "Multiple blue balls, all outside boxes"
Ground Truth Answer: "No"
Program Output: "Yes"
Corrected Program:
def execute_command(image):
    image_patch = ImagePatch(image)
    ball_patches = image_patch.find("ball")
    if not ball_patches:
        return "No"  # No ball found
    blue_ball = [ball for ball in ball_patches if ball.verify_property("color", "blue")]
    if not blue_ball:
        return "No"  # No blue ball found
    box_patches = image_patch.find("box")
    if not box_patches:
        return "No"  # No box found
    blue_ball_patch = blue_ball[0]
    for box_patch in box_patches:
        if (box_patch.left <= blue_ball_patch.left and
            box_patch.right >= blue_ball_patch.right and
            box_patch.upper <= blue_ball_patch.upper and
            box_patch.lower >= blue_ball_patch.lower):
            return "Yes" 
    return "No"

Query: INSERT_QUERY_HERE
Incorrect Program:
INSERT_CODE_HERE
Test Cases:
INSERT_UNIT_TEST_OUTPUTS_HERE
Corrected Program:
\end{lstlisting}

    \begin{lstlisting}[caption=Reprompting with Unit Tests ITM, label=code:viunit_reprompting_itm]
INSERT_IMAGE_PATCH_API

You are provided a Python program that answers a query about an image, with a set of tests with the corresponding outputs and exected responses. 
Correct the Python program such that it passes the tests. 
- Ensure the corrected program is different than the incorrect program provided. 

Query: "Verify image matches text="An airplane is flying in the sky, and birds are flying below it.""
Incorrect Program: 
def execute_command(image):
    image_patch = ImagePatch(image)
    airplane = image_patch.find("airplane")
    birds = image_patch.find("birds")
    if not airplane or not birds:
        return "No"
    if airplane[0].vertical_center >= birds[0].vertical_center:
        return "Yes"
    return "No"
Test Cases:
Test A
Image Content: "An airplane flying high in the sky with birds below it."
Ground Truth Answer: "Yes"
Program Output: "Yes"
Test B
Image Content: "Birds are flying above and below an airplane in the sky."
Ground Truth Answer: "No"
Program Output: "Yes"
Test C
Image Content: "An airplane and birds flying side by side."
Ground Truth Answer: "No"
Program Output: "Yes"
Test D
Image Content: "Only an airplane is flying in the sky."
Ground Truth Answer: "No"
Program Output: "No"
Test E
Image Content: "Birds flying in the sky with no airplane present."
Ground Truth Answer: "No"
Program Output: "No"
Corrected Program::
def execute_command(image):
    image_patch = ImagePatch(image)
    airplane_patches = image_patch.find("airplane")
    bird_patches = image_patch.find("bird")
    if not airplane_patches or not bird_patches:
        return "No"
    airplane = airplane_patches[0]
    birds_below = all(bird.vertical_center > airplane.vertical_center for bird in bird_patches)
    return "Yes" if birds_below else "No"

Query: "Verify image matches text="The bird is flying above the tree, and a cat is sitting under the tree.""
Incorrect Program:
def execute_command(image):
    image_patch = ImagePatch(image)
    tree = image_patch.find("tree")
    bird = image_patch.find("bird")
    cat = image_patch.find("cat")
    if not tree or not bird or not cat:
        return "No"
    if bird[0].vertical_center < tree[0].vertical_center and cat[0].vertical_center > tree[0].vertical_center:
        return "Yes"
    return "No"
Test Cases:
Test A
Image Content: "A bird flying above a tree and a cat under the tree."
Ground Truth Answer: "Yes"
Program Output: "Yes"
Test B
Image Content: "A cat sitting above the tree and a bird flying below it."
Ground Truth Answer: "No"
Program Output: "Yes"
Test C
Image Content: "A bird sitting in the tree with no cat around."
Ground Truth Answer: "No"
Program Output: "No"
Test D
Image Content: "A cat climbing the tree while a bird flies overhead."
Ground Truth Answer: "No"
Program Output: "Yes"
Test E
Image Content: "A bird flying above a tree with a dog under the tree."
Ground Truth Answer: "No"
Program Output: "No"
Corrected Program:
def execute_command(image):
    image_patch = ImagePatch(image)
    tree_patches = image_patch.find("tree")
    bird_patches = image_patch.find("bird")
    cat_patches = image_patch.find("cat")
    if not tree_patches or not bird_patches or not cat_patches:
        return "No"
    tree = tree_patches[0]
    bird_above = all(bird.vertical_center < tree.vertical_center for bird in bird_patches)
    cat_below = all(cat.vertical_center > tree.vertical_center for cat in cat_patches)
    return "Yes" if bird_above and cat_below else "No"

Query: "Verify image matches text="A car is parked near a tree, and a bird is sitting on the tree.""
Incorrect Program:
def execute_command(image):
    image_patch = ImagePatch(image)
    car = image_patch.find("car")
    tree = image_patch.find("tree")
    bird = image_patch.find("bird")
    if not car or not tree or not bird:
        return "No"
    if car.horizontal_center - tree.horizontal_center < 100 and bird.vertical_center < tree.vertical_center:
        return "Yes"
    return "No"
Test Cases:
Test A
Image Content: "A car parked near a tree with a bird sitting on it."
Ground Truth Answer: "Yes"
Program Output: AttributeError: 'list' object has no attribute 'horizontal_center'
Test B
Image Content: "A car far from a tree with a bird on the ground."
Ground Truth Answer: "No"
Program Output: AttributeError: 'list' object has no attribute 'horizontal_center'
Test C
Image Content: "A tree with a bird on it but no car nearby."
Ground Truth Answer: "No"
Program Output: "No"
Test D
Image Content: "A car parked near a tree with no bird in sight."
Ground Truth Answer: "No"
Program Output: AttributeError: 'list' object has no attribute 'horizontal_center'
Test E
Image Content: "A car and a bird but no tree present."
Ground Truth Answer: "No"
Program Output: AttributeError: 'list' object has no attribute 'horizontal_center'
Corrected Program:
def execute_command(image):
    image_patch = ImagePatch(image)
    car_patches = image_patch.find("car")
    tree_patches = image_patch.find("tree")
    bird_patches = image_patch.find("bird")
    if not car_patches or not tree_patches or not bird_patches:
        return "No"
    car = car_patches[0]
    tree = tree_patches[0]
    bird = bird_patches[0]
    car_near_tree = abs(car.horizontal_center - tree.horizontal_center) < 100
    bird_on_tree = bird.vertical_center < tree.vertical_center
    return "Yes" if car_near_tree and bird_on_tree else "No"

Query: "Verify image matches text="A man is holding a red balloon, and a child is reaching up to grab it.""
Incorrect Program:
def execute_command(image):
    image_patch = ImagePatch(image)
    man = image_patch.find("man")
    balloon = image_patch.find("balloon")
    child = image_patch.find("child")
    if not man or not balloon or not child:
        return "No"
    if balloon[0].verify_property("red") and child[0].vertical_center < balloon[0].vertical_center:
        return "Yes"
    return "No"
Test Cases:
Test A
Image Content: "A man holding a red balloon, with a child reaching up."
Ground Truth Answer: "Yes"
Program Output: TypeError: verify_property() missing 1 required positional argument: 'visual_property'
Test B
Image Content: "A man holding a blue balloon, with a child below him."
Ground Truth Answer: "No"
Program Output: TypeError: verify_property() missing 1 required positional argument: 'visual_property'
Test C
Image Content: "A man holding a flower, with a child next to him."
Ground Truth Answer: "No"
Program Output: "No"
Corrected Program:
def execute_command(image):
    image_patch = ImagePatch(image)
    man_patches = image_patch.find("man")
    balloon_patches = image_patch.find("balloon")
    child_patches = image_patch.find("child")
    if not man_patches or not balloon_patches or not child_patches:
        return "No"
    balloon = balloon_patches[0]
    is_red_balloon = balloon.verify_property("balloon", "red")
    child_below_balloon = all(child.vertical_center < balloon.vertical_center for child in child_patches)
    return "Yes" if is_red_balloon and child_below_balloon else "No"

Query: "Verify image matches text="A cat is sitting on the table, and a book is lying beside it.""
Incorrect Program:
def execute_command(image):
    image_patch = ImagePatch(image)
    cat = image_patch.find("cat")
    book = image_patch.find("book")
    if not cat or not book:
        return "No"
    if abs(book[0].horizontal_center - cat[0].horizontal_center) < 50:
        return "Yes"
    return "No"
Test Cases:
Test A
Image Content: "A cat sitting on the table with a book beside it."
Ground Truth Answer: "Yes"
Program Output: "Yes"
Test B
Image Content: "A cat sitting on the floor with a book beside it."
Ground Truth Answer: "No"
Program Output: "Yes"
Test C
Image Content: "A cat sitting on the table with no book around."
Ground Truth Answer: "No"
Program Output: "No"
Test D
Image Content: "A book lying on the table with no cat in sight."
Ground Truth Answer: "No"
Program Output: "No"
Test E
Image Content: "A cat sitting on the table with a book on the floor."
Ground Truth Answer: "No"
Program Output: "Yes"
Corrected Program:
def execute_command(image):
    image_patch = ImagePatch(image)
    cat_patches = image_patch.find("cat")
    book_patches = image_patch.find("book")
    table_patches = image_patch.find("table")
    if not cat_patches or not book_patches or not table_patches:
        return "No"
    cat = cat_patches[0]
    book = book_patches[0]
    table = table_patches[0]
    is_cat_on_table = cat.vertical_center < table.vertical_center and abs(cat.horizontal_center - table.horizontal_center) < 50
    is_book_beside_cat = abs(book.horizontal_center - cat.horizontal_center) < 50
    return "Yes" if is_cat_on_table and is_book_beside_cat else "No"

Query: INSERT_QUERY_HERE
Incorrect Program:
INSERT_CODE_HERE
Test Cases:
INSERT_UNIT_TEST_OUTPUTS_HERE
Corrected Program:
\end{lstlisting}

    \begin{lstlisting}[caption=VQA Unit Test Generation In Context Examples, label=code:ut_gen_ice_vqa]
Query: Is there a cat or dog in the image?
Tests:
1. Image Caption: "A grey tabby cat peacefully napping on a plush sofa" Answer: yes
2. Image Caption: "A lively golden retriever bounding across a grassy field in the park" Answer: yes
3. Image Caption: "Twin Siamese cats playfully swatting at a bright yellow ball" Answer: yes
4. Image Caption: "A cluster of wild horses trotting along the sandy shores of a sunlit beach" Answer: no
5. Image Caption: "An orange cat and a black Labrador playfully tugging on a rope toy" Answer: yes
6. Image Caption: "A modern living room featuring sleek furniture and devoid of any pets" Answer: no

Query: Is there a red truck or bus in the image?
Tests:
1. Image Caption: "A vibrant red Ford pickup parked beside a country road" Answer: yes
2. Image Caption: "A red double-decker bus navigating through a busy downtown street" Answer: yes
3. Image Caption: "A large blue semi-truck cruising down an interstate highway" Answer: no
4. Image Caption: "A quiet suburban street devoid of any large vehicles like buses or trucks" Answer: no
5. Image Caption: "A shiny red Ferrari speeding on a professional race track" Answer: no
6. Image Caption: "An array of red delivery trucks lined up in a distribution center parking lot" Answer: yes
7. Image Caption: "Several bright yellow school buses parked in a row at a local school" Answer: no

Query: What color is the largest car in the image?
Tests:
1. Image Caption: "A large blue Ford pickup truck driving on a busy highway" Answer: blue
2. Image Caption: "A city street empty of any large vehicles like buses or trucks" Answer: no answer
3. Image Caption: "A row of green food trucks serving lunch in an urban park" Answer: green
4. Image Caption: "A scene with a green public bus next to a smaller blue pickup at an intersection" Answer: green

Query: Is the vase to the left or right of the center?
Tests:
1. Image Caption: "A delicate porcelain vase positioned on the right end of a mahogany dining table" Answer: right
2. Image Caption: "A tall glass vase sitting on the left side of a neatly made bed in a sunlit room" Answer: left
3. Image Caption: "A ceramic vase centrally placed on a round table surrounded by chairs" Answer: center

Query: What is the highest object in the image?
Tests:
1. Image Caption: "A massive skyscraper dominating the skyline among lower city buildings" Answer: skyscraper
2. Image Caption: "A lone oak tree surpassing the height of the cottage it stands next to" Answer: tree
3. Image Caption: "Colorful balloons drifting above the treetops in a clear sky" Answer: balloons
4. Image Caption: "A commercial jet flying high above the city's tallest skyscrapers" Answer: plane
5. Image Caption: "A majestic eagle soaring high above a vast canyon landscape" Answer: eagle
6. Image Caption: "A figure standing on the peak of a grassy hill under a blue sky" Answer: person

Query: INSERT_QUERY_HERE
Tests:
\end{lstlisting}

    \begin{lstlisting}[caption=ITM Unit Test Generation In Context Examples, label=code:ut_gen_ice_itm]
Query: Is the drawing of a tree on the hill, and a river that flows at the bottom of the hill?
Tests:
1. Image Caption: "A solitary tree stands atop a gentle hill, with a flowing river winding below it." Answer: yes
2. Image Caption: "A tree on a grassy hill under a clear sky." Answer: no
3. Image Caption: "A river meandering through a dense forest of tall trees." Answer: no
4. Image Caption: "A panoramic view of rolling hills in the desert, with a river at the bottom." Answer: no
5. Image Caption: "A vast plain with a river running through fields of wildflowers." Answer: no
6. Image Caption: Image Caption: "A hill with multiple trees and a river flowing nearby." Answer: yes

Query: Is the drawing of an airplane flying in the sky, and birds flying below it?
Tests:
1. Image Caption:  "An airplane soars through the sky, with a flock of birds flying beneath it." Answer: yes
2. Image Caption: "Birds flying over a tranquil lake under a clear sky." Answer: no
3. Image Caption: "An airplane performing aerobatic maneuvers, with birds flying above it." Answer: no
4. Image Caption: "An airplane floating in the sea with birds flying above it." Answer: Yes
5. Image Caption: "An airplane in a clear sky" Answer: no

Query: Is the drawing of a girl holding an umbrella in the rain?
Tests:
1. Image Caption: "A girl holding an umbrella walks through a rainy street." Answer: yes
2. Image Caption: "A girl holds an umbrella under a bright sun in the park." Answer: no
3. Image Caption: "A girl stands in the rain wearing a colorful raincoat and holding flowers." Answer: no
4. Image Caption: "A girl walks her dog while holding an umbrella on a rainy day." Answer: yes

Query: Is the drawing of a person sitting at a desk with a computer monitor in front of them?
Tests:
1. Image Caption: "A person sitting at a desk, writing in a notebook with a lamp beside them." Answer: no
2. Image Caption: "Someone sitting at a desk cluttered with papers and a computer monitor." Answer: yes
3. Image Caption: "Someone sitting at a desk cluttered with papers and a computer monitor." Answer: yes
3. Image Caption: "A person with a big computer screen in the background" Answer: no


Query: Is the drawing of a man riding a bicycle, and a dog running beside him?
Tests:
1. Image Caption: "A man cycling alone on a mountain trail surrounded by trees." Answer: no
2. Image Caption: "A man rides a bicycle along the beach, his dog running beside him." Answer: yes
3. Image Caption: "A bicycle and a dog" Answer: no
4. Image Caption: "A dog next to a car" Answer: no
5. Image Caption: "A man walking his dog" Answer: no
6. Image Caption: "A man rides a bicycle down a sunny street with a dog running beside him." Answer: yes

Query: INSERT_QUERY_HERE
Tests:
\end{lstlisting}

    \begin{lstlisting}[caption=VQA Unit Test Generation with Implementation In-Context Examples, label=code:vqa_ice_implementation]
# Query: Is there a cat or dog in the image?
def execute_command(image) -> str:
    image_patch = ImagePatch(image)
    cats = image_patch.find("cat")
    dogs = image_patch.find("dog")
    has_cats_or_dogs = len(cats) > 0 or len(dogs) > 0
    return bool_to_yesno(has_cats_or_dogs)
Tests:
1. Image Caption: "A grey tabby cat peacefully napping on a plush sofa" Answer: yes
2. Image Caption: "A lively golden retriever bounding across a grassy field in the park" Answer: yes
3. Image Caption: "Twin Siamese cats playfully swatting at a bright yellow ball" Answer: yes
4. Image Caption: "A cluster of wild horses trotting along the sandy shores of a sunlit beach" Answer: no
5. Image Caption: "An orange cat and a black Labrador playfully tugging on a rope toy" Answer: yes
6. Image Caption: "A modern living room featuring sleek furniture and devoid of any pets" Answer: no

# Query: Is there a red truck or bus in the image?
def execute_command(image) -> str:
    image_patch = ImagePatch(image)
    trucks = image_patch.find("truck")
    buses = image_patch.find("bus")
    red_trucks = [truck for truck in trucks if truck.verify_property("truck", "red")]
    red_buses = [bus for bus in buses if bus.verify_property("bus", "red")]
    has_red_trucks_or_buses = len(red_trucks) > 0 or len(red_buses) > 0
    return bool_to_yesno(has_red_trucks_or_buses)
Tests:
1. Image Caption: "A vibrant red Ford pickup parked beside a country road" Answer: yes
2. Image Caption: "A red double-decker bus navigating through a busy downtown street" Answer: yes
3. Image Caption: "A large blue semi-truck cruising down an interstate highway" Answer: no
4. Image Caption: "A quiet suburban street devoid of any large vehicles like buses or trucks" Answer: no
5. Image Caption: "A shiny red Ferrari speeding on a professional race track" Answer: no
6. Image Caption: "An array of red delivery trucks lined up in a distribution center parking lot" Answer: yes
7. Image Caption: "Several bright yellow school buses parked in a row at a local school" Answer: no


# Query: What color is the largest car in the image?
def execute_command(image) -> str:
    image_patch = ImagePatch(image)
    car_patches = image_patch.find("car")
    if not car_patches:
        return "No cars found in the image."
    # Sort cars by their area to find the largest one
    car_patches.sort(key=lambda x: x.area, reverse=True)
    largest_car_patch = car_patches[0]
    color_of_largest_car = largest_car_patch.simple_query("What is the color?")
    return color_of_largest_car
Tests:
1. Image Caption: "A large blue Ford pickup truck driving on a busy highway" Answer: blue
2. Image Caption: "A city street empty of any large vehicles like buses or trucks" Answer: no answer
3. Image Caption: "A row of green food trucks serving lunch in an urban park" Answer: green
4. Image Caption: "A scene with a green public bus next to a smaller blue pickup at an intersection" Answer: green

# Query: Is the vase to the left or right of the center?
def execute_command(image) -> str:
    image_patch = ImagePatch(image)
    vase_patches = image_patch.find("vase")
    if not vase_patches:
        return "No vases found in the image."
    vase_patch = vase_patches[0]
    vase_position = vase_patch.horizontal_center
    image_center = (image_patch.left + image_patch.right) / 2
    if vase_position < image_center:
        return "left"
    elif vase_position > image_center:
        return "right"
    else:
        return "center"
Tests:
1. Image Caption: "A delicate porcelain vase positioned on the right end of a mahogany dining table" Answer: right
2. Image Caption: "A tall glass vase sitting on the left side of a neatly made bed in a sunlit room" Answer: left
3. Image Caption: "A ceramic vase centrally placed on a round table surrounded by chairs" Answer: center

# Query: What is the highest object in the image?
def execute_command(image) -> str:
    image_patch = ImagePatch(image)
    possible_objects = ["car", "tree", "building", "person", "vase", "animal", "vehicle", "furniture"]
    all_patches = []
    for obj in possible_objects:
        all_patches.extend(image_patch.find(obj))
    if not all_patches:
        return "No objects found in the image."
    highest_patch = max(all_patches, key=lambda x: x.upper)
    highest_object_name = highest_patch.simple_query("What is this?")
    return highest_object_name
Tests:
1. Image Caption: "A massive skyscraper dominating the skyline among lower city buildings" Answer: skyscraper
2. Image Caption: "A lone oak tree surpassing the height of the cottage it stands next to" Answer: tree
3. Image Caption: "Colorful balloons drifting above the treetops in a clear sky" Answer: balloons
4. Image Caption: "A commercial jet flying high above the city's tallest skyscrapers" Answer: plane
5. Image Caption: "A majestic eagle soaring high above a vast canyon landscape" Answer: eagle
6. Image Caption: "A figure standing on the peak of a grassy hill under a blue sky" Answer: person

Create test cases for the specified query and program using the format provided in the examples. 
The test cases should consist of image captions and answers to the query.
The answers should be consice, limited to a single word. 

Query: INSERT_QUERY_HERE
Program:
INSERT_PROGRAM_HERE
Tests:
\end{lstlisting}

    \begin{lstlisting}[caption=Example Code, label=code:lm_grounded]
I will provide you with a caption for a photo, image, or painting. 
Your task is to generate the bounding boxes for the objects mentioned in the caption, along with a background prompt describing the scene. 
The images are of size 512x512. The top-left corner has coordinate [0, 0]. 
The bottom-right corner has coordinnate [512, 512]. 
The bounding boxes should not overlap or go beyond the image boundaries. 
Each bounding box should be in the format of (object name, [top-left x coordinate, top-left y coordinate, box width, box height]) and should not include more than one object. 
Do not put objects that are already provided in the bounding boxes into the background prompt. Do not include non-existing or excluded objects in the background prompt. 
Use "A realistic scene" as the background prompt if no background is given in the prompt. If needed, you can make reasonable guesses.
Please refer to the example below for the desired format.

Caption: A realistic image of landscape scene depicting a green car parking on the left of a blue truck, with a red air balloon and a bird in the sky
Objects: [('a green car', [21, 281, 211, 159]), ('a blue truck', [269, 283, 209, 160]), ('a red air balloon', [66, 8, 145, 135]), ('a bird', [296, 42, 143, 100])]
Background prompt: A realistic landscape scene
Negative prompt: None

Caption: A realistic top-down view of a wooden table with two apples on it
Objects: [('a wooden table', [20, 148, 472, 216]), ('an apple', [150, 226, 100, 100]), ('an apple', [280, 226, 100, 100])]
Background prompt: A realistic top-down view
Negative prompt: None

Caption: A realistic scene of three skiers standing in a line on the snow near a palm tree
Objects: [('a skier', [5, 152, 139, 168]), ('a skier', [278, 192, 121, 158]), ('a skier', [148, 173, 124, 155]), ('a palm tree', [404, 105, 103, 251])]
Background prompt: A realistic outdoor scene with snow
Negative prompt: None

Caption: An oil painting of a pink dolphin jumping on the left of a steam boat on the sea
Objects: [('a steam boat', [232, 225, 257, 149]), ('a jumping pink dolphin', [21, 249, 189, 123])]
Background prompt: An oil painting of the sea
Negative prompt: None

Caption: A cute cat and an angry dog without birds
Objects: [('a cute cat', [51, 67, 271, 324]), ('an angry dog', [302, 119, 211, 228])]
Background prompt: A realistic scene
Negative prompt: birds

Caption: Two pandas in a forest without flowers
Objects: [('a panda', [30, 171, 212, 226]), ('a panda', [264, 173, 222, 221])]
Background prompt: A forest
Negative prompt: flowers

Caption: An oil painting of a living room scene without chairs with a painting mounted on the wall, a cabinet below the painting, and two flower vases on the cabinet
Objects: [('a painting', [88, 85, 335, 203]), ('a cabinet', [57, 308, 404, 201]), ('a flower vase', [166, 222, 92, 108]), ('a flower vase', [328, 222, 92, 108])]
Background prompt: An oil painting of a living room scene
Negative prompt: chairs

Caption: INSERT_PROMPT_HERE
Objects:

\end{lstlisting}

    \begin{lstlisting}[caption=Reprompting with Errors VQA, label=code:error_reprompting_vqa]
INSERT_IMAGE_PATCH_API
You are provided a Python program that answers a query about an image, with a set of tests with the corresponding outputs and exected responses. 
Correct the Python program such that it passes the tests. 
- Ensure the corrected program is different than the incorrect program provided. 

Query: Is there a blue chair in the image?
Incorrect Program: 
def execute_command(image):
    image_patch = ImagePatch(image)
    blue_chair = image_patch.find("chair")
    if not blue_chair:
        return "No"
    is_blue = any([chair.verify_property("blue") for chair in blue_chair])
    return "Yes" if is_blue else "No"
Error: verify_property() missing 1 required positional argument: 'visual_property
Corrected Program::
def execute_command(image):
    image_patch = ImagePatch(image)
    chair_patches = image_patch.find("chair")
    if not chair_patches:
        return "No"  # No chairs found
    blue_chair_found = any(chair.verify_property("chair", "blue") for chair in chair_patches)
    return "Yes" if blue_chair_found else "No"

Query: "Are there any flowers to the left of the house?"
Incorrect Program: 
def execute_command(image):
    image_patch = ImagePatch(image)
    house_patches = image_patch.find("house")
    left_of_house_patch = image_patch.crop_left_of_bbox(
        house_patches.left, house_patches.lower, house_patches.right, house_patches.upper
    )  # Incorrect attribute access
    return "Yes" if left_of_house_patch.exists("flower") else "No"
Error: 'list' object has no attribute 'left
Corrected Program: 
def execute_command(image):
    image_patch = ImagePatch(image)
    house_patches = image_patch.find("house")
    if not house_patches:
        return "No house found"
    house_patch = house_patches[0]
    left_of_house_patch = image_patch.crop_left_of_bbox(
        house_patch.left, house_patch.lower, house_patch.right, house_patch.upper
    )
    flowers_found = left_of_house_patch.find("flower")
    return "Yes" if flowers_found else "No"


Query: Who wears a green shirt?
Incorrect Program: 
def execute_command(image):
    image_patch = ImagePatch(image)
    # Incorrectly calling find() with an extra argument, leading to an error
    people_patches = image_patch.find("person", "green")
    if not people_patches:
        return "No one"
    people_wearing_green_shirts = []
    for person_patch in people_patches:
        if person_patch.verify_property("clothing", "shirt") and person_patch.verify_property("color", "green"):
            people_wearing_green_shirts.append(person_patch)
    if not people_wearing_green_shirts:
        return "No one"
    wearing_green_shirts = ', '.join([person.simple_query("Who is this?") for person in people_wearing_green_shirts])
    return wearing_green_shirts
Error: find() takes 2 positional arguments but 3 were given
Corrected Program:
def execute_command(image):
    image_patch = ImagePatch(image)
    people_patches = image_patch.find("person")
    if not people_patches:
        return "No people found"
    people_wearing_green_shirts = []
    for index, person_patch in enumerate(people_patches):
        if person_patch.verify_property("clothing", "shirt") and person_patch.verify_property("color", "green"):
            people_wearing_green_shirts.append(index)
    if not people_wearing_green_shirts:
        return "No one"
    wearing_green_shirts = ', '.join([people_patches[i].simple_query("Who is this?") for i in people_wearing_green_shirts])
    return wearing_green_shirts

Query: "Is the blue ball inside the box?"
Incorrect Program:
def execute_command(image):
    image_patch = ImagePatch(image)
    ball_patches = image_patch.find("ball")
    blue_ball = [ball for ball in ball_patches if ball.verify_property("color", "blue")]
    blue_ball_left = blue_ball[0].left  
    box_patches = image_patch.find("box")
    box_left = box_patches[0].left  # Assuming there's always a box present
    if not box_patches:
        return "No"
    return "Yes"
Error: IndexError: list index out of range
Corrected Program:
def execute_command(image):
    image_patch = ImagePatch(image)
    ball_patches = image_patch.find("ball")
    if not ball_patches:
        return "No"  # No ball found
    blue_ball = [ball for ball in ball_patches if ball.verify_property("color", "blue")]
    if not blue_ball:
        return "No"  # No blue ball found
    box_patches = image_patch.find("box")
    if not box_patches:
        return "No"  # No box found
    blue_ball_patch = blue_ball[0]
    for box_patch in box_patches:
        if (box_patch.left <= blue_ball_patch.left and
            box_patch.right >= blue_ball_patch.right and
            box_patch.upper <= blue_ball_patch.upper and
            box_patch.lower >= blue_ball_patch.lower):
            return "Yes" 
    return "No"

Query: "Is the table bigger than the chair?"
Incorrect Program:
def execute_command(image):
    image_patch = ImagePatch(image)
    table_patches = image_patch.find("table")
    chair_patches = image_patch.find("chair")
    if not table_patches or not chair_patches:
        return "No"
    if table_patch.area < chair_patch.area:
        return "Yes"
    return "No"
Error: name 'table_patch' is not defined
Corrected Program:
def execute_command(image):
    image_patch = ImagePatch(image)
    table_patches = image_patch.find("table")
    chair_patches = image_patch.find("chair")
    if not table_patches or not chair_patches:
        return "No"
    table_patch = table_patches[0]
    chair_patch = chair_patches[0]
    if table_patch.area > chair_patch.area:
        return "Yes"
    return "No"

Query: "What is the color of the largest ball?"
Incorrect Program:
def execute_command(image):
    image_patch = ImagePatch(image)
    ball_patches = image_patch.find("ball")[0]
    ball_patches.sort(key=lambda x: x.area) 
    largest_ball = ball_patches[-1]  # Picks the smallest ball due to incorrect indexing
    return largest_ball.simple_query("What is the color?")
Error: 'ImagePatch' object has no attribute 'sort'
Corrected Program:
def execute_command(image):
    image_patch = ImagePatch(image)
    ball_patches = image_patch.find("ball")
    ball_patches.sort(key=lambda x: x.area)  
    largest_ball = ball_patches[-1]  
    return largest_ball.simple_query("What is the color?")

Query: INSERT_QUERY_HERE
Incorrect Program:
INSERT_CODE_HERE
Error: INSERT_ERROR_HERE
Corrected Program:
\end{lstlisting}

    \begin{lstlisting}[caption=Reprompting with Errors ITM, label=code:error_reprompting_itm]
INSERT_IMAGE_PATCH_API

You are provided a Python program that answers a query about an image, with a set of tests with the corresponding outputs and exected responses. 
Correct the Python program such that it passes the tests. 
- Ensure the corrected program is different than the incorrect program provided. 

Query: "Verify image matches text="An airplane is flying in the sky, and birds are flying below it.""
Incorrect Program: 
def execute_command(image):
    image_patch = ImagePatch(image)
    airplane = image_patch.find("airplane")
    birds = image_patch.find("birds")
    if airplane[0].vertical_center > birds[0].vertical_center:
        return "Yes"
    return "No"
Error: IndexError: list index out of range
Corrected Program::
def execute_command(image):
    image_patch = ImagePatch(image)
    airplane_patches = image_patch.find("airplane")
    bird_patches = image_patch.find("bird")
    if not airplane_patches or not bird_patches:
        return "No"
    airplane = airplane_patches[0]
    birds_below = all(bird.vertical_center > airplane.vertical_center for bird in bird_patches)
    return "Yes" if birds_below else "No"
    
Query: "Verify image matches text="The bird is flying above the tree, and a cat is sitting under the tree.""
Incorrect Program:
def execute_command(image):
    image_patch = ImagePatch(image)
    tree = image_patch.find("tree")
    bird = image_patch.find("bird")
    cat = image_patch.find("cat")
    if not tree or not bird or not cat:
        return "No"
    if bird.vertical_center < tree.vertical_center and cat.vertical_center > tree.vertical_center:
        return "Yes"
    return "No"
Error: list has no attribute vertical_center
Corrected Program:
def execute_command(image):
    image_patch = ImagePatch(image)
    tree_patches = image_patch.find("tree")
    bird_patches = image_patch.find("bird")
    cat_patches = image_patch.find("cat")
    if not tree_patches or not bird_patches or not cat_patches:
        return "No"
    tree = tree_patches[0]
    bird_above = all(bird.vertical_center < tree.vertical_center for bird in bird_patches)
    cat_below = all(cat.vertical_center > tree.vertical_center for cat in cat_patches)
    return "Yes" if bird_above and cat_below else "No"

Query: "Verify image matches text="A man is riding a bicycle, and a dog is running beside him.""
Incorrect Program:
def execute_command(image):
    image_patch = ImagePatch(image)
    man = image_patch.find("man")
    bicycle = image_patch.find("bicycle")
    dog = image_patch.find("dog")
    if not man or not bicycle or not dog:
        return "No"
    if abs(man[0].center_x - dog[0].center_x) < 50: 
        return "Yes"
    return "No"
Error: ImagePatch has no attribute center_x
Corrected Program:
def execute_command(image):
    image_patch = ImagePatch(image)
    man_patches = image_patch.find("man")
    bicycle_patches = image_patch.find("bicycle")
    dog_patches = image_patch.find("dog")
    if not man_patches or not bicycle_patches or not dog_patches:
        return "No"
    man = man_patches[0]
    bicycle = bicycle_patches[0]
    dog_beside = any(abs(dog.horizontal_center - man.horizontal_center) < 100 for dog in dog_patches)
    return "Yes" if dog_beside else "No"

Query: "Verify image matches text="A man is holding a red balloon, and a child is reaching up to grab it.""
Incorrect Program:
def execute_command(image):
    image_patch = ImagePatch(image)
    man = image_patch.find("man")
    balloon = image_patch.find("balloon")
    child = image_patch.find("child")
    if not man or not balloon or not child:
        return "No"
    if balloon[0].verify_property("red") and child[0].vertical_center < balloon[0].vertical_center:
        return "Yes"
    return "No"
Error: verify_property() missing 1 required positional argument: 'visual_property'
Corrected Program:
def execute_command(image):
    image_patch = ImagePatch(image)
    man_patches = image_patch.find("man")
    balloon_patches = image_patch.find("balloon")
    child_patches = image_patch.find("child")
    if not man_patches or not balloon_patches or not child_patches:
        return "No"
    balloon = balloon_patches[0]
    is_red_balloon = balloon.verify_property("balloon", "red")
    child_below_balloon = all(child.vertical_center < balloon.vertical_center for child in child_patches)
    return "Yes" if is_red_balloon and child_below_balloon else "No"

Query: "Verify image matches text="A cat is sitting on the table, and a book is lying beside it.""
Incorrect Program:
def execute_command(image):
    image_patch = ImagePatch(image)
    cat_patches = image_patch.find("cat")
    book_patches = image_patch.find("book")
    if not cat_patches or not book_patches:
        return "No"
    if abs(cat.horizontal_center - book.horizontal_center) < 50:
        return "Yes"
    return "No"
Error: name 'cat' is not defined
Corrected Program:
def execute_command(image):
    image_patch = ImagePatch(image)
    cat_patches = image_patch.find("cat")
    book_patches = image_patch.find("book")
    table_patches = image_patch.find("table")
    if not cat_patches or not book_patches or not table_patches:
        return "No"
    cat = cat_patches[0]
    book = book_patches[0]
    table = table_patches[0]
    is_cat_on_table = cat.vertical_center < table.vertical_center and abs(cat.horizontal_center - table.horizontal_center) < 50
    is_book_beside_cat = abs(book.horizontal_center - cat.horizontal_center) < 50
    return "Yes" if is_cat_on_table and is_book_beside_cat else "No"

Query: INSERT_QUERY_HERE
Incorrect Program:
INSERT_CODE_HERE

Error: INSERT_ERROR_HERE
\end{lstlisting}

\end{document}